\documentclass[11pt]{article}

\usepackage[margin=1in]{geometry}

\usepackage{times,upgreek,morenotations,rotating}
\usepackage{graphicx} 
\usepackage{subfigure} 

\usepackage{natbib}

\usepackage{algorithm}
\usepackage{algorithmic}
\usepackage{hyperref}

\usepackage{makecell}

\usepackage{tabularx}
\usepackage{color, colortbl}
\definecolor{Gray}{gray}{0.9}

\newcommand{\mybs}{{\tiny $\blacksquare$}}

\usepackage[framemethod=TikZ]{mdframed}
\mdfdefinestyle{MyFrame}{%
    linecolor=gray,
    outerlinewidth=2pt,
    roundcorner=10pt,
    innertopmargin=0.5\baselineskip,
    innerbottommargin=0.5\baselineskip,
    innerrightmargin=10pt,
    innerleftmargin=10pt}

\author{
  Richard Nock\\
NICTA \& the Australian National University\\
  \texttt{richard.nock@nicta.com.au}
}

\begin{document} 

\title{Learning Games and Rademacher Observations Losses}

\maketitle 

\begin{abstract} 
It has recently been shown that supervised learning with the popular
logistic loss is equivalent to optimizing the exponential loss over
sufficient statistics about the class: \textit{Rademacher
  observations} (rados). 
We first show that this unexpected equivalence can actually be
\textit{generalized} to other example / rado losses, with necessary
and sufficient conditions for the equivalence, exemplified on four
losses that bear popular
names in various fields: exponential (boosting), mean-variance
(finance), Linear Hinge (on-line learning), ReLU (deep learning), and
unhinged (statistics). Second, we show that the generalization unveils a surprising new connection to \textit{regularized}
learning, and in particular a sufficient condition under which regularizing the loss over examples is
equivalent to \textit{regularizing the rados} (with Minkowski sums) in
the equivalent rado loss. This brings simple and powerful rado-based
learning algorithms for \textit{sparsity-controlling
regularization}, that we exemplify on a boosting algorithm
for the regularized exponential rado-loss,
which formally boosts over \textit{four} types of regularization, including the popular
ridge and lasso, and the recently coined \slope~  --- we obtain the first proven boosting
algorithm for this last regularization. Through our first contribution
on the equivalence of rado and example-based losses,
\radaboostNoL~appears to be an efficient proxy to boost the
regularized logistic loss \textit{over examples} using
\textit{whichever} of the four regularizers (and any linear combination
of them, \textit{e.g.}, for elastic net regularization). 
We are not aware of any
regularized logistic loss formal boosting algorithm with such a wide spectrum of regularizers.
Experiments display that
regularization consistently improves performances of rado-based
learning, and may challenge or beat the state of
the art of example-based
learning even when learning over small sets of rados. Finally, we connect regularization to
$\varepsilon$-differential privacy, and display how tiny budgets
(\textit{e.g.} $\varepsilon < 10^{-3}$) can be afforded on
big
domains while beating (protected) example-based learning.
\end{abstract} 

\section{Introduction}

A recent result has shown that minimising the popular logistic loss
over examples in supervised learning is equivalent to the minimisation
of the exponential loss over
\textit{sufficient statistics about the class} known as \textit{Rademacher
  observations} (rados, \citep{npfRO}), for the \textit{same}
classifier. In short, we fit a
classifier over data that is different from examples, and the same classifier generalizes well to
new \textit{observations}. It is known that sufficient
statistics carry the intractability of certain
processes that would otherwise be easy with data \citep{mCI}. In the case of
rados, such a computational caveat turns
out to be a \textit{big} advantage as privacy is
becoming crucial \citep{ecTE}. Indeed, rados allow to protect
data not just from a computational complexity
standpoint, but also from geometric, algebraic and statistical standpoints \citep{npfRO}, while still allowing to learn
accurate classifiers.

Two key problems remain: learning from
rados can compete experimentally with learning from examples, but there is a
gap to reduce for rados to be not just a
good material to learn from in a privacy setting, but also a serious
alternative to learning from examples \textit{at large}, yielding new
avenues to supervised learning. Second, theoretically
speaking, it is crucial to understand if this equivalence holds only for the logistic
and exponential losses, or if it can be generalised and shed new
light on losses and their minimisation.

In this paper, we provide answers to these two questions, with four main contributions.
Our first contribution is to show that this generalization indeed holds: other example losses admit equivalent
losses in the rado world, meaning in particular that their minimiser
classifier is the \textit{same}, regardless of the dataset of
examples. The technique we use exploits a two-player zero sum game
representation of convex losses, that has been very
useful to analyse boosting algorithms \citep{sTB,tAP}, with one key
difference: payoffs are non-linear convex, eventually non-differentiable. These also resemble the entropic
dual losses \citep{rfwmGM}, with the difference that 
we do not enforce conjugacy over the simplex. The
conditions of the game are slightly different for examples and rados. We provide necessary and
sufficient conditions for the resulting losses over examples and rados
to be equivalent. Informally, equivalence happens iff the convex
functions of the games satisfy a symmetry relationship \textit{and} the weights satisfy a
linear system of equations. We give \textit{four cases} of this equivalence. It turns out that the losses involved bear
popular names in different communities, even when not all of them are systematically
used as losses \textit{per se}: exponential, logistic, square, mean-variance, ReLU,
linear Hinge, and unhinged losses \citep{nhRL,gwLH,nnOT,tAP,vSL,vmwLW} (and
many others).

Our second contribution came unexpectedly through this
equivalence. Regularizing a loss is common in machine learning \citep{bjmoOW}. We show a
sufficient condition for the equivalence under which regularizing the
example loss is equivalent to regularizing \textit{the rados} in the
rado loss, \textit{i.e.} making a Minkowski sum of the rado set with a
classifier-based set. This
property is \textit{independent} of the regularizer, and holds for \textit{all}
four cases of equivalence. 

Third, we propose a boosting algorithm, \radaboostNoL, that learns a
classifier from rados using the exponential regularized rado loss,
with regularization choice belonging to the ridge, lasso, $\ell_\infty$, or the
recently coined \slope~\citep{bvsscSA}. 
A key
property is that \radaboostNoL~\textit{bypasses} the Minkowski sums to
compute regularized rados. It is therefore computationally
efficient.
Experiments display that
\radaboostNoL~is all the better vs
\adaboostSS~(unregularized and $\ell_1$-regularized) as
the domain gets larger, and is able to learn both accurate and sparse classifiers,
making it a good contender for supervised learning at large on big domains. From a theoretical standpoint, we show that
for \textit{any} of these four regularizations, \radaboostNoL~is a boosting
algorithm --- thus, through our first contribution, \radaboostNoL~is an efficient proxy to boost the
regularized logistic loss \textit{over examples} using
\textit{whichever} of the four regularizers, and by extension, any
linear combination of them (\textit{e.g.}, for elastic net
regularization \citep{zhRA}). We are not aware of any
regularized logistic loss formal boosting algorithm with such a wide spectrum of regularizers.

Our fourth contribution is a direct application of our findings to
$\varepsilon$-differential privacy (DP). We protect directly the \textit{examples}, granting the property that \textit{all} subsequent stages are
DP as well. We show
theoretically that a most popular mechanism \citep{drTA} used to protect
examples in rados amounts to a surrogate
form of regularization of the \textit{clean} examples' loss; furthermore,
the amount of noise can be
commensurate to the one for a direct protection of
\textit{examples}. In other words, since rados' norm may be much larger
than examples' (\textit{e.g.} on big domains), we can expect noise to be much less
damaging if learning from protected rados, and afford
tiny budgets (\textit{e.g.} $\varepsilon \approx 10^{-4}$) at
little cost in accuracy. Experiments validate this intuition.

The rest of this paper is as follows. $\S$\ref{secequ}, \ref{secreg}
and \ref{secboo} respectively present the equivalence between example
and rado losses, its extension to regularized learning and
\radaboostNoL. $\S$\ref{secdp}, \ref{secexp} and \ref{seccon}
respectively present differential
privacy vs regularized rado losses, detail experiments, and
conclude. In order not to laden the paper's body, an appendix, starting page \pageref{secappendix} of this draft, contains the proofs and additional
theoretical and experimental results.

\section{Games and equivalent example/rado losses}\label{secequ}

We first start by defining and analysing our general two players game
setting. To avoid notational load, we shall not put immediately the
learning setting at play, considering for the moment that the learner
fits a general vector $\ve{z}\in {\mathbb{R}}^m$, which depends both
on data (examples or rados) and classifier.
Let $[m] \defeq \{1, 2, ..., m\}$ and $\Sigma_m \defeq \{-1,1\}^m$,
for $m>0$. Let $\varphi_{\lab{e}} : {\mathbb{R}}\rightarrow
{\mathbb{R}}$ and $\varphi_{\lab{r}} : {\mathbb{R}}\rightarrow
{\mathbb{R}}$ two convex and lower-semicontinuous \textit{generators}. We define functions
${\mathcal{L}}_{\lab{e}} : {\mathbb{R}}^m\times {\mathbb{R}}^m
\rightarrow {\mathbb{R}}$ and ${\mathcal{L}}_{\lab{r}} :
{\mathbb{R}}^{2^m}\times {\mathbb{R}}^m
\rightarrow {\mathbb{R}}$:
\begin{eqnarray}
{\mathcal{L}}_{\lab{e}}(\ve{p}, \ve{z}) & \defeq & \sum_{i\in [m]} p_i z_i
+ \upmu_{\lab{e}} \sum_{i\in [m]} \varphi_{\lab{e}}(p_i)\:\:,\label{defopte}\\
{\mathcal{L}}_{\lab{r}}(\ve{q}, \ve{z}) & \defeq & 
\sum_{{\mathcal{I}} \subseteq [m]} q_{{\mathcal{I}}} \sum_{i \in
  {\mathcal{I}}} z_i + \upmu_{\lab{r}} \sum_{{\mathcal{I}} \subseteq
  [m]} \varphi_{\lab{r}} (q_{{\mathcal{I}}})\:\:,\label{defoptr}
\end{eqnarray}
where $\upmu_{\lab{e}},\upmu_{\lab{r}} > 0$ do not depend on $\bm{z}$. For the notation to be meaningful, the coordinates in $\ve{q}$ are
assumed (wlog) to be in bijection with $2^{[m]}$. The dependence of
both problems in their respective \textit{generators} is implicit and shall be
clear from context. The adversary's goal is to fit
\begin{eqnarray}
\ve{p}^*(\ve{z}) & \defeq & \arg\min_{\ve{p}\in {\mathbb{R}}^m}
{\mathcal{L}}_{\lab{e}}(\ve{p}, \ve{z})\:\:, \label{opte}\\
\ve{q}^*(\ve{z}) & \defeq & \arg\min_{\ve{q}\in {\mathbb{H}}^{2^m}}
{\mathcal{L}}_{\lab{r}}(\ve{q}, \ve{z})\:\:,\label{optr}
\end{eqnarray}
with ${\mathbb{H}}^{2^m} \defeq \{\ve{q} \in {\mathbb{R}}^{2^m} :
\ve{1}^\top \ve{q} = 1\}$, so as to attain
\begin{eqnarray}
{\mathcal{L}}_{\lab{e}}^*(\ve{z}) & \defeq &
{\mathcal{L}}_{\lab{e}}(\ve{p}^*(\ve{z}),\ve{z})\:\:, \label{defle}\\
{\mathcal{L}}_{\lab{r}}^*(\ve{z}) & \defeq &
{\mathcal{L}}_{\lab{r}}(\ve{q}^*(\ve{z}),\ve{z})\:\:,\label{deflr}
\end{eqnarray}
and let $\partial {\mathcal{L}}^*_{\lab{e}}(\ve{z})$ and $\partial
{\mathcal{L}}^*_{\lab{r}}(\ve{z})$ denote their subdifferentials. We view the learner's task as the problem of maximising the
corresponding problems in eq. (\ref{defle}) (with examples) or
(\ref{deflr}) (with rados), or equivalently minimising negative the
corresponding function, then called a \textit{loss function}. The question of when these two problems are
equivalent from the learner's standpoint motivates the following definition.
\begin{definition}\label{sequiv}
Two generators $\varphi_{\lab{e}}, \varphi_{\lab{r}}$ are said
\textbf{proportionate} iff for any $m>0$, there exists
$(\upmu_{\lab{e}},\upmu_{\lab{r}})$ such that
\begin{eqnarray}
{\mathcal{L}}_{\lab{e}}^*(\ve{z}) & = &
{\mathcal{L}}_{\lab{r}}^*(\ve{z}) + b\:\:,\forall \ve{z}\in
{\mathbb{R}}^m\:\:.\label{ppa}
\end{eqnarray}
\end{definition}
($b$ does not depend on $\ve{z}$) $\forall m \in {\mathbb{N}}_*$,
let
\begin{eqnarray}
\matrice{G}_m & \defeq & 
\left[
\begin{array}{cc}
\ve{0}^\top_{2^{m-1}} & \ve{1}^\top_{2^{m-1}}\\
\matrice{G}_{m-1} & \matrice{G}_{m-1}
\end{array}
\right] \:\: (\in \{0,1\}^{m\times 2^m})
\end{eqnarray}
if $m>1$, and $\matrice{G}_1 \defeq [0\:\: 1]$ otherwise ($\ve{z}_d$
denotes a vector in ${\mathbb{R}}^d$). Each column of
$\matrice{G}_m$  is the binary indicator vector for the
edge vectors summed in a rado; wlog, we let these to
give the bijection between $2^{[m]}$ and coordinates of
$\ve{q}^{(*)}(\ve{z})$. 
\begin{theorem}\label{thmstrongrc}
$\varphi_{\lab{e}}, \varphi_{\lab{r}}$ are proportionate iff the
optimal solutions $\ve{p}^*(\ve{z})$ and $\ve{q}^*(\ve{z})$ to eqs
(\ref{opte}) and (\ref{optr}) satisfy
\begin{eqnarray}
\ve{p}^*(\ve{z}) & \in & \partial {\mathcal{L}}^*_{\lab{r}}(\ve{z})\:\:,\label{pp2}\\
\matrice{G}_m \ve{q}^*(\ve{z}) & \in & \partial  {\mathcal{L}}^*_{\lab{e}}(\ve{z})\:\:.\label{qq2}
\end{eqnarray}
In the case where $\varphi_{\lab{e}}, \varphi_{\lab{r}}$ are
differentiable, they are proportionate iff $\ve{p}^*(\ve{z}) = \matrice{G}_m \ve{q}^*(\ve{z})$.
\end{theorem}
(Proof in Appendix, Subsection
\ref{proof_thm_thmstrongrc}) 
Theorem \ref{thmstrongrc} gives a necessary and sufficient condition
for two generators to be proportionate. It does not say how
to construct one from the other, if possible. We now show that it is indeed
possible and prune the search space: if $\varphi_{\lab{e}}$
is proportionate to some $\varphi_{\lab{r}}$, then it has to be a ``symmetrized''
version of $\varphi_{\lab{r}}$, according to the following definition.
\begin{definition}\label{defsymz}
Let $\varphi_{\lab{r}}$ such that $\mathrm{dom}(\varphi_{\lab{r}})
\supseteq (0,1)$. We call $\varphi_{\lab{s}({\lab{r}})} (z) \defeq \varphi_{\lab{r}}(z) + \varphi_{\lab{r}}(1-z)$ the symmetrisation of $\varphi_{\lab{r}}$.
\end{definition}

\begin{lemma}\label{lemgens}
If $\varphi_{\lab{e}}$ and $\varphi_{\lab{r}}$ are
proportionate, then $\varphi_{\lab{e}}(z) =
(\upmu_{\lab{r}}/\upmu_{\lab{e}})\cdot \varphi_{\lab{s}({\lab{r}})}(z)
+ (b/\upmu_{\lab{e}})$, where $b$ appears in eq. (\ref{ppa}).
\end{lemma}
(Proof in Appendix, Subsection
\ref{proof_lem_lemgens}) To summarize, $\varphi_{\lab{e}}$ and $\varphi_{\lab{r}}$ are
proportionate iff (i) they meet the structural property that $\varphi_{\lab{e}}$ is
  (proportional to) the symmetrized version of $\varphi_{\lab{r}}$ (according to Definition \ref{defsymz}),
  and (ii) the optimal solutions $\ve{p}^*(\ve{z})$ and $\ve{q}^*(\ve{z})$
  to problems (\ref{defopte}) and (\ref{defoptr})
  satisfy the conditions of Theorem \ref{thmstrongrc}. Depending on
  the direction, we have two cases to craft proportionate
  generators. First, if we have $\varphi_{\lab{r}}$, then necessarily
  $\varphi_{\lab{e}} \propto \varphi_{\lab{s}({\lab{r}})}$ so we merely
  have to check Theorem \ref{thmstrongrc}. Second, if we have
  $\varphi_{\lab{e}}$, then it matches Definition
  \ref{defsymz}\footnote{Alternatively, $-\varphi_{\lab{e}}$ is permissible
  \citep{kmOTj}.}. In this case, we 
have to find $\varphi_{\lab{r}} = f + g$ where
$g(z) = -g(1-z)$ and $\varphi_{\lab{e}}(z) = f(z) + f(1-z)$.


We now come back to
${\mathcal{L}}_{\lab{e}}^*(\ve{z})$,
${\mathcal{L}}_{\lab{r}}^*(\ve{z})$ as defined in Definition \ref{sequiv},
and make the connection with
$\labsmall{e}$xample and $\labsmall{r}$ado losses. In the next definition, an $\lab{e}$-loss
$\ell_{\lab{e}}(\ve{z})$ is a function defined over the
coordinates of $\ve{z}$, and a $\lab{r}$-loss
$\ell_{\lab{r}}(\ve{z})$ is a function defined over the
subsets of sums of coordinates. 
Functions can depend on other
parameters as well.
\begin{definition}\label{defeqer}
Suppose $\lab{e}$-loss $\ell_{\lab{e}}(\ve{z})$ and $\lab{r}$-loss
$\ell_{\lab{r}}(\ve{z})$ are such that there exist (i) $f_{\lab{e}} : {\mathbb{R}} \rightarrow
{\mathbb{R}}$ and $f_{\lab{r}}(z) : {\mathbb{R}} \rightarrow
{\mathbb{R}}$ both strictly increasing and such that $\forall \ve{z}\in {\mathbb{R}}^m$,
\begin{eqnarray}
-{\mathcal{L}}_{\lab{e}}^*(\ve{z}) & = & f_{\lab{e}}\left(\ell_{\lab{e}}(\ve{z})\right)\:\:, \label{losse}\\
-{\mathcal{L}}_{\lab{r}}^*(\ve{z}) & = & f_{\lab{r}}\left(\ell_{\lab{r}}(\ve{z})\right)\:\:.\label{lossr}
\end{eqnarray}
Then the couple $(\ell_{\lab{e}},\ell_{\lab{r}})$ is
called a couple of equivalent example-rado losses.
\end{definition}
Hereafter, we just write $\varphi_{\lab{s}}$ instead of
$\varphi_{\lab{s}(\lab{r})}$.
\begin{lemma}\label{srclog}
$\varphi_{\lab{r}}(z) \defeq z\log z - z$  is proportionate to $\varphi_{\lab{e}}
\defeq \varphi_{\lab{s}} = z\log z + (1-z)\log(1-z) - 1$, whenever $\upmu_{\lab{e}} = \upmu_{\lab{r}}$.
\end{lemma}
(Proof in Appendix, Subsection
\ref{proof_lem_srclog})

\begin{corollary}\label{corlog}
The following example and rado losses are equivalent for any $\upmu > 0$:
\begin{eqnarray}
\ell_{\lab{e}}(\ve{z}, \upmu) & = & \sum_{i\in [m]} \log \left(1+\exp\left(-\frac{1}{\upmu}
  \cdot z_i\right)\right)\:\:,\label{ploge}\\
\ell_{\lab{r}}(\ve{z}, \upmu) & = & \sum_{{\mathcal{I}} \subseteq [m]} {\exp\left(-\frac{1}{\upmu}
  \cdot \sum_{i \in
  {\mathcal{I}}} z_i\right)}\:\:.
\end{eqnarray}
\end{corollary}
(Proof in Appendix, Subsection
\ref{proof_cor_corlog})

\begin{lemma}\label{srcsql}
$\varphi_{\lab{r}}(z) \defeq (1/2)\cdot z^2$ is proportionate to $\varphi_{\lab{e}}
\defeq \varphi_{\lab{s}} = (1/2)\cdot(1-2z(1-z))$ whenever $\upmu_{\lab{e}} = \upmu_{\lab{r}}/2^{m-1}$.
\end{lemma}
(Proof in Appendix, Subsection
\ref{proof_lem_srcsql})

\begin{corollary}\label{corsql}
The following example and rado losses are equivalent, for any $\upmu > 0$:
\begin{eqnarray}
\ell_{\lab{e}}(\ve{z}, \upmu) & = & \sum_{i\in [m]} 
\left(1- \frac{1}{\upmu}
  \cdot z_i\right)^2\:\:,\label{psqle}\\
\ell_{\lab{r}}(\ve{z}, \upmu) & = & -\left(\expect_{{\mathcal{I}} }\left[ \frac{1}{\upmu}\cdot \sum_{i \in
  {\mathcal{I}}} z_i\right]  - \upmu \cdot \var_{{\mathcal{I}} }\left[ \frac{1}{\upmu}\cdot \sum_{i \in
  {\mathcal{I}}} z_i\right]\right)\:\:,\label{psqlr}
\end{eqnarray}
where $\expect_{{\mathcal{I}} }[X({\mathcal{I}})]$ and $\var_{{\mathcal{I}}
  }[X({\mathcal{I}})]$ denote the expectation and variance of $X$ wrt
    uniform weights on ${\mathcal{I}} \subseteq [m]$.
\end{corollary}
(Proof in Appendix, Subsection
\ref{proof_cor_corsql}) We now investigate cases of non differentiable
proportionate generators, the first of which is self-proportionate
($\varphi_{\lab{e}}=\varphi_{\lab{r}}$). We let $\upchi_{\mathcal{A}}(z)$ be
the indicator function: $\upchi_{{\mathcal{A}}}(z) \defeq 0$ if $z \in
{\mathcal{A}}$ (and $+\infty$ otherwise),
convex since ${\mathcal{A}} = [0,1]$ is convex.
\begin{lemma}\label{srcrelu}
\hspace{-0.1cm}$\varphi_{\lab{r}}(z) \defeq \upchi_{[0,1]}(z)$ is self-proportionate,
\hspace{-0.1cm}$\forall \upmu_{\lab{e}}, \upmu_{\lab{r}}$.
\end{lemma}
(Proof in Appendix, Subsection
\ref{proof_lem_srcrelu})
\begin{corollary}\label{correlu}
The following example and rado losses are equivalent, for any
$\upmu_{\lab{e}} , \upmu_{\lab{r}}$:
\begin{eqnarray}
\ell_{\lab{e}}(\ve{z}, \upmu_{\lab{e}}) & = & \sum_{i\in [m]} \max\left\{0, - \frac{1}{\upmu_{\lab{e}}}\cdot z_i\right\}\:\:,\label{pmaxe}\\
\ell_{\lab{r}}(\ve{z}, \upmu_{\lab{r}}) & = & \max \left\{0,
  \max_{{\mathcal{I}} \subseteq [m]} \left\{-
\frac{1}{\upmu_{\lab{r}}} \cdot \sum_{i\in {\mathcal{I}}} z_i\right\}\right\}\:\:.\label{pmaxr}
\end{eqnarray}
\end{corollary}
(Proof in Appendix, Subsection
\ref{proof_cor_correlu})

\begin{lemma}\label{srcunh}
$\varphi_{\lab{r}}(z) \defeq \upchi_{\left[\frac{1}{2^m}, \frac{1}{2}\right]}(z)$ is proportionate to $\varphi_{\lab{e}}
\defeq \varphi_{\lab{s}} = \upchi_{\left\{\frac{1}{2}\right\}}(z)$, for any
$\upmu_{\lab{e}} , \upmu_{\lab{r}}$.
\end{lemma}
(Proof in Appendix, Subsection
\ref{proof_lem_srcunh})
\begin{corollary}\label{corunh}
The following example and rado losses are equivalent, for any
$\upmu_{\lab{e}} , \upmu_{\lab{r}}$:
\begin{eqnarray}
\ell_{\lab{e}}(\ve{z}, \upmu_{\lab{e}}) & = & \sum_i -\frac{1}{\upmu_{\lab{e}}}\cdot  z_i\:\:,\label{punhe}\\
\ell_{\lab{r}}(\ve{z}, \upmu_{\lab{r}}) & = & \expect_{{\mathcal{I}} }\left[ -\frac{1}{\upmu_{\lab{r}}}\cdot \sum_{i \in
  {\mathcal{I}}} z_i\right]  \:\:.\label{punhr}
\end{eqnarray}
\end{corollary}
(Proof in Appendix, Subsection
\ref{proof_cor_corunh}) Table \ref{sumeq} summarizes the four equivalent example and
rado losses.
\begin{table*}[t]
\begin{center}
{
\begin{tabular}{c|cc|c|cc|c}
\hline \hline
  $\#$ & $\ell_{\lab{e}}(\ve{z}, \upmu_{\lab{e}})$ & $\ell_{\lab{r}}(\ve{z},
  \upmu_{\lab{r}})$ & $\varphi_{\lab{r}}(z)$ & 
  $\upmu_{\lab{e}}$ and $\upmu_{\lab{r}}$ &  $a_{\lab{e}}$ & Ref\\\hline
 I & $\sum_{i\in [m]} \log \left(1+\exp\left(z^{\lab{e}}_i\right)\right)$ & $\sum_{{\mathcal{I}} \subseteq [m]} {\exp\left(z^{\lab{r}}_{\mathcal{I}}\right)}$ & $z\log z - z$ & $\forall
\upmu_{\lab{e}} = \upmu_{\lab{r}}$ &  $\upmu_{\lab{e}}$  & Cor. \ref{corlog}\\
 II & $\sum_{i\in [m]} 
\left(1+z^{\lab{e}}_i\right)^2$ & $-(\expect_{{\mathcal{I}} }\left[ -z^{\lab{r}}_{\mathcal{I}} \right]  - \upmu_{\lab{r}} \cdot \var_{{\mathcal{I}} }\left[ -z^{\lab{r}}_{\mathcal{I}} \right])$ & $(1/2)\cdot z^2$ & $\forall
\upmu_{\lab{e}} = \upmu_{\lab{r}}$ &  $\upmu_{\lab{e}}/4$ & Cor. \ref{corsql}\\
 III & $\sum_{i\in [m]} \max\left\{0, z^{\lab{e}}_i\right\}$ & $\max\left\{0,\max_{{\mathcal{I}} \subseteq [m]}  \{z^{\lab{r}}_{\mathcal{I}}\}\right\}$
& $\upchi_{[0,1]}(z)$ & $\forall
\upmu_{\lab{e}}, \upmu_{\lab{r}}$ &  $\upmu_{\lab{e}}$ & Cor. \ref{correlu}\\
IV & $\sum_i z^{\lab{e}}_i$ & $\expect_{{\mathcal{I}} }\left[ z^{\lab{r}}_{\mathcal{I}} \right]$ & $\upchi_{\left[\frac{1}{2^m},
  \frac{1}{2}\right]}(z) $ & $\forall
\upmu_{\lab{e}}, \upmu_{\lab{r}}$ &  $\upmu_{\lab{e}}$ & Cor. \ref{corunh}\\
\hline\hline
\end{tabular}
}
\end{center}
\caption{Examples of equivalent example and rado
  losses. Names of the rado-losses $\ell_{\lab{r}}(\ve{z},
  \upmu_{\lab{r}})$ are respectively the Exponential (I),
  Mean-variance (II), ReLU (III) and Unhinged (IV) rado loss.
We use shorthands $z^{\lab{e}}_i \defeq
  -(1/\upmu_{\lab{e}})\cdot z_i$ and $z^{\lab{r}}_{\mathcal{I}} \defeq
  -(1/\upmu_{\lab{r}})\cdot \sum_{i \in
  {\mathcal{I}}} z_i$. Parameter $a_{\lab{e}}$ appears
in eq. (\ref{pfe}). Column
  ``$\upmu_{\lab{e}}$ and $\upmu_{\lab{r}}$'' gives the constraints
  for the equivalence to hold (see text for details).}\label{sumeq}
\end{table*}

\section{Learning with (rado) regularized losses}\label{secreg}


We now plug the learning setting. The learner is given a set of examples ${\mathcal{S}} = \{(\ve{x}_i,
y_i), i = 1, 2, ..., m\}$ where $\ve{x}_i \in {\mathbb{R}}^d$, $y_i\in
\Sigma_1$ (for $i=1, 2, ..., m$). It returns a classifier $h : {\mathbb{R}}^d
\rightarrow {\mathbb{R}}$ from a predefined set ${\mathcal{H}}$. Let $z_i(h) \defeq y h(\bm{x}_i)$ and define
$\ve{z}(h)$ as the corresponding vector in ${\mathbb{R}}^m$, which we
plug in the
losses of Table \ref{sumeq} to obtain the corresponding example and
rado losses. Losses simplify conveniently when ${\mathcal{H}}$ consists of linear classifiers, $h(\bm{x})
\defeq \ve{\theta}^\top \ve{x}$ for some $\ve{\theta}\in
\Theta\subseteq {\mathbb{R}}^d$. In this
case, the example loss can be described using edge vectors
${\mathcal{S}}_{\lab{e}} \defeq \{y_i\cdot \ve{x}_i, i = 1, 2, ..., m\}$
since $z_i = \ve{\theta}^\top (y_i\cdot \ve{x}_i)$, and the rado loss
can be described using rademacher observations \citep{npfRO}, since $\sum_{i \in
  {\mathcal{I}}} z_i = \ve{\theta}^\top \rado_{\ve{\sigma}}$ for
$\sigma_i = y_i$ iff $i\in {\mathcal{I}}$ (and $-y_i$ otherwise) and
$\rado_{\ve{\sigma}}\defeq (1/2)\cdot \sum_i (\sigma_i + y_i)\cdot
\ve{x}_i$. Let us define ${\mathcal{S}}^*_{\lab{r}} \defeq
\{\rado_{\ve{\sigma}},\ve{\sigma}\in \Sigma_m\}$ the set of all rademacher
observations. We rewrite any couple of equivalent example and rado
losses as $\ell_{\lab{e}}({\mathcal{S}}_{\lab{e}}, \ve{\theta})$ and
$\ell_{\lab{r}}({\mathcal{S}}^*_{\lab{r}}, \ve{\theta})$
respectively\footnote{To prevent notational overload, we blend the
  notions of (pointwise) loss and (samplewise) risk, as just ``losses''.},
omitting parameters $\upmu_{\lab{e}}$ and $\upmu_{\lab{r}}$,
assumed to be fixed beforehand for the equivalence to hold (see Table
\ref{sumeq}). Let us regularize the example loss, 
so that the learner's goal is to minimize
\begin{eqnarray}
\ell_{\lab{e}}({\mathcal{S}}_{\lab{e}}, \ve{\theta},\Omega) & \defeq &
\ell_{\lab{e}}({\mathcal{S}}_{\lab{e}}, \ve{\theta}) + \Omega(\ve{\theta})\:\:,\label{defreg}
\end{eqnarray}
with $\Omega$ a
regularizer \citep{bjmoOW}. The following
shows that when $f_{\lab{e}}$ in eq. (\ref{losse}) is linear, there is
a rado-loss equivalent to this regularized loss,
\textit{regardless} of $\Omega$. 
\begin{theorem}\label{regrado}
Suppose ${\mathcal{H}}$ contains linear classifiers. Let
$(\ell_{\lab{e}}({\mathcal{S}}_{\lab{e}}, \ve{\theta}),
\ell_{\lab{r}}({\mathcal{S}}^*_{\lab{r}}, \ve{\theta}))$ be any couple of
equivalent example-rado losses such that
$f_{\lab{e}}$ in eq. (\ref{losse}) is linear:
\begin{eqnarray}
f_{\lab{e}}(z) & = & a_{\lab{e}}\cdot z + b_{\lab{e}}\:\:,\label{pfe}
\end{eqnarray}
for some $a_{\lab{e}}>0, b_{\lab{e}}\in {\mathbb{R}}$. 
Then for any regularizer $\Omega(.)$, the regularized example loss
$\ell_{\lab{e}}({\mathcal{S}}_{\lab{e}}, \ve{\theta},\Omega)$ is
equivalent to rado loss
$\ell_{\lab{r}}({\mathcal{S}}^{*,\Omega,\ve{\theta}}_{\lab{r}}, \ve{\theta})$
computed over
\textbf{regularized} rados:
\begin{eqnarray}
{\mathcal{S}}^{*,\Omega,\ve{\theta}}_{\lab{r}} & \defeq &
{\mathcal{S}}^*_{\lab{r}} \oplus \{-\tilde{\Omega}(\ve{\theta}) \cdot \ve{\theta}\}\:\:,
\end{eqnarray}
where $\oplus$ is Minkowski sum and $\tilde{\Omega}(\ve{\theta})
\defeq a_{\lab{e}}\cdot\Omega(\ve{\theta})/\|\ve{\theta}\|_2^2$ if $\ve{\theta} \neq
\ve{0}$ (and $0$ otherwise, assuming wlog $\Omega(\ve{0}) = 0$).
\end{theorem}
(Proof in Appendix, Subsection
\ref{proof_thm_regrado})
Theorem \ref{regrado} applies to all rado losses (I-IV) in Table
\ref{sumeq}. The effect of regularization on rados is intuitive from
the margin standpoint: assume that a ``good'' classifier $\ve{\theta}$ is one that
ensures lowerbounded inner products $\ve{\theta}^\top \ve{z} \geq \tau$ for some
\textit{margin threshold} $\tau$. Then any good classifier on a
regularized rado $\ve{\rado}_{\ve{\sigma}}$ shall actually meet, over \textit{examples}, 
\begin{eqnarray}
\sum_{i:y_i = \sigma_i} \ve{\theta}^\top (y_i\cdot \ve{x}_i)  & \geq & \tau + a_{\lab{e}}\cdot
\Omega(\ve{\theta})\label{eqrado}\:\:.
\end{eqnarray} 
Notice that ineq (\ref{eqrado}) ties an "accuracy" of $\ve{\theta}$
(edges, left
hand-side) and its sparsity (right-hand side). One important question is the way the minimisation of
the regularized rado loss impacts the minimisation of the regularized examples loss when
one \textit{subsamples} the rados, and learns $\ve{\theta}$ from some
${\mathcal{S}}_{\lab{r}}\subseteq {\mathcal{S}}^*_{\lab{r}}$ with
eventually $|{\mathcal{S}}_{\lab{r}}|\ll |{\mathcal{S}}^*_{\lab{r}}|$. We
give an answer for the log-loss \citep{npfRO} (row I
in Table \ref{sumeq}), and
for this objective define the $\Omega$-regularized exp-rado-loss computed over
${\mathcal{S}}_{\lab{r}}$, with $|{\mathcal{S}}_{\lab{r}}| = n$ and
$\upomega > 0$ user-fixed:
\begin{eqnarray}
\lefteqn{\ell^{\tiny{\mathrm{exp}}}_{\lab{r}}({\mathcal{S}}_r,
  \ve{\theta}, \Omega)}\nonumber\\
 & \defeq &  \frac{1}{n}\cdot \sum_{j \in [n]}
 \exp\left(-\ve{\theta}^\top \left(\ve{\rado}_{j} - \upomega\cdot \frac{\Omega(\ve{\theta})}{\|\ve{\theta}\|_2^2}\cdot
    \ve{\theta}\right)\right)\:\:,\label{defregexp}
\end{eqnarray}
whenever $\ve{\theta}\neq \ve{0}$ (otherwise, we discard the
factor depending on $\upomega$ in the formula). We assume that
$\Omega$ is a norm, and let $\ell^{\tiny{\mathrm{exp}}}_{\lab{r}}({\mathcal{S}}_r,
  \ve{\theta})$ denote the unregularized loss ($\upomega = 0$ in
  eq. (\ref{defregexp})), and we let $\ell^{\tiny{\mathrm{log}}}_{\lab{e}}\left({\mathcal{S}}_{\lab{e}}, \ve{\theta}, \Omega\right) \defeq (1/m)\sum_{i}
\log\left(1+\exp\left(-\ve{\theta}^\top
    (y_i \cdot \ve{x}_i)\right)\right) + \Omega(\ve{\theta})$ denote
the $\Omega$-regularized log-loss. Notice that we normalize losses. We 
define the open ball ${\mathcal{B}}_{\Omega}(\ve{0},r) \defeq \{\ve{x} \in {\mathbb{R}}^d :
\Omega(\ve{x}) < r\}$ and $r^\star_\pi \defeq (1/m) \cdot \max_{{\mathcal{S}}^*_{\lab{r}}}
  \Omega^\star(\ve{\rado}_{\ve{\sigma}})$, where $\Omega^\star$
  is the dual norm of $\Omega$. The following Theorem is a direct
  application of Theorem 3 in
\citep{npfRO}, and shows mild conditions on ${\mathcal{S}}_{\lab{r}}\subseteq {\mathcal{S}}^*_{\lab{r}}$ for the minimization of $\ell^{\tiny{\mathrm{exp}}}_{\lab{r}}({\mathcal{S}}_r,
  \ve{\theta}, \Omega)$ to indeed yield that of $\ell^{\tiny{\mathrm{log}}}_{\lab{e}}\left({\mathcal{S}}_{\lab{e}}, \ve{\theta}, \Omega\right)$. 
\begin{theorem}\label{thm_concentration}
Assume $\Theta \subseteq {\mathcal{B}}_{\|.\|_2}(\ve{0},r_\theta)$, with
$r_\theta > 0$. Let $\varrho(\ve{\theta}) \defeq (\sup_{\ve{\theta}' \in \Theta}
\max_{\ve{\rado}_{\ve{\sigma}} \in {\mathcal{S}}^*_{\lab{r}}} \exp(-\ve{\theta}'^\top
\ve{\rado}_{\ve{\sigma}}))/\ell^{\tiny{\mathrm{exp}}}_{\lab{r}}({\mathcal{S}}^*_r,
  \ve{\theta})$. 
Then if $m$ is sufficiently large, $\forall \updelta > 0$, there is probability $\geq 1 -
\updelta$ over the sampling of ${\mathcal{S}}_r$ that any
$\ve{\theta}\in \Theta$ satisfies:
\begin{eqnarray}
\ell^{\tiny{\mathrm{log}}}_{\lab{e}}\left({\mathcal{S}}_{\lab{e}},
  \ve{\theta}, \Omega\right) \hspace{-0.2cm} & \leq & \hspace{-0.2cm}
\log 2 + (1/m) \cdot \log \ell^{\tiny{\mathrm{exp}}}_{\lab{r}}({\mathcal{S}}_r,
  \ve{\theta}, \Omega)\nonumber\\
 & &  \hspace{-0.2cm} + O\left( \frac{\varrho(\ve{\theta})}{m^\beta}
    \cdot \sqrt{\frac{r_\theta r^\star_\pi}{n} +
  \frac{d}{n m}\log \frac{n}{d \updelta} } \right) \:\:,\nonumber
\end{eqnarray}
as long as $\upomega \geq u m$ for some constant $u>0$.
\end{theorem}

\section{Boosting with (rado) regularized losses}\label{secboo}

\begin{algorithm}[t]
\caption{\radaboostNoL}\label{algoRAdaBoostLGEN}
\begin{algorithmic}
\STATE  \textbf{Input} rados ${\mathcal{S}}_{\lab{r}} \defeq
\{\ve{\rado}_{1},\ve{\rado}_{2}, ...,
\ve{\rado}_{n}\}$; $T\in {\mathbb{N}}_*$; $\upomega \in
{\mathbb{R}}_{+}$; $\upgamma \in (0,1)$;
\STATE  Step 1 : let $\ve{\theta}_0 \leftarrow \ve{0}$, $\ve{w}_0
\leftarrow (1/n)\ve{1}$ ; 
\STATE  Step 2 : \textbf{for} $t = 1, 2, ..., T$
\STATE  \hspace{0.5cm} Step 2.1 : call the weak learner
\begin{mdframed}[style=MyFrame]
\begin{eqnarray}
(\iota(t), r_t) & \leftarrow & 
\omegaweak({\mathcal{S}}_{\lab{r}}, \ve{w}_t,
    \upgamma, \upomega, \ve{\theta}_{t-1})\:\:;
\end{eqnarray} 
\end{mdframed}
\STATE  \hspace{0.5cm} Step 2.2 : let
\begin{eqnarray}
\hspace{-1.15cm}\alpha_{\iota(t)} & \leftarrow & \frac{1}{2\rado_{*\iota(t)}}
\log \frac{1 + r_t}{1 -
  r_t}\:\:;\label{defalpha}
\end{eqnarray}
\begin{mdframed}[style=MyFrame]
\begin{eqnarray}
\delta_{t} & \leftarrow & \upomega\cdot
    (\Omega(\ve{\theta}_{t})-\Omega(\ve{\theta}_{t-1})) \:\:;
\end{eqnarray}
\end{mdframed}
\STATE  \hspace{0.5cm} Step 2.3 : \textbf{for} $j = 1, 2, ..., n$
\begin{mdframed}[style=MyFrame]
\begin{eqnarray}
w_{tj} & \leftarrow & \frac{w_{(t-1)j}}{Z_t} \cdot
    \exp\left(-\alpha_{t} \rado_{j \iota(t)} + \delta_t\right) \:\:;\label{normz}
\end{eqnarray}
\end{mdframed}
\STATE \textbf{Return} $\ve{\theta}_{T}$;
\end{algorithmic}
\end{algorithm}

\begin{algorithm}[t]
\caption{\omegaweak, for $\Omega \in \{\|.\|_1, \|.\|_{\Gamma}^2,
  \|.\|_\infty, \|.\|_\Phi\}$}\label{weakl}
\begin{algorithmic}
\STATE  \textbf{Input} set of rados ${\mathcal{S}}_{\lab{r}} \defeq
\{\ve{\rado}_{1},\ve{\rado}_{2}, ...,
\ve{\rado}_{n}\}$; weights $\ve{w} \in \bigtriangleup_n$; parameters
$\upgamma \in  (0,1)$, $\upomega \in {\mathbb{R}}_+$;
classifier $\ve{\theta} \in {\mathbb{R}}^d$;
\STATE  Step 1 : pick weak feature $\iota_*\in [d]$;
\begin{mdframed}[style=MyFrame]
Optional --- use preference
order:
\begin{eqnarray}
\iota \succeq \iota' & \Leftrightarrow & |r_{\iota}| -  \delta_{\iota}
\geq |r_{\iota'}| - \delta_{\iota'}\:\:;\\
& & (\updelta_\iota \defeq \upomega\cdot(\Omega(\ve{\theta} +
\alpha_{\iota}\cdot \ve{1}_{\iota}) - \Omega(\ve{\theta})))\nonumber
\end{eqnarray}
// $r_{\iota}$ is given in (\ref{defRt}), $\alpha_{\iota}$ is given in
(\ref{defalpha})
\end{mdframed}
\STATE  Step 2 : \textbf{if} $\Omega = \|.\|_\Gamma^2$ \textbf{then} 
\begin{eqnarray}
r_* & \leftarrow & \left\{\begin{array}{ccl}
r_{\iota_*} & \mbox{ if } & r_{\iota_*} \in [-\upgamma, \upgamma]\\
 \mathrm{sign}\left(r_{\iota_*}\right)\cdot \upgamma &
 \multicolumn{2}{l}{\mbox{ otherwise}}
\end{array}\right.\:\:;\label{defmuL2}
\end{eqnarray}
\STATE  \hspace{1.1cm} \textbf{else} $r_* \leftarrow r_{\iota_*}$;
\STATE \textbf{Return} $(\iota_*, r_*)$;
\end{algorithmic}
\end{algorithm}

\radaboostNoL~presents our approach to
learning with rados regularized with regularizer $\Omega$
to minimise loss 
$\ell^{\tiny{\mathrm{exp}}}_{\lab{r}}({\mathcal{S}}_r,
  \ve{\theta}, \Omega)$ in eq. (\ref{defregexp}). Classifier $\ve{\theta}_t$ is
defined as $\ve{\theta}_t \defeq \sum_{t'=1}^t \alpha_{\iota(t')} \cdot \ve{1}_{\iota(t')}$,
where $\ve{1}_k$ is the $k^{th}$ canonical basis vector.
A key
property is that \radaboostNoL~\textit{bypasses} the Minkowski sums to
compute regularized rados. It is therefore computationally
efficient.
Frameboxes highlight the differences with
\radoboost~\citep{npfRO}. The \textit{expected edge} $r_t$ used to
compute $\alpha_t$ in eq. (\ref{defalpha}) is based on the following basis
assignation:
\begin{eqnarray}
r_{\iota(t)} & \leftarrow & \frac{1}{\rado_{*\iota(t)}}
\sum_{j=1}^{n} {w_{tj} \rado_{j \iota(t)}} \:\: (\in [-1,1])\:\:.\label{defRt}
\end{eqnarray}
The computation of $r_t$ is eventually tweaked by the weak learner, as
displayed in Algorithm \omegaweak. We investigate four choices for
$\Omega$. For \textit{each} of them, we prove the boosting ability of \radaboostNoL~($\Upgamma$ is symmetric positive definite, $S_d$ is the
symmetric group of order $d$, $|\ve{\theta}|$ is the vector whose
coordinates are the absolute values of the coordinates of $\ve{\theta}$): 
\begin{eqnarray}
\Omega(\ve{\theta}) &\hspace{-0.2cm} = \hspace{-0.2cm} & \left\{
\begin{array}{lcll}
\|\ve{\theta}\|_1 & \defeq&  |\ve{\theta}|^\top \ve{1}& \mbox{Lasso} \\
\|\ve{\theta}\|^2_\Gamma & \defeq&  \ve{\theta}^\top
\Upgamma\ve{\theta} & \mbox{Ridge} \\
\|\ve{\theta}\|_\infty & \defeq & \max_k |\theta_k|& \mbox{$\ell_\infty$} \\
\|\ve{\theta}\|_\Phi & \defeq  & \max_{\matrice{m} \in S_d}
(\matrice{m}|\ve{\theta}|)^\top \ve{\xi} & \mbox{\slope}
\end{array}
\right. \label{defregs}
\end{eqnarray}
\citep{bjmoOW,bvsscSA,dsEL,scSI}. The
coordinates of $\ve{\xi}$ in \slope~are $\xi_k \defeq \Phi^{-1}(1-kq/(2d))$ where
$\Phi^{-1}(.)$ is the quantile of the standard normal distribution and
$q\in (0,1)$;
thus, the largest coordinates (in absolute value) of $\ve{\theta}$ are more penalized. We now establish the boosting ability of \radaboostNoL. We give no direction for Step 1 in
\omegaweak, which is consistent with the definition of a weak learner
in the boosting theory: 
all we require from the weak learner
 is $|r_{.}|$ no smaller than some weak learning threshold
$\gwl > 0$. 
\begin{definition}
Fix any constant $\gwl\in (0,1)$. \omegaweak~is said
to be a $\gwl$-Weak Learner iff the feature
$\iota(t)$ it picks at iteration $t$ satisfies $|r_{\iota(t)}|\geq
\gwl$, for any $t = 1, 2, ..., T$.
\end{definition}
We also provide an optional step for the weak learner in \omegaweak, which we
exploit in the experimentations, which gives a total preference order on
features to optimise further the convergence of \radaboostNoL. 
\begin{theorem}\label{thadarL2}
(\textbf{boosting with ridge}). Take $\Omega(.) = \|.\|^2_\Gamma$. Fix any $0<a<1/5$, and suppose that $\upomega$ and the number of
iterations $T$ of \radaboostNoL~are chosen so that
\begin{eqnarray}
\upomega & < & (2a\min_{k} \max_j \rado_{jk}^2)/(T \uplambda_{\Gamma})\:\:,\label{bomegaL2}
\end{eqnarray}
where $\uplambda_{\Gamma}>0$ is the largest eigenvalue of $\Gamma$. Then there exists some $\upgamma > 0$ (depending on $a$, and given to \omegaweak) such that for any fixed $0 < \gwl <
\upgamma$, if \omegaweak~is a $\gwl$-Weak Learner, then \radaboostNoL~returns at
the end of the $T$ boosting iterations a classifier $\ve{\theta}_T$
which meets:
\begin{eqnarray}
\ell^{\tiny{\mathrm{exp}}}_{\lab{r}}({\mathcal{S}}_r,
  \ve{\theta}_T, \|.\|_\Gamma^2) & \leq & \exp(-a \gwl^2 T/2)\:\:.
\end{eqnarray}
Furthermore, if we fix $a = 1/7$, then we can fix $\upgamma = 0.98$,
and if we consider $a = 1/10$, then we can fix $\upgamma = 0.999$.
\end{theorem}
(Proof in Appendix, Subsection
\ref{proof_thm_thadarL2}) Two remarks are in order. First, the cases $a=1/7, 1/10$ show that \omegaweak~can still obtain large edges in eq. (\ref{defRt}), so
even a ``strong'' weak learner might fit in for \omegaweak, without clamping 
edges. Second, the right-hand side of ineq. (\ref{bomegaL2}) may be
very large if we consider that $\min_{k} \max_j \rado_{jk}^2$ may be
proportional to $m^2$. So the constraint on $\upomega$ is in fact 
loose, and $\upomega$ may easily meet the constraint of Thm
\ref{thm_concentration}.
\begin{theorem}\label{thadarINF}
(\textbf{boosting with lasso or $\ell_\infty$}). Take $\Omega(.) \in \{
\|.\|_1, \|.\|_\infty\}$. Suppose \omegaweak~is a
$\gwl$-Weak Learner for some $\gwl>0$. Suppose $\exists 0<a<3/11$
s. t.
$\upomega$ satisfies:
\begin{eqnarray}
\upomega & = & a \gwl \min_{k} \max_j |\rado_{jk}|\:\:.\label{bomegaLINF}
\end{eqnarray}
Then \radaboostNoL~returns at
the end of the $T$ boosting iterations a classifier $\ve{\theta}_T$
which meets:
\begin{eqnarray}
\ell^{\tiny{\mathrm{exp}}}_{\lab{r}}({\mathcal{S}}_r,
  \ve{\theta}_T, \Omega) & \leq & \exp(- \tilde{T} \gwl^2/2)\:\:,
\end{eqnarray}
where 
\begin{eqnarray}
\tilde{T} & \defeq & \left\{
\begin{array}{ccl}
a \gwl T & \mbox{ if } & \Omega = \|.\|_1\\
(T - T_*) + a \gwl \cdot T_*& \mbox{ if } & \Omega = \|.\|_\infty
\end{array}\right.\:\:,
\end{eqnarray}
 and $T_*$ is the
number of iterations where the feature computing the $\ell_\infty$ norm
was updated\footnote{If several features match this criterion, $T_*$
  is the total number of iterations for all these features.}.
\end{theorem}
(Proof in Appendix, Subsection
\ref{proof_thm_thadarINF}) We finally investigate the \slope~
choice. The Theorem is proven for $\upomega = 1$ in \radaboostNoL, for
two reasons: it matches the original definition \citep{bvsscSA} and
furthermore it unveils an interesting connection between boosting and
the \slope~ properties \citep{scSI}.
\begin{theorem}\label{thadarSLOPE}
(\textbf{boosting with \slope}). Take $\Omega(.) = \|.\|_\Phi$. Suppose wlog
$|\theta_{Tk}| \geq |\theta_{T(k+1)}|, \forall k$, and fix $\upomega = 1$. Let
\begin{eqnarray}
a & \defeq & \min \left\{\frac{3 \gwl}{11}, \frac{\Phi^{-1}(1-q/(2d))}{
    \min_ k \max_j
  |\rado_{jk}|}\right\}\label{defaaMF}\:\:.
\end{eqnarray}
Suppose (i) \omegaweak~is a
$\gwl$-Weak Learner for some $\gwl>0$, and (ii) the $q$-value is
chosen to meet:
\begin{eqnarray*}
q \hspace{-0.2cm} & \geq & \hspace{-0.2cm} 2\cdot \max_k \left\{\biggl( 1 - \Phi\left(\frac{3\gwl}{11}\cdot \max_j
  |\rado_{jk}|\right) \biggr) \bigg/ \left( \frac{k}{d} \right)\right\}\:\:.
\end{eqnarray*}
Then classifier $\ve{\theta}_T$ returned by \radaboostNoL~at
the end of the $T$ boosting iterations satisfies:
\begin{eqnarray}
\ell^{\tiny{\mathrm{exp}}}_{\lab{r}}({\mathcal{S}}_r,
  \ve{\theta}_T, \|.\|_\Phi) & \leq & \exp(-a \gwl^2 T/2)\:\:.
\end{eqnarray}
\end{theorem}
(Proof in Appendix, Subsection
\ref{proof_thm_thadarSLOPE}) Constraint (ii) on $q$ is interesting in
the light of the properties of
\slope~\citep{bvsscSA,scSI}. Modulo some assumptions, \slope~yields a control the
false discovery rate (FDR) --- that is, sparsity errors, negligible coefficients in the
"true'' linear model $\ve{\theta}^*$ that are actually found
significant in the learned $\ve{\theta}$ ---. Constraint (ii) links
the "small'' achievable FDR (upperbounded by $q$) to the "boostability'' of the data:
the fact that each feature
$k$ can be chosen by the weak learner for a "large'' $\gwl$, or has $\max_j
  |\rado_{jk}|$ large, precisely flags potential significant
  features, thus reducing the risk of sparsity errors, and allowing small $q$, which is constraint
  (ii). Using the second order approximation of normal quantiles
  \citep{scSI}, a sufficient condition for (ii) is that, for some constant $K$,
\begin{eqnarray}
\gwl \min_j \max_j
  |\rado_{jk}| & \geq & K\cdot \sqrt{\log d + \log q^{-1}}\:\:;\label{edq}
\end{eqnarray}
but $\min_j \max_j
  |\rado_{jk}|$ is proportional to $m$, so ineq. (\ref{edq}), and thus
  (ii), may hold
  even for small samples and $q$-values.

We can now have a look at the
regularized log-loss of $\ve{\theta}_T$ over examples, as depicted in Theorem
\ref{thm_concentration}, and show that it is guaranteed a monotonic
decrease with $T$, with high probability, for any applicable choice of
regularization, since we get indeed that the regularized log-loss of
$\ve{\theta}_T$ output by \radaboostNoL, computed on \textit{examples}, satisfies with high probability
$\ell^{\tiny{\mathrm{log}}}_{\lab{e}}\left({\mathcal{S}}_{\lab{e}},
  \ve{\theta}, \Omega\right) \leq \log 2 - \upkappa \cdot T + \uptau(m)$,
with $\uptau(m)\rightarrow 0$ when $m\rightarrow \infty$, and
$\upkappa$ does not depend on $T$. Hence, \radaboostNoL~is an
efficient proxy to boost the regularized log-loss
over examples, using \textit{whichever} of the ridge, lasso, $\ell_\infty$ or
\slope~regularization, establishing the first boosting algorithm for
this last choice. Notice finally that we can also choose any linear
combinations of the regularizers and still keep the formal boosting
property, thereby extending our results \textit{e.g.}, to the popular
elastic nets regularization \citep{zhRA}.

\begin{algorithm}[t]
\caption{\dprad}\label{dpr}
\begin{algorithmic}
\STATE  \textbf{Input} rados ${\mathcal{S}}_{\lab{r}} \defeq
\{\ve{\rado}_{1},\ve{\rado}_{2}, ...,
\ve{\rado}_{n}\}$; budget $\varepsilon
> 0$;
\STATE  Step 1 : let ${\mathcal{S}}^{\textsc{dp}}_{\lab{r}} \leftarrow
\emptyset$;
\STATE  Step 2 : \textbf{for} $j = 1, 2, ..., n$
\STATE  \hspace{0.5cm} Step 2.1 : sample $\ve{z}_j$ as $z_{jk} \sim \mathrm{Lap}(z|n
r_{\lab{e}}/\varepsilon) \:\:, \forall k$;
\STATE  \hspace{0.5cm} Step 2.2 : ${\mathcal{S}}^{\textsc{dp}}_{\lab{r}}   \leftarrow 
{\mathcal{S}}^{\textsc{dp}}_{\lab{r}} \cup \{\ve{\rado}_{j}  + \ve{z}_j\}$;
\STATE \textbf{Return} ${\mathcal{S}}^{\textsc{dp}}_{\lab{r}}$;
\end{algorithmic}
\end{algorithm}

\begin{table*}[t]
\begin{center}
{\tiny
\begin{tabular}{|crr||r||rr|rrr|rrr|rrr|rrr|}
\hline \hline
 & \hspace{-0.2cm}& \hspace{-0.4cm}&   \adaboostSS$\wedge$  &
 \multicolumn{14}{c|}{{\scriptsize \radaboostNoL}} \\ 
 & \hspace{-0.2cm}& \hspace{-0.4cm}& $\ell_1$-\adaboostSS\hspace{-0.19cm}& \multicolumn{2}{c|}{$\upomega = 0$} &  \multicolumn{3}{c|}{$\Omega = \|.\|_{\matrice{\tiny{i}}_d}^2$} & \multicolumn{3}{c|}{$\Omega = \|.\|_1$} &
\multicolumn{3}{c|}{$\Omega = \|.\|_\infty$}&
\multicolumn{3}{c|}{$\Omega = \|.\|_{\Phi}$}\\
domain & \hspace{-0.2cm}$m$ & \hspace{-0.4cm}$d$ &
\multicolumn{1}{c||}{err$\pm\sigma$ \hspace{-0.19cm} } & &
\multicolumn{1}{c|}{err$\pm\sigma$} & & \multicolumn{1}{c}{err$\pm\sigma$}
& \multicolumn{1}{c|}{$\Delta$} & & \multicolumn{1}{c}{err$\pm\sigma$}
& \multicolumn{1}{c|}{$\Delta$} & & \multicolumn{1}{c}{err$\pm\sigma$}
& \multicolumn{1}{c|}{$\Delta$} & & \multicolumn{1}{c}{err$\pm\sigma$}
& \multicolumn{1}{c|}{$\Delta$} \\ \hline
\hspace{-0.27cm} Fertility & \hspace{-0.49cm} 100 & \hspace{-0.4cm} 9 & $\diamond$ 40.00$\pm$14.1
\hspace{-0.19cm} & & 40.00$\pm$14.9 &
& 41.00$\pm$16.6 & 8.00 & \hspace{-0.2cm}$\bullet$\hspace{-0.2cm} &
41.00$\pm$14.5 & 4.00 & \hspace{-0.2cm}$\circ$\hspace{-0.2cm} & 41.00$\pm$21.3 &
6.00 & & \cellcolor{Gray}38.00$\pm$14.0 & 7.00\\
\hspace{-0.27cm} Sonar & \hspace{-0.49cm} 208 & \hspace{-0.4cm} 60 & \mybs$\diamond$ 24.57$\pm$9.11
\hspace{-0.19cm} & \hspace{-0.2cm}$\bullet$\hspace{-0.2cm} & 27.88$\pm$4.33 & & 25.05$\pm$7.56 &
8.14 & & \cellcolor{Gray}24.05$\pm$8.41 & 4.83 & & 24.52$\pm$8.65 & 10.12 &\hspace{-0.2cm}$\circ$\hspace{-0.2cm} 
& 25.00$\pm$13.4 & 3.83\\
\hspace{-0.27cm} Haberman & \hspace{-0.49cm} 306 & \hspace{-0.4cm} 3 &  $\blacksquare$$\diamond$ 25.15$\pm$6.53
\hspace{-0.19cm} & \hspace{-0.2cm}$\bullet$\hspace{-0.2cm} & 25.78$\pm$7.18 &
\hspace{-0.2cm}$\bullet$\hspace{-0.2cm} & \cellcolor{Gray}24.83$\pm$6.18
& 1.62 & & 25.80$\pm$6.71 & 1.32 & & 25.48$\pm$7.37 & 1.62 &\hspace{-0.2cm}$\circ$\hspace{-0.2cm} 
& 25.78$\pm$7.18 & 1.65\\
\hspace{-0.27cm} Ionosphere & \hspace{-0.49cm} 351 & \hspace{-0.4cm} 33 &  $\blacksquare$$\diamond$ 13.11$\pm$6.36
\hspace{-0.19cm} & \hspace{-0.2cm}$\bullet$\hspace{-0.2cm} & 14.51$\pm$7.36 &
& 13.64$\pm$5.99 & 5.43 & & 14.24$\pm$6.15 & 2.83 & \hspace{-0.2cm}$\circ$\hspace{-0.2cm} & \cellcolor{Gray}13.38$\pm$4.44 & 3.15
& \hspace{-0.2cm}$\circ$\hspace{-0.2cm} & 14.25$\pm$5.04 & 3.41\\
\hspace{-0.27cm} Breastwisc & \hspace{-0.49cm} 699 & \hspace{-0.4cm} 9 &  $\blacksquare$$\diamond$  3.00$\pm$1.96
\hspace{-0.19cm} & & 3.43$\pm$2.25 & & \cellcolor{Gray}2.57$\pm$1.62 &
1.14 & \hspace{-0.2cm}$\circ$\hspace{-0.2cm} & 3.29$\pm$2.24 & 0.86 & & 2.86$\pm$2.13 & 0.86 & \hspace{-0.2cm}$\bullet$\hspace{-0.2cm} & 3.00$\pm$2.18 &
0.29\\
\hspace{-0.27cm} Transfusion & \hspace{-0.49cm} 748 & \hspace{-0.4cm} 4 &  $\blacksquare$$\diamond$  39.17$\pm$7.01
\hspace{-0.19cm} & \hspace{-0.2cm}$\circ$\hspace{-0.2cm} & 37.97$\pm$7.42 &
& 37.57$\pm$5.60 & 2.40 & \hspace{-0.2cm}$\circ$\hspace{-0.2cm} & 36.50$\pm$6.78 & 2.14 & \hspace{-0.2cm}$\circ$\hspace{-0.2cm} & 37.43$\pm$8.08 & 1.21
& \hspace{-0.2cm}$\bullet$\hspace{-0.2cm} & \cellcolor{Gray}36.10$\pm$8.06 & 3.21\\
\hspace{-0.27cm} Banknote & \hspace{-0.49cm} 1 372 & \hspace{-0.4cm} 4 &  $\blacksquare$$\diamond$ 2.70$\pm$1.46
\hspace{-0.19cm} &\hspace{-0.2cm}$\circ$\hspace{-0.2cm} 
& 14.00$\pm$4.16 & \hspace{-0.2cm}$\bullet$\hspace{-0.2cm} & \cellcolor{Gray}12.02$\pm$2.74
& 0.73 & & 13.63$\pm$2.75 & 1.39 &
\hspace{-0.2cm}$\bullet$\hspace{-0.2cm} & 12.17$\pm$2.77 & 0.80 &
& 13.63$\pm$3.02 & 1.39\\
\hspace{-0.27cm} Winered & \hspace{-0.49cm} 1 599 & \hspace{-0.4cm} 11 &  $\blacksquare$$\diamond$ 26.33$\pm$2.75
\hspace{-0.19cm} &\hspace{-0.2cm}$\circ$\hspace{-0.2cm} 
& 28.02$\pm$3.32 & \hspace{-0.2cm}$\bullet$\hspace{-0.2cm} & 27.83$\pm$3.95
& 1.19 & \hspace{-0.2cm}$\bullet$\hspace{-0.2cm} & \cellcolor{Gray}27.45$\pm$4.17 & 1.00 & \hspace{-0.2cm}$\circ$\hspace{-0.2cm} & 27.58$\pm$3.76 & 1.12 &
\hspace{-0.2cm}$\bullet$\hspace{-0.2cm} & \cellcolor{Gray}27.45$\pm$3.34 & 1.25\\
\hspace{-0.27cm} Abalone & \hspace{-0.49cm} 4 177 & \hspace{-0.4cm} 10 &  $\blacksquare$ 22.98$\pm$2.70
\hspace{-0.19cm} &
\hspace{-0.2cm}$\bullet$\hspace{-0.2cm} & 26.57$\pm$2.31 & \hspace{-0.2cm}$\circ$\hspace{-0.2cm} & 24.18$\pm$2.51
& 0.00 & & 24.13$\pm$2.48 & 0.14 & \hspace{-0.2cm}$\circ$\hspace{-0.2cm} & 24.18$\pm$2.51 & 0.00 &
& \cellcolor{Gray}24.11$\pm$2.59 & 0.07\\
\hspace{-0.27cm} Winewhite & \hspace{-0.49cm} 4 898 & \hspace{-0.4cm} 11 &  $\blacksquare$$\diamond$
 30.73$\pm$2.20
\hspace{-0.19cm} 
& \hspace{-0.2cm}$\bullet$\hspace{-0.2cm} & 32.63$\pm$2.52 &
\hspace{-0.2cm}$\bullet$\hspace{-0.2cm} & \cellcolor{Gray}31.85$\pm$1.66 & 1.18 & & 32.16$\pm$1.73 & 1.31 & & 32.16$\pm$2.02 & 0.90
& \hspace{-0.2cm}$\circ$\hspace{-0.2cm} & 31.97$\pm$2.26 & 1.12\\
\hspace{-0.27cm} Smartphone & \hspace{-0.49cm} 7 352 &  \hspace{-0.4cm} 561 &
 0.00$\pm$0.00 \hspace{-0.19cm} & \hspace{-0.2cm}$\circ$\hspace{-0.2cm} & 0.67$\pm$0.25 & &
\cellcolor{Gray} 0.19$\pm$0.22 & 0.00 &
\hspace{-0.2cm}$\circ$\hspace{-0.2cm} & 0.44$\pm$0.29 & 0.03 &
\hspace{-0.2cm}$\bullet$\hspace{-0.2cm} & 0.20$\pm$0.24 & 0.01
& & \cellcolor{Gray} 0.19$\pm$0.22 & 0.04\\
\hspace{-0.27cm} Firmteacher & \hspace{-0.49cm} 10 800 &  \hspace{-0.4cm} 16 & $\blacksquare$$\diamond$
44.44$\pm$1.34 \hspace{-0.19cm} & & 40.58$\pm$4.87 & \hspace{-0.2cm}$\bullet$\hspace{-0.2cm} &
40.89$\pm$3.95 & 2.35 & & 39.81$\pm$4.37 & 2.89 & \hspace{-0.2cm}$\circ$\hspace{-0.2cm}  & 38.91$\pm$4.51 &
3.56 &\hspace{-0.2cm}$\circ$\hspace{-0.2cm}  & \cellcolor{Gray}38.01$\pm$6.15 & 5.02\\
\hspace{-0.27cm} Eeg & \hspace{-0.49cm} 14 980 & \hspace{-0.4cm} 14 & $\diamond$
45.38$\pm$2.04 \hspace{-0.19cm} &
\hspace{-0.2cm}$\bullet$\hspace{-0.2cm} & 44.09$\pm$2.32 & \hspace{-0.2cm}$\circ$\hspace{-0.2cm} & 44.01$\pm$1.48 &
0.40 & \hspace{-0.2cm}$\bullet$\hspace{-0.2cm} & 43.89$\pm$2.19 & 0.89
& \hspace{-0.2cm}$\circ$\hspace{-0.2cm} & 44.07$\pm$2.02 & 0.81 & \hspace{-0.2cm}$\bullet$\hspace{-0.2cm} & \cellcolor{Gray}43.87$\pm$1.40
& 0.95\\
\hspace{-0.27cm} Magic & \hspace{-0.49cm} 19 020 & \hspace{-0.4cm} 10 &  21.07$\pm$1.09
\hspace{-0.19cm} &
\hspace{-0.2cm}$\circ$\hspace{-0.2cm} & 37.51$\pm$0.46 & \hspace{-0.2cm}$\bullet$\hspace{-0.2cm} &
\cellcolor{Gray}22.11$\pm$1.32 & 0.28 & \hspace{-0.2cm}$\circ$\hspace{-0.2cm} & 26.41$\pm$1.08 & 0.00 & &
23.00$\pm$1.71 & 0.66 & \hspace{-0.2cm}$\circ$\hspace{-0.2cm} &
26.41$\pm$1.08 & 0.00\\
\hspace{-0.27cm} Hardware & \hspace{-0.49cm}  28 179 & \hspace{-0.4cm}  96 &
 16.77$\pm$0.73 \hspace{-0.19cm} & \hspace{-0.2cm}$\circ$\hspace{-0.2cm} & 9.41$\pm$0.71 & &
6.43$\pm$0.74 & 0.18 & \hspace{-0.2cm}$\circ$\hspace{-0.2cm} & 11.72$\pm$1.24 & 0.41 & \hspace{-0.2cm}$\bullet$\hspace{-0.2cm} & 6.50$\pm$0.67 &
0.10 & & \cellcolor{Gray}6.42$\pm$0.69 & 0.13\\
\Xhline{2\arrayrulewidth}
\hspace{-0.27cm} Marketing & \hspace{-0.49cm} 45 211 & \hspace{-0.4cm} 27 & $\blacksquare$ 30.68$\pm$1.01 \hspace{-0.19cm} & & 27.70$\pm$0.69 & &
27.33$\pm$0.73 & 0.33 &  \hspace{-0.2cm}$\circ$\hspace{-0.2cm} &
28.02$\pm$0.47 & 0.00 & \hspace{-0.2cm}$\bullet$\hspace{-0.2cm}  & \cellcolor{Gray}27.19$\pm$0.87 &
0.51 & \hspace{-0.2cm}$\circ$\hspace{-0.2cm}  & 28.02$\pm$0.47 & 0.00\\
\hspace{-0.27cm} Kaggle & \hspace{-0.49cm} 120 269 & \hspace{-0.4cm} 11 & $\blacksquare$ 47.80$\pm$0.47 \hspace{-0.19cm} & \hspace{-0.2cm}$\bullet$\hspace{-0.2cm} & 39.22$\pm$8.47 & \hspace{-0.2cm}$\circ$\hspace{-0.2cm} &
16.90$\pm$0.51 & 0.00 & & 16.90$\pm$0.51 & 0.00 & & \cellcolor{Gray}16.89$\pm$0.50 &
0.01 & & 16.90$\pm$0.51 & 0.00\\
\hline\hline
\end{tabular}
}
\end{center}
\caption{Best result of \adaboostSS/$\ell_1$-\adaboostSS~\citep{ssIBj,xxrsSA}, vs \radaboostNoL~(with or without
  regularization, trained with $n=m$ random rados (above bold horizontal
  line) / $n=10 000$ rados (below bold horizontal line)), according to the
  expected true error. Table shows the best result
  over all $\upomega$s, as well as 
  the difference between the worst and best
  ($\Delta$). Shaded cells display the best result of
  \radaboostNoL. For each domain, the sparsest of \radaboostNoL's method (in
  average) is indicated with "$\circ$", and the least sparse is
  indicated with "$\bullet$". When
  \adaboostSS~(resp. $\ell_1$-\adaboostSS) yields the least sparse
  (resp. the sparsest) of
  \textit{all} methods (including \radaboostNoL), it is shown with
  "$\blacksquare$" (resp. "$\diamond$"). All domains but Kaggle are
  UCI \citep{blUR}.}
  \label{tc1_errs_rr}
\end{table*}

\section{Regularized losses and differential privacy}\label{secdp}

\begin{sidewaystable}[t]
\begin{center}
{\tiny
\begin{tabular}{ccc}\hline\hline
\includegraphics[width=0.3\columnwidth]{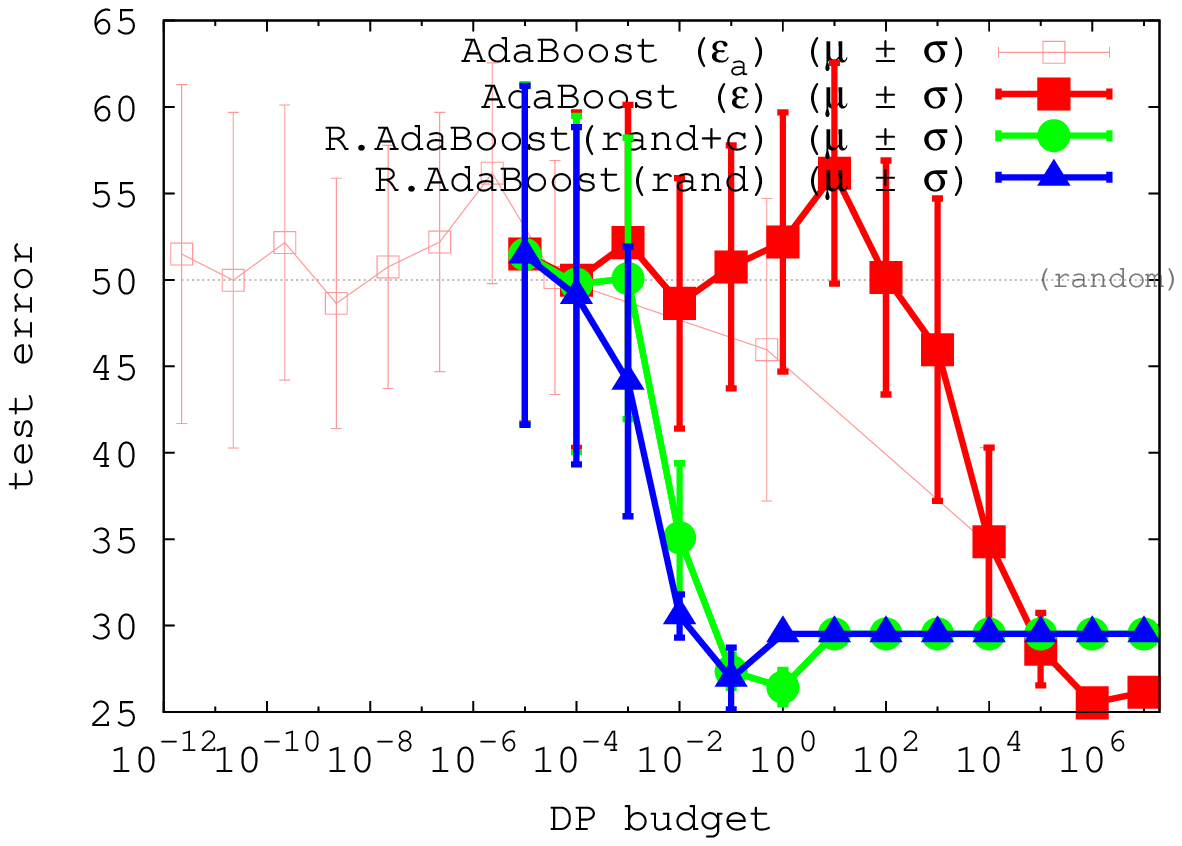}
& \includegraphics[width=0.3\columnwidth]{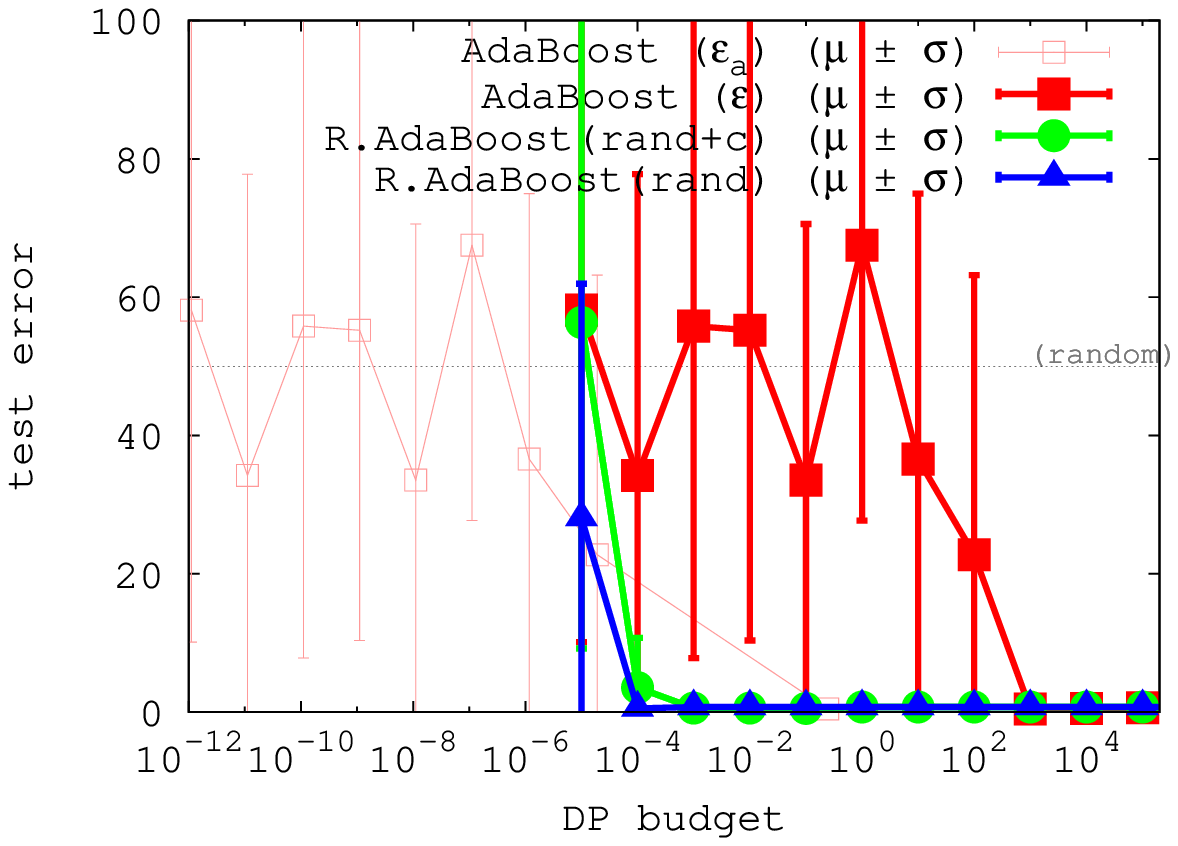}
& \includegraphics[width=0.3\columnwidth]{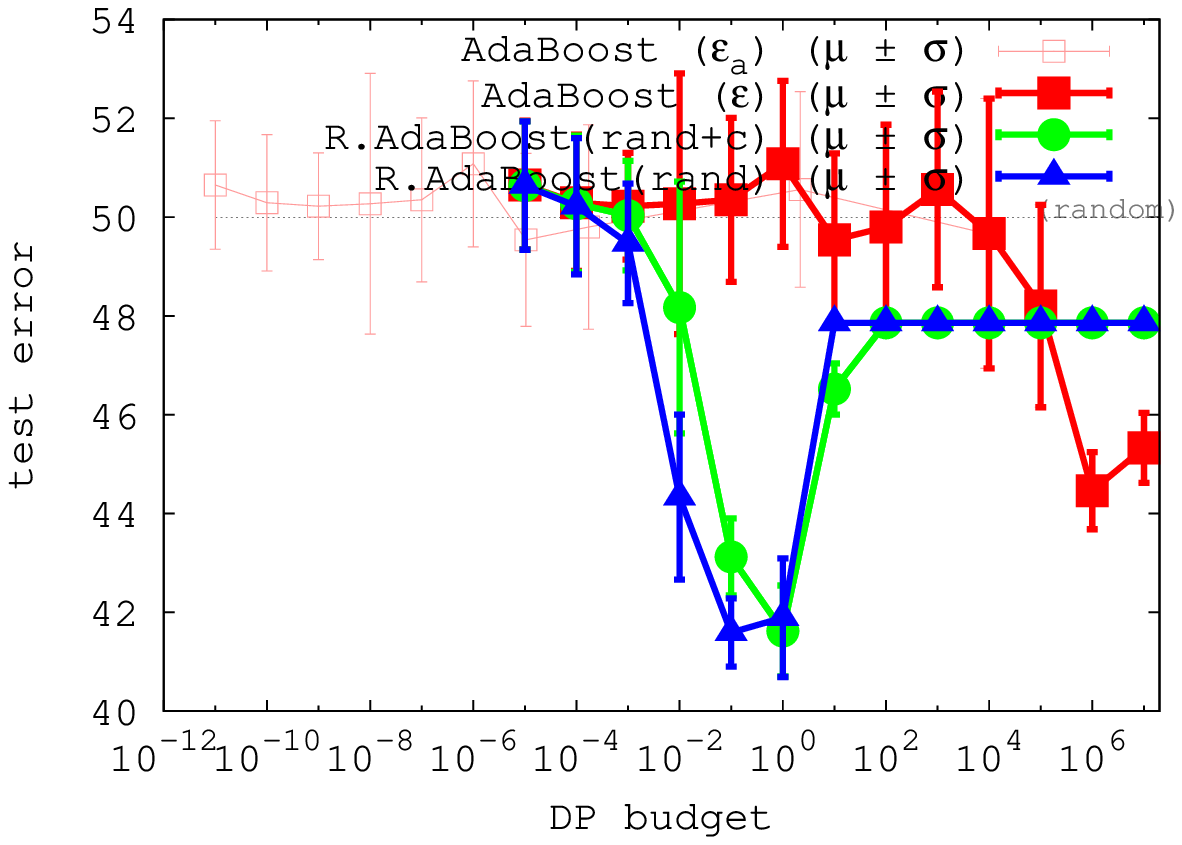}\\
SuSy ($5\times 10^6$) & GaussNLin ($10^7$)  & Higgs ($1.1\times 10^7)$\\\hline\hline
\end{tabular}
}
\end{center}
\caption{DP experiments on big domains ($\upomega = 0$ in \radaboostNoL). Test error as a function of
  privacy budgets $\varepsilon$ and $\varepsilon_a$ for \adaboostSS, and as a
  function of $\varepsilon$ for 
  \radaboostNoL~trained with plain random rados (rand) or class-wise random rados (rand+c).}\label{dp1_complete}
\end{sidewaystable}

We show here that the standard differential privacy (DP) mechanism \citep{drTA} to protect
examples in rados --- not investigated in \citep{npfRO} ---, amounts to
a surrogate form of randomized regularization over \textit{clean}
examples. We let $\mathrm{Lap}(z|b) \defeq
(1/2b)\exp(-|z|/b)$ denote the pdf of 
the Laplace distribution. Algorithm \dprad~states the protection
mechanism. Let us
define two training samples ${\mathcal{S}}_{\lab{e}}$ and
${\mathcal{S}}'_{\lab{e}}$ as being neighbours, noted
${\mathcal{S}}_{\lab{e}} \approx {\mathcal{S}}'_{\lab{e}}$, iff they differ from one
example. We show how the Laplace
mechanism of \dprad~can give $\varepsilon$-DP, and
furthermore the minimisation of a rado-loss over protected rados
resembles the minimisation of an optimistic bound on a 
regularization of the equivalent example loss over clean
examples. We make the assumption that any two edge vectors $\ve{e}, \ve{e}'$ satisfy $\|\ve{e}
-\ve{e}'\|_1 \leq r_{\lab{e}}$, which is ensured \textit{e.g.} if all examples
belong to a $\ell_1$-ball of diameter $r_{\lab{e}}$.

\begin{theorem}\label{thdp1}
\dprad~delivers $\varepsilon$-differential
privacy. Furthermore, pick $(\Omega, \Omega^\star)$ any couple of dual
norms and assume ${\mathcal{S}}_{\lab{r}} = {\mathcal{S}}^*_{\lab{r}}$ ($|{\mathcal{S}}_{\lab{r}}|=2^m$).
Then $\forall \ve{\theta}$, $\ell^{\tiny{\mathrm{exp}}}_{\lab{r}}({\mathcal{S}}^{*,\textsc{dp}}_{\lab{r}},
  \ve{\theta}) \leq \exp \{m \cdot \ell^{\tiny{\mathrm{log}}}_{\lab{e}}\left({\mathcal{S}}_{\lab{e}},
   \ve{\theta}, (1/m)\cdot \max_{j} \Omega^\star(\ve{z}_{j})\cdot\Omega\right)\}$.
\end{theorem}
(Proof in Appendix, Subsection
\ref{proof_thm_thdp1}) 

\section{Experiments}\label{secexp}

We have implemented \omegaweak~using the order suggested 
to retrieve the topmost feature in the order. Hence, the weak learner
returns the feature maximising $|r_\iota|-\delta_\iota$. The rationale
for this comes from the proofs of Theorems \ref{thadarL2} --- \ref{thadarSLOPE}, showing
that $\prod_t \exp(-(r_{\iota(t)}^2/2 - \delta_{\iota(t)}))$ is an
upperbound on the exponential regularized rado-loss. We do not clamp the weak learner for $\Omega(.) =
\|.\|_\Gamma^2$, so the weak learner is restricted to the framebox in \omegaweak\footnote{the
values for $\upomega$ that we test, in $\{10^{-u}, u \in \{0, 1, 2, 3,
4, 5\}\}$, are very small with respect to the upperbound in
ineq. (\ref{bomegaL2}) given the number of boosting iterations ($T=1000$), and would yield on most domains a maximal
$\upgamma \approx 1$.}. We have tested two types of random rados, the
plain random rados \citep{npfRO}, and \textit{class-wise} rados, for
which we first pick 
a class at random and then sample a subset of its
examples to compute one rado (and repeat for $n$
rados). The
supplementary information (Appendix, Section
\ref{exp_expes}) provides the complete experiments, whose
summary is given here.

\paragraph{Experiments I: (regularized) rados vs examples} The
objective of these experiments is to evaluate \radaboostNoL~as a contender
for supervised
learning \textit{per se}. We compared
\radaboostNoL~to
\adaboostSS/$\ell_1$-\adaboostSS~\citep{ssIBj,xxrsSA}. 
All algorithms are run for a total of $T=1000$ iterations, and at the
end of the iterations, the classifier in the sequence that minimizes
the empirical loss is kept. Notice therefore that rado-based
classifiers are evaluated on the training set which computes the rados
(in a privacy setting, the learner send the sequence of classifiers to the
data handler, which then selects the best according to its training
sample). To obtain very
sparse solutions for $\ell_1$-\adaboostSS, we pick its $\upomega$
($\beta$ in \citep{xxrsSA}) in $\{10^{-4},1,10^4\}$. The results we
give, in Table
\ref{tc1_errs_rr}, report only the lowest error of all of
\adaboostSS~variants. The Appendix (Subsection
\ref{exp_compl_tc1_errs_rr}) details the support results, that are
summarized in Table \ref{tc1_errs_rr}. Experiments support several
key observations. First, regularizing consistently reduces the test error of
\radaboostNoL, by more than $15\%$ on Magic, and $20\%$ on Kaggle. Second, \radaboostNoL~is able to obtain both
very sparse \textit{and} accurate classifiers (Magic, Hardware,
Marketing, Kaggle). Third, with
the sole exception of domain Banknote, \radaboostNoL~competes or beats
\adaboostSS~on \textit{all} domains, and is all the better as the domain gets
bigger. Fourth, it is important to have several choices of
regularizers at hand. Fifth, as already remarked \citep{npfRO},
significantly subsampling rados (\textit{e.g.} Marketing, Kaggle) still
yields very accurate classifiers. Finally, regularization in \radaboostNoL~successfully \textit{reduces} 
sparsity to learn more accurate classifiers on
several domains (Transfusion, Banknote, Winered, Magic,
Marketing), achieving efficient \textit{adaptive} sparsity control. 

\paragraph{Experiments II: differential privacy} 

We have tested \dprad~for a fixed
number of rados of $n=100$. Such a small number of rados has three
advantages: (i) the privacy budget does not blow up, (ii) accurate
classifiers can still be learned with a small number of rados
\citep{npfRO}, (iii) with such a small
number of rados, we are within the reach of additional privacy
guarantees \citep{npfRO}. We have 
compared with \adaboostSS, trained over a subset of $n=100$
(protected) examples, randomly sampled out of the full training fold. To make sure that
this does not impair the algorithm just because the sample is too
small, we compute the test error for very large values of
$\varepsilon$ as a baseline. Last,
for tight comparisons, we
use the same set of random vectors $\ve{z}$ to protect the rados \textit{and}
the examples. This choice is justified and discussed at the end of the proof of Theorem
\ref{thdp1} (Appendix, Subsection
\ref{proof_thm_thdp1}). Yet, as we shall see, the results are
exceedingly in favor of \radaboostNoL~in this case. To give a more
balanced picture, we chose to compute an ``approximate'' example-equivalent privacy budget
$\varepsilon_a = \varepsilon_a(\varepsilon, n, m)$ for \adaboostSS~and $n$ examples, which we fix to be
\begin{eqnarray}
\varepsilon_a & \defeq & n \cdot \ln\left(1+\frac{\exp(\varepsilon/n)-1}{m}\right)\label{defepse}\:\:.
\end{eqnarray}
We always have $\varepsilon_a < \varepsilon$. The ``optimal'' DP picture
of \adaboostSS~shall thus be representable as a stretching of its curves in
between the figures for $\varepsilon_a$ and $\varepsilon$. We insist
on the fact that the noise for $\varepsilon$ is conservative but
always safe, while computing $\varepsilon$ from $\varepsilon_a$ would
sometimes fail to provide $\varepsilon_a$-DP (Appendix, Subsection
\ref{proof_thm_thdp1}).

Table \ref{dp1_complete} presents the results obtained for three big domains
($m$ indicated in parenthesis), in which we have run unregularized
algorithms for a fixed number of $T=1000$ iterations, keeping the last
classifier $\ve{\theta}_{1000}$ for testing. GaussNLin is a $d=2$ simulated domain, non linearly separable
but for which the optimal linear classifier has error $<2\%$.
The results are a clear advocacy in favor of using rados
against examples for the straight DP protection:
with plain random rados, test errors that \textit{compete with clean
  data} can be
observed for privacy budget $\varepsilon \approx 10^{-4}$, that is,
more than a hundred times smaller than most reported studies
\citep{hghknprDP}. In comparison, \adaboostSS's results,
even plotted against the weak protection budget $\varepsilon_a$, are
very significantly worse. Finally, on
UCI domains SuSy and Higgs, non-trivial protections (typically,
$\varepsilon \in [0.01, 1]$) allow to \textit{beat} classification on
clean data, as witnessed by a $6\%+$ test error reduction for
Higgs. In addition to ``coming for free'' \citep{wfsPF} in machine
learning, DP may thus also be a worthwhile companion to improve learning.

\section{Conclusion}\label{seccon}

We have shown that the equivalence between the log loss over examples and the
exponential loss over rados, as shown in \citep{npfRO}, can be
generalized to other losses via a principled representation of a loss
function in a two-player zero-sum game. Furthermore, we have shown that this
equivalence extends to regularized losses, where the regularization in
the rado loss is performed over the rados themselves with Minkowski
sums. Because regularization with rados has such a simple form, it is
relatively easy to derive efficient learning algorithms working with various forms of
regularization, as exemplified with ridge, lasso, $\ell_\infty$ and
\slope~regularizations in a formal boosting algorithm that we introduce,
\radaboostNoL. Experiments confirm that this freedom in the choice
of regularization is a clear strength of the
algorithm, and that regularization dramatically improves the
performances over non-regularized rado learning. \radaboostNoL~efficiently \textit{controls}
sparsity, and may be a worthy contender
for supervised learning at large outside the privacy framework.
Experiments also display that
\slope~regularization tends to achieve top performances, and call for an
extension to rados of the formal sparsity
results already known \citep{scSI}.

\section{Acknowledgments}

Thanks are also due to Stephen
Hardy and Giorgio Patrini for many stimulating
discussions and feedback on the subject. NICTA is funded by the Australian Government through the Department of Communications and the Australian Research Council through the ICT Center of Excellence Program.

\bibliography{bibgen}

\begin{thebibliography}{23}
\providecommand{\natexlab}[1]{#1}
\providecommand{\url}[1]{\texttt{#1}}
\expandafter\ifx\csname urlstyle\endcsname\relax
  \providecommand{\doi}[1]{doi: #1}\else
  \providecommand{\doi}{doi: \begingroup \urlstyle{rm}\Url}\fi

\bibitem[Bach et~al.(2011)Bach, Jenatton, Mairal, and Obozinski]{bjmoOW}
Bach, F., Jenatton, R., Mairal, J., and Obozinski, G.
\newblock Optimization with sparsity-inducing penalties.
\newblock \emph{Foundations and Trends in Machine Learning}, 4:\penalty0
  1--106, 2011.

\bibitem[Bache \& Lichman(2013)Bache and Lichman]{blUR}
Bache, K. and Lichman, M.
\newblock {UCI} machine learning repository, 2013.

\bibitem[Bogdan et~al.(2015)Bogdan, {van den Berg}, Sabatti, Su, and
  Cand\`es]{bvsscSA}
Bogdan, M, {van den Berg}, E., Sabatti, C., Su, W., and Cand\`es, E.-J.
\newblock {SLOPE} -- adaptive variable selection via convex optimization.
\newblock \emph{Annals of Applied Statistics}, 2015.
\newblock Also arXiv:1310.1969v2.

\bibitem[Duchi \& Singer(2009)Duchi and Singer]{dsEL}
Duchi, J.-C. and Singer, Y.
\newblock Efficient learning using forward-backward splitting.
\newblock In \emph{NIPS*22}, pp.\  495--503, 2009.

\bibitem[Dwork \& Roth(2014)Dwork and Roth]{drTA}
Dwork, C. and Roth, A.
\newblock The algorithmic foudations of differential privacy.
\newblock \emph{Foundations and Trends in Theoretical Computer Science},
  9:\penalty0 211--407, 2014.

\bibitem[Enserink \& Chin(2015)Enserink and Chin]{ecTE}
Enserink, M. and Chin, G.
\newblock The end of privacy.
\newblock \emph{Science}, 347:\penalty0 490--491, 2015.

\bibitem[Gentile \& Warmuth(1998)Gentile and Warmuth]{gwLH}
Gentile, C. and Warmuth, M.
\newblock Linear hinge loss and average margin.
\newblock In \emph{NIPS*11}, pp.\  225--231, 1998.

\bibitem[Hsu et~al.(2014)Hsu, Gaboardi, Haeberlen, Khanna, Narayan, Pierce, and
  Roth]{hghknprDP}
Hsu, J., Gaboardi, M., Haeberlen, A., Khanna, S., Narayan, A., Pierce, B.-C.,
  and Roth, A.
\newblock Differential privacy: An economic method for choosing epsilon.
\newblock In \emph{Proc.\ of the 27$^{th}$ IEEE CSFS}, pp.\  398--410, 2014.

\bibitem[Kearns \& Mansour(1999)Kearns and Mansour]{kmOTj}
Kearns, M.J. and Mansour, Y.
\newblock On the boosting ability of top-down decision tree learning
  algorithms.
\newblock \emph{J. Comp. Syst. Sc.}, 58:\penalty0 109--128, 1999.

\bibitem[Montanari(2014)]{mCI}
Montanari, A.
\newblock Computational implications of reducing data to sufficient statistics.
\newblock Technical Report 2014-12, Stanford U., 2014.

\bibitem[Nair \& Hinton(2010)Nair and Hinton]{nhRL}
Nair, V. and Hinton, G.
\newblock Rectified linear units improve restricted boltzmann machines.
\newblock In \emph{27$^{th}$ ICML}, pp.\  807--814, 2010.

\bibitem[Nock \& Nielsen(2008)Nock and Nielsen]{nnOT}
Nock, R. and Nielsen, F.
\newblock On the efficient minimization of classification-calibrated
  surrogates.
\newblock In \emph{NIPS*21}, pp.\  1201--1208, 2008.

\bibitem[Nock et~al.(2015)Nock, Patrini, and Friedman]{npfRO}
Nock, R., Patrini, G., and Friedman, A.
\newblock Rademacher observations, private data, and boosting.
\newblock In \emph{32$^{nd}$ ICML}, pp.\  948--956, 2015.

\bibitem[Reid et~al.(2015)Reid, Frongillo, Williamson, and Mehta]{rfwmGM}
Reid, M.-D., Frongillo, R.-M., Williamson, R.-C., and Mehta, N.-A.
\newblock Generalized mixability via entropic duality.
\newblock In \emph{28$^{~th}$ COLT}, pp.\  1501--1522, 2015.

\bibitem[Schapire(2003)]{sTB}
Schapire, R.-E.
\newblock The boosting approach to machine learning: An overview.
\newblock In Denison, D.-D., Hansen, M.-H., Holmes, C.-C., Mallick, B., and Yu,
  B. (eds.), \emph{Nonlinear Estimation and Classification}, volume 171 of
  \emph{Lecture Notes in Statistics}, pp.\  149--171. Springer Verlag, 2003.

\bibitem[Schapire \& Singer(1999)Schapire and Singer]{ssIBj}
Schapire, R.~E. and Singer, Y.
\newblock Improved boosting algorithms using confidence-rated predictions.
\newblock \emph{MLJ}, 37:\penalty0 297--336, 1999.

\bibitem[Su \& Cand\`es(2015)Su and Cand\`es]{scSI}
Su, W. and Cand\`es, E.-J.
\newblock {SLOPE} is adaptive to unkown sparsity and asymptotically minimax.
\newblock \emph{CoRR}, abs/1503.08393, 2015.

\bibitem[Telgarsky(2012)]{tAP}
Telgarsky, M.
\newblock A primal-dual convergence analysis of boosting.
\newblock \emph{JMLR}, 13:\penalty0 561--606, 2012.

\bibitem[{van Rooyen} et~al.(2015){van Rooyen}, Menon, and Williamson]{vmwLW}
{van Rooyen}, B., Menon, A., and Williamson, R.-C.
\newblock Learning with symmetric label noise: The importance of being
  unhinged.
\newblock In \emph{NIPS*28}, 2015.

\bibitem[Vapnik(1998)]{vSL}
Vapnik, V.
\newblock \emph{Statistical Learning Theory}.
\newblock John Wiley, 1998.

\bibitem[Wang et~al.(2015)Wang, Fienberg, and Smola]{wfsPF}
Wang, Y.-X., Fienberg, S.E., and Smola, A.-J.
\newblock Privacy for free: Posterior sampling and stochastic gradient {Monte
  Carlo}.
\newblock In \emph{32$^{nd}$ ICML}, pp.\  2493--2502, 2015.

\bibitem[Xi et~al.(2009)Xi, Xiang, Ramadge, and Schapire]{xxrsSA}
Xi, Y.-T., Xiang, Z.-J., Ramadge, P.-J., and Schapire, R.-E.
\newblock Speed and sparsity of regularized boosting.
\newblock In \emph{12$^{th}$ AISTATS}, pp.\  615--622, 2009.

\bibitem[Zou \& Hastie(2005)Zou and Hastie]{zhRA}
Zou, H. and Hastie, T.
\newblock Regularization and variable selection via the elastic net.
\newblock \emph{Journal of the Royal Statistical Society B}, 67:\penalty0
  301--321, 2005.

\end{thebibliography}
\bibliographystyle{icml2015}

\newpage

\section*{Appendix --- Table of contents}\label{secappendix}

\noindent \textbf{Proofs} \hrulefill Pg \pageref{proof_proofs}\\
\noindent Proof of Theorem \ref{thmstrongrc}\hrulefill Pg \pageref{proof_thm_thmstrongrc}\\
\noindent Proof of Lemma \ref{lemgens}\hrulefill Pg \pageref{proof_lem_lemgens}\\
\noindent Proof of Lemma \ref{srclog}\hrulefill Pg \pageref{proof_lem_srclog}\\
\noindent Proof of Corollary \ref{corlog}\hrulefill Pg \pageref{proof_cor_corlog}\\
\noindent Proof of Lemma \ref{srcsql}\hrulefill Pg \pageref{proof_lem_srcsql}\\
\noindent Proof of Corollary \ref{corsql}\hrulefill Pg \pageref{proof_cor_corsql}\\
\noindent Proof of Lemma \ref{srcrelu}\hrulefill Pg \pageref{proof_lem_srcrelu}\\
\noindent Proof of Corollary \ref{correlu}\hrulefill Pg \pageref{proof_cor_correlu}\\
\noindent Proof of Lemma \ref{srcunh}\hrulefill Pg \pageref{proof_lem_srcunh}\\
\noindent Proof of Corollary \ref{corunh}\hrulefill Pg \pageref{proof_cor_corunh}\\
\noindent Proof of Theorem \ref{regrado}\hrulefill Pg \pageref{proof_thm_regrado}\\
\noindent Proof of Theorem \ref{thadarL2}\hrulefill Pg \pageref{proof_thm_thadarL2}\\
\noindent Proof of Theorem \ref{thadarINF}\hrulefill Pg \pageref{proof_thm_thadarINF}\\
\noindent Proof of Theorem \ref{thadarSLOPE}\hrulefill Pg \pageref{proof_thm_thadarSLOPE}\\
\noindent Proof of Theorem \ref{thdp1}\hrulefill Pg \pageref{proof_thm_thdp1}\\

\noindent \textbf{Additional Experiments} \hrulefill Pg
\pageref{exp_expes}\\
\noindent Supports for rados (complement to Table \ref{tc1_errs_rr})\hrulefill Pg \pageref{exp_compl_tc1_errs_rr}\\
\noindent Experiments on \textit{class-wise} rados\hrulefill Pg \pageref{exp_compl_tc1}\\
\noindent Test errors and supports for rados (comparison last vs
  best empirical classifier)\hrulefill Pg \pageref{exp_compl_tc1_errs_supp_all}\\

\newpage

\section{Proofs}\label{proof_proofs}

\subsection{Proof of Theorem \ref{thmstrongrc}}\label{proof_thm_thmstrongrc}

We split the proof in two parts, the first concerning the case where
both generators are differentiable since some of the derivations shall
be used hereafter, and then the case where they are not. Remark that
because of Lemma \ref{lemgens}, we do not have to cover the case
where just one of the two generators would be differentiable.

\noindent \textbf{Case 1}: $\varphi_{\lab{e}}, \varphi_{\lab{r}}$ are
differentiable. We show in this case that being proportionate is
equivalent to having:
\begin{eqnarray}
\ve{p}^*(\ve{z}) & = & \matrice{G}_m \ve{q}^*(\ve{z})\:\:.\label{eq111}
\end{eqnarray}
Solving
eqs. (\ref{opte}) and (\ref{optr}) bring respectively:
\begin{eqnarray}
p^*_i(\ve{z}) & = &
{\varphi'_{\lab{e}}}^{-1}\left(-\frac{1}{\upmu_{\lab{e}}}
  \cdot z_i\right)\:\:, \label{pp1}\\
q^*_{\mathcal{I}}(\ve{z}) & = & {\varphi'_{\lab{r}}}^{-1}\left(-\frac{1}{\upmu_{\lab{r}}}
  \cdot \sum_{i \in
  {\mathcal{I}}} z_i + \frac{\uplambda}{\upmu_{\lab{r}}} \right)\:\:, \label{pp2}
\end{eqnarray}
where $\uplambda$ is picked so that $\ve{q}^*(\ve{z}) \in
{\mathbb{H}}^{2^m}$, that is,
\begin{eqnarray}
\sum_{{\mathcal{I}} \subseteq [m]} {\varphi'_{\lab{r}}}^{-1}\left(-\frac{1}{\upmu_{\lab{r}}}
  \cdot \sum_{i \in
  {\mathcal{I}}} z_i + \frac{\uplambda}{\upmu_{\lab{r}}} \right) & = & 1\:\:.\label{plambda}
\end{eqnarray}
We obtain
\begin{eqnarray}
{\mathcal{L}}_{\lab{e}}^*(\ve{z}) & = & -\upmu_{\lab{e}} \sum_{i\in [m]} 
{\varphi_{\lab{e}}^{\star}} \left(-\frac{1}{\upmu_{\lab{e}}}
  \cdot z_i\right)\:\:,\label{simpe}\\
{\mathcal{L}}_{\lab{r}}^*(\ve{z}) & = & \uplambda -\upmu_{\lab{r}} \sum_{{\mathcal{I}} \subseteq [m]} 
{\varphi_{\lab{r}}}^{\star} \left(-\frac{1}{\upmu_{\lab{r}}}
  \cdot \sum_{i \in
  {\mathcal{I}}} z_i + \frac{\uplambda}{\upmu_{\lab{r}}}\right)\:\:,\label{simpr}
\end{eqnarray}
where $\varphi^\star(z) \defeq \sup_{z'} \{zz'-\varphi(z')\}$ denotes the convex
conjugate of $\varphi$. It follows from eq. (\ref{simpe}) that:
\begin{eqnarray}
\frac{\partial}{\partial z_i}
{\mathcal{L}}_{\lab{e}}^*(\ve{z}) & = &  {\varphi^\star_{\lab{e}}}' \left(-\frac{1}{\upmu_{\lab{e}}}
  \cdot z_i\right) \nonumber\\
 & = & {\varphi'_{\lab{e}}}^{-1} \left(-\frac{1}{\upmu_{\lab{e}}}
  \cdot z_i\right) \label{pppp1}\\
 & = & p^*_i(\ve{z})\:\:,\label{parte}
\end{eqnarray}
where eq. (\ref{pppp1}) follows from properties of $\varphi^\star$.
We also have
\begin{eqnarray}
\lefteqn{\frac{\partial}{\partial z_i}
{\mathcal{L}}_{\lab{r}}^*(\ve{z})}\nonumber\\
 & = & \left(1 - \sum_{{\mathcal{I}} \subseteq [m]} 
{\varphi'_{\lab{r}}}^{-1} \left(-\frac{1}{\upmu_{\lab{r}}}
  \cdot \sum_{j \in
  {\mathcal{I}}} z_j + \frac{\uplambda}{\upmu_{\lab{r}}}\right)\right) \cdot \frac{\partial
\uplambda}{\partial z_i} \nonumber\\
 & & + \sum_{{\mathcal{I}} \subseteq [m] : i \in {\mathcal{I}}} 
{\varphi'_{\lab{r}}}^{-1} \left(-\frac{1}{\upmu_{\lab{r}}}
  \cdot \sum_{j \in
  {\mathcal{I}}} z_j + \frac{\uplambda}{\upmu_{\lab{r}}}\right) \nonumber\\
 & = & \frac{\partial
\uplambda}{\partial z_i} \nonumber\\
 & & + \sum_{{\mathcal{I}} \subseteq [m]} 
\left(1_{i \in {\mathcal{I}}} - \frac{\partial
\uplambda}{\partial z_i} \right) {\varphi'_{\lab{r}}}^{-1} \left(-\frac{1}{\upmu_{\lab{r}}}
  \cdot \sum_{j \in
  {\mathcal{I}}} z_j +
\frac{\uplambda}{\upmu_{\lab{r}}}\right)\nonumber\\
& =&  \frac{\partial
\uplambda}{\partial z_i} + \sum_{{\mathcal{I}} \subseteq [m]} 
\left(1_{i \in {\mathcal{I}}} - \frac{\partial
\uplambda}{\partial z_i}  \right) \cdot q^*_{\mathcal{I}}(\ve{z}) \nonumber\\
& =& \frac{\partial
\uplambda}{\partial z_i} \cdot \left( 1 - \sum_{{\mathcal{I}}
  \subseteq [m]} q^*_{\mathcal{I}}(\ve{z}) \right) +
\sum_{{\mathcal{I}} \subseteq [m]} 1_{i \in {\mathcal{I}}} \cdot q^*_{\mathcal{I}}(\ve{z}) \nonumber\\
& =& \sum_{{\mathcal{I}} \subseteq [m]} 1_{i \in {\mathcal{I}}} \cdot q^*_{\mathcal{I}}(\ve{z})\:\:,\label{partr}
\end{eqnarray}
since $\ve{q}^*(\ve{z}) \in
{\mathbb{H}}^{2^m}$. 

Now suppose $\varphi_{\lab{e}}$ and $\varphi_{\lab{r}}$ proportionate. It
comes that there exists $(\upmu_{\lab{e}},\upmu_{\lab{r}})$ such that
the gradients of eq. (\ref{ppa}) yield $\nabla {\mathcal{L}}_{\lab{e}}^*(\ve{z}) =
\nabla {\mathcal{L}}_{\lab{r}}^*(\ve{z})$, and from
eqs. (\ref{parte}) and (\ref{partr}) we obtain $\ve{p}^*(\ve{z}) =
\matrice{G}_m \ve{q}^*(\ve{z})$. 

Reciprocally, having $\ve{p}^*(\ve{z}) = \matrice{G}_m
\ve{q}^*(\ve{z})$ for some $\varphi_{\lab{e}}, \varphi_{\lab{r}}$ and
$\upmu_{\lab{e}},\upmu_{\lab{r}} > 0$ implies as well $\nabla {\mathcal{L}}_{\lab{e}}^*(\ve{z}) =
\nabla {\mathcal{L}}_{\lab{r}}^*(\ve{z})$ from eqs. (\ref{parte}) and (\ref{partr}), and therefore
eq. (\ref{ppa}) holds as well. This ends the proof of Case 1 for
Theorem \ref{thmstrongrc}.\\

\noindent \textbf{Case 2}: $\varphi_{\lab{e}}, \varphi_{\lab{r}}$ are not
differentiable. To simplify the statement and proofs, we assume
that $\ve{\mu}_{\lab{e}} = \ve{\mu}_{\lab{r}} = 1$. We define the following problems
\begin{eqnarray}
{\mathcal{L}}_{\lab{e}}(\ve{z}) & \defeq & \inf_{\ve{p}\in {\mathbb{R}}^m} \ve{z}^\top \ve{p}
+ \varphi_{\lab{e}}(\ve{p})\:\:,\label{defoptegen}\\
{\mathcal{L}}_{\lab{r}}(\ve{z}) & \defeq & \inf_{\ve{q}\in
  {\mathbb{H}}^{2^m}} \ve{z}^\top
\matrice{G}_m \ve{q} + \varphi_{\lab{r}} (\ve{q})\:\:,\label{defoptrgen}
\end{eqnarray}
where $\varphi_{\lab{e}} : {\mathbb{R}}^m\rightarrow {\mathbb{R}}$ and
$\varphi_{\lab{r}} : {\mathbb{R}}^{2^m}\rightarrow {\mathbb{R}}$ are
convex. Recall that
$\partial {\mathcal{L}}_{\lab{e}}$ and $\partial
{\mathcal{L}}_{\lab{r}}$ are their subdifferentials, and $\ve{p}(\ve{z})$
and $\ve{q}(\ve{z})$ the arguments of the infima, assuming without
loss of generality that they are
finite. We now show that being proportionate is equivalent to having,
for any $\ve{z}$,
\begin{eqnarray}
\ve{p}(\ve{z}) & \in & \partial {\mathcal{L}}_{\lab{r}}(\ve{z})\:\:,\label{pp2}\\
\matrice{G}_m \ve{q}(\ve{z}) & \in & \partial  {\mathcal{L}}_{\lab{e}}(\ve{z})\:\:.\label{qq2}
\end{eqnarray}
This property is an immediate consequence of the following property,
which we shall in fact show:
\begin{eqnarray}
\ve{p}(\ve{z}) & \in & \partial {\mathcal{L}}_{\lab{e}}(\ve{z})\:\:,\label{pp}\\
\matrice{G}_m \ve{q}(\ve{z}) & \in & \partial {\mathcal{L}}_{\lab{r}}(\ve{z})\:\:.\label{qq}
\end{eqnarray}
Granted all (\ref{pp2}---\ref{qq}) hold, Eq. (\ref{eq111}) of Theorem
\ref{thmstrongrc} follows whenever subgradients are singletons. To see
why the statement of the Theorem follows from
(\ref{pp2}--\ref{qq2}), if the functions are proportionate, then their
subdifferentials match from Definition \ref{sequiv} and we immediately get (\ref{pp2}) and
(\ref{qq2}) from  (\ref{pp}) and
(\ref{qq}). If, on the other hand, we have both (\ref{pp2}) and
(\ref{qq2}), then we get from (\ref{pp}) and
(\ref{qq}) that $\partial {\mathcal{L}}_{\lab{e}}(\ve{z})
\cap \partial {\mathcal{L}}_{\lab{r}}(\ve{z}) \neq \emptyset, \forall
\ve{z}$ and so $\ve{0}\in \partial ({\mathcal{L}}_{\lab{e}}(\ve{z}) -
{\mathcal{L}}_{\lab{r}}(\ve{z}))$, yieding the fact that the
epigraphs of ${\mathcal{L}}_{\lab{e}}(\ve{z})$ and
${\mathcal{L}}_{\lab{r}}(\ve{z})$ match by a translation of some $b$
that does not depend on $\ve{z}$, and by extension, the fact that
$\varphi_{\lab{e}}$ and $\varphi_{\lab{r}}$ meet Definition \ref{sequiv} and are
proportionate.

To show
 (\ref{pp}), we first remark that $-\ve{z}'\in \partial
\varphi_{\lab{e}}(\ve{p}(\ve{z}'))$ for any $\ve{z}'$ because of the definition of
$\ve{p}$ in (\ref{defoptegen}). So, from the definition of
subdifferentials, for any $\ve{z}$,
\begin{eqnarray}
\varphi_{\lab{e}}(\ve{p}(\ve{z}')) +
(-\ve{z}')^\top(\ve{p}(\ve{z})-\ve{p}(\ve{z}')) & \leq & \varphi_{\lab{e}}(\ve{p}(\ve{z}))\:\:.\nonumber
\end{eqnarray}
Reorganising and substracting $\ve{z}^\top\ve{p}(\ve{z})$ to both
sides, we get
\begin{eqnarray}
\lefteqn{-\varphi_{\lab{e}}(\ve{p}(\ve{z}'))-\ve{z}'^\top\ve{p}(\ve{z}')}\nonumber\\
 & \geq & -\varphi_{\lab{e}}(\ve{p}(\ve{z}))-\ve{z}^\top\ve{p}(\ve{z}) + (-\ve{p}(\ve{z}))^\top(\ve{z}'-\ve{z})\:\:,\nonumber
\end{eqnarray}
which shows that $-\ve{p}(\ve{z}) \in \partial
-(\varphi_{\lab{e}}(\ve{p}(\ve{z}))+\ve{z}^\top\ve{p}(\ve{z}))$, and
so $\ve{p}(\ve{z}) \in \partial {\mathcal{L}}_{\lab{e}}(\ve{z})$. 

We then tackle (\ref{qq}). We show that there exists $\uplambda \in
{\mathbb{R}}$ such that $\uplambda\cdot \ve{1}_{2^m} - \matrice{g}^\top_m\ve{z}
\in \partial \varphi_{\lab{r}}(\ve{q}(\ve{z}))$ at the optimal
$\ve{q}(\ve{z})$. Suppose it is not the case. Then because of the definition
of subgradients, for any $\uplambda \in {\mathbb{R}}$, there exists
$\ve{q}\in {\mathbb{H}}^{2^m}, \ve{q} \neq \ve{q}(\ve{z})$ such that
\begin{eqnarray}
\varphi_{\lab{r}}(\ve{q}(\ve{z})) + (\uplambda\cdot \ve{1}_{2^m} -
\matrice{g}^\top_m\ve{z})^\top (\ve{q}-\ve{q}(\ve{z})) & > & \varphi_{\lab{r}}(\ve{q})\:\:.\nonumber
\end{eqnarray}
Reorganising and using the fact that $\ve{q}, \ve{q}_* \in
{\mathbb{H}}^{2^m}$, we get $\varphi_{\lab{r}}(\ve{q}(\ve{z})) +
\ve{z}^\top \matrice{g}_m\ve{q}(\ve{z}) > \varphi_{\lab{r}}(\ve{q}) +
\ve{z}^\top \matrice{g}_m\ve{q}$, contradicting the optimality
of $\ve{q}(\ve{z})$. Consider any $\ve{z}'$ and its corresponding
optimal $\ve{q}(\ve{z}')$. Since $\uplambda'\cdot \ve{1}_{2^m} - \matrice{g}^\top_m\ve{z}
\in \partial \varphi_{\lab{r}}(\ve{q}(\ve{z}))$ for some $\uplambda' \in
{\mathbb{R}}$, we get from the
definition of subgradients that
\begin{eqnarray}
\lefteqn{\varphi_{\lab{r}}(\ve{q}(\ve{z}))}\nonumber\\
 & \geq & \varphi_{\lab{r}}(\ve{q}(\ve{z}')) + (\uplambda'\cdot \ve{1}_{2^m} -
\matrice{g}^\top_m\ve{z}')^\top
(\ve{q}(\ve{z})-\ve{q}(\ve{z}')) \nonumber\:\:.
\end{eqnarray}
Reorganising and using the fact that $\ve{q}(\ve{z}),
\ve{q}(\ve{z}')\in {\mathbb{H}}^{2^m}$, we get
\begin{eqnarray}
\lefteqn{-(\varphi_{\lab{r}}(\ve{q}(\ve{z}')) +
  \ve{z}'^\top\matrice{g}_m\ve{q}(\ve{z}'))}\nonumber\\
 & \geq & -(\varphi_{\lab{r}}(\ve{q}(\ve{z})) +
  \ve{z}^\top\matrice{g}_m\ve{q}(\ve{z})) \nonumber\\
 & & +
  (-\matrice{g}_m\ve{q}(\ve{z}))^\top(\ve{z}' -\ve{z})\:\:,
\end{eqnarray}
showing that $-\matrice{g}_m\ve{q}(\ve{z}) \in \partial -(\varphi_{\lab{r}}(\ve{q}(\ve{z})) +
  \ve{z}^\top\matrice{g}_m\ve{q}(\ve{z}))$, and so $\matrice{g}_m\ve{q}(\ve{z}) \in \partial  {\mathcal{L}}_{\lab{r}}(\ve{z})$.

\subsection{Proof of Lemma \ref{lemgens}}\label{proof_lem_lemgens}

Take $m=1$, and replace $\ve{z}$ by real $z_1$. We have
${\mathcal{L}}_{\lab{e}}(p, z_1) = pz_1 + \varphi_{\lab{e}}(z_1)$ and
${\mathcal{L}}_{\lab{r}}(\ve{q}, z) = q_{\{1\}}z_1 +
\varphi_{\lab{r}}(q_{\{1\}}) +
\varphi_{\lab{r}}(q_{\emptyset})$. Remark that we can drop the
constraint $\ve{q}\in {\mathbb{H}}^{2}$ since then $q_{\emptyset} = 1
- q_{\{1\}}$. So we get
\begin{eqnarray*}
{\mathcal{L}}^*_{\lab{r}}(\ve{q}) & = & \min_{q\in {\mathbb{R}}} qz_1 +
\upmu_{\lab{r}}\varphi_{\lab{r}}(q) + \upmu_{\lab{r}}\varphi_{\lab{r}}(1-q)\\
 & = & \min_{q\in {\mathbb{R}}} qz_1 +
 \upmu_{\lab{r}}\varphi_{\lab{s}({\lab{r}})}(q)\\
 & = & -
 \upmu_{\lab{r}}\varphi^\star_{\lab{s}({\lab{r}})}\left(-\frac{1}{\upmu_{\lab{r}}}
   \cdot z_1\right)\:\:,
\end{eqnarray*}
whereas 
\begin{eqnarray*}
{\mathcal{L}}^*_{\lab{e}}(p) & = & -
 \upmu_{\lab{e}}\varphi^\star_{\lab{r}}\left(-\frac{1}{\upmu_{\lab{e}}}
   \cdot z_1\right)\:\:,
\end{eqnarray*}
and since $\varphi_{\lab{e}}$ and $\varphi_{\lab{r}}$ are
proportionate, then
\begin{eqnarray}
\varphi^\star_{\lab{r}}\left(-\frac{1}{\upmu_{\lab{e}}}
   \cdot z_1\right) & = & \frac{\upmu_{\lab{r}}}{\upmu_{\lab{e}}}\cdot \varphi^\star_{\lab{s}({\lab{r}})}\left(-\frac{1}{\upmu_{\lab{r}}}
   \cdot z_1\right) - \frac{b}{\upmu_{\lab{e}}}\:\:.
\end{eqnarray}
We then make the variable change $z \defeq -z_1 / \upmu_{\lab{e}}$ and
get
\begin{eqnarray}
\varphi^\star_{\lab{e}}(z) & = & \frac{\upmu_{\lab{r}}}{\upmu_{\lab{e}}}\cdot \varphi^\star_{\lab{s}({\lab{r}})}\left(\frac{\upmu_{\lab{e}}}{\upmu_{\lab{r}}}
   \cdot z\right) - \frac{b}{\upmu_{\lab{e}}}\:\:,
\end{eqnarray}
which yields, since $\varphi_{\lab{e}}$, $\varphi_{\lab{r}}$, and by
extension $\varphi_{\lab{s}(\lab{r})}$, are all convex and lower-semicontinuous,
\begin{eqnarray}
\varphi_{\lab{e}}(z) & = & \frac{\upmu_{\lab{r}}}{\upmu_{\lab{e}}}\cdot \varphi_{\lab{s}({\lab{r}})}(z) + \frac{b}{\upmu_{\lab{e}}}\:\:,
\end{eqnarray}
as claimed.

\subsection{Proof of Lemma \ref{srclog}}\label{proof_lem_srclog}

We use the fact that whenever $\varphi$ is differentiable,
$\varphi^\star(z) \defeq z \cdot {\varphi'}^{-1}(z)- \varphi({\varphi'}^{-1}(z))$.
We have $\varphi'_{\lab{r}}(z) = \log z$,
${\varphi'_{\lab{r}}}^{-1}(z) = \exp z =
\varphi^\star_{\lab{r}}(z)$. Therefore, the Lagrange multiplier
$\uplambda$ in (\ref{plambda}) is 
\begin{eqnarray}
\uplambda & = & -\upmu_{\lab{r}}\cdot \log\left(\sum_{{\mathcal{I}} \subseteq [m]} {\exp\left(-\frac{1}{\upmu_{\lab{r}}}
  \cdot \sum_{i \in
  {\mathcal{I}}} z_i\right)}\right)\:\:,\label{deflambdaexp}
\end{eqnarray}
which yields from (\ref{pp2}):
\begin{eqnarray}
q^*_{\mathcal{I}}(\ve{z}) & = & \frac{\exp\left(-\frac{1}{\upmu_{\lab{r}}}
  \cdot \sum_{i \in
  {\mathcal{I}}} z_i \right)}{\sum_{{\mathcal{J}} \subseteq [m]} {\exp\left(-\frac{1}{\upmu_{\lab{r}}}
  \cdot \sum_{j \in
  {\mathcal{J}}} z_j\right)}} \:\:, \forall {\mathcal{I}} \subseteq [m]\:\:.\nonumber
\end{eqnarray}
On the other hand, we also have $\varphi'_{\lab{s}} (z) =
\log(z/(1-z))$, ${\varphi'_{\lab{s}}}^{-1} (z) =
\exp(z)/(1+\exp(z))$ and $\varphi^\star_{\lab{s}}(z) = 1+
\log(1+\exp(z))$, which yields from (\ref{pp1}):
\begin{eqnarray}
p^*_i(\ve{z}) & = &
\frac{\exp\left(-\frac{1}{\upmu_{\lab{e}}}
  \cdot z_i\right)}{1+\exp\left(-\frac{1}{\upmu_{\lab{e}}}
  \cdot z_i\right)}\:\:, \forall i \in [m]\:\:.\label{elog1}
\end{eqnarray}
We then check that for any $i\in [m]$, we indeed have
\begin{eqnarray}
\lefteqn{\sum_{{\mathcal{I}} \subseteq [m]} 1_{i \in {\mathcal{I}}}
  \cdot q^*_{\mathcal{I}}(\ve{z})}\nonumber\\
 & = & \sum_{{\mathcal{I}} \subseteq [m]} 1_{i \in {\mathcal{I}}}
  \cdot \frac{\exp\left(-\frac{1}{\upmu_{\lab{r}}}
  \cdot \sum_{i' \in
  {\mathcal{I}}} z_{i'} \right)}{\sum_{{\mathcal{J}} \subseteq [m]} {\exp\left(-\frac{1}{\upmu_{\lab{r}}}
  \cdot \sum_{j \in
  {\mathcal{J}}} z_j\right)}} \nonumber\\
 & = & \exp\left(-\frac{1}{\upmu_{\lab{e}}}
  \cdot z_i \right)\cdot \frac{\sum_{{\mathcal{J}} \subseteq
    [m]\backslash \{i\}} {\exp\left(-\frac{1}{\upmu_{\lab{r}}}
  \cdot \sum_{j \in
  {\mathcal{I}}} z_{j}\right)}}{\sum_{{\mathcal{J}} \subseteq [m]} {\exp\left(-\frac{1}{\upmu_{\lab{r}}}
  \cdot \sum_{j \in
  {\mathcal{J}}} z_j\right)}}\nonumber\\
 & = & \exp\left(-\frac{1}{\upmu_{\lab{e}}}
  \cdot z_i \right)\cdot \frac{c}{\left( 1+\exp\left(-\frac{1}{\upmu_{\lab{e}}}
  \cdot z_i\right)\right)\cdot c}\nonumber\\
 & = & \frac{\exp\left(-\frac{1}{\upmu_{\lab{r}}}
  \cdot z_i \right)}{1+\exp\left(-\frac{1}{\upmu_{\lab{r}}}
  \cdot z_i\right)} \:\:,\label{llog1}
\end{eqnarray}
with $c \defeq \sum_{{\mathcal{J}} \subseteq
    [m]\backslash \{i\}} {\exp\left(-\frac{1}{\upmu_{\lab{r}}}
  \cdot \sum_{j \in
  {\mathcal{I}}} z_{j}\right)}$. We check that eq. (\ref{llog1})
equals eq. (\ref{elog1}) whenever $\upmu_{\lab{e}} = \upmu_{\lab{r}}$.
Hence eq. (\ref{eq111}) holds. We conclude that
$\varphi_{\lab{r}}$ and $\varphi_{\lab{e}}
= \varphi_{\lab{s}}$ are proportionate whenever $\upmu_{\lab{e}} = \upmu_{\lab{r}}$.

\subsection{Proof of Corollary \ref{corlog}}\label{proof_cor_corlog}

Consider $\varphi_{\lab{r}}(z) \defeq z\log z - z$ and $\varphi_{\lab{e}}
= \varphi_{\lab{s}}$. We obtain from eq. (\ref{simpe}):
\begin{eqnarray}
\lefteqn{-{\mathcal{L}}_{\lab{e}}^*(\ve{z})}\nonumber\\
 & = & f_{\lab{e}} \left(\sum_{i\in [m]} 
\log \left(1+\exp\left(-\frac{1}{\upmu_{\lab{e}}}
  \cdot z_i\right)\right)\right) \:\:,\nonumber
\end{eqnarray}
with $f_{\lab{e}}(z) =
\upmu_{\lab{e}}  \cdot z + 
\upmu_{\lab{e}}m$. We have also $\varphi_{\lab{r}}^\star (z) =
\exp(z)$, and so using $\uplambda$ in eq. (\ref{deflambdaexp}) and eq. (\ref{simpr}),
we obtain
\begin{eqnarray}
\lefteqn{-{\mathcal{L}}_{\lab{r}}^*(\ve{z})}\nonumber\\
 & = & \upmu_{\lab{r}}\cdot \log\left(\sum_{{\mathcal{I}} \subseteq [m]} {\exp\left(-\frac{1}{\upmu_{\lab{r}}}
  \cdot \sum_{i \in
  {\mathcal{I}}} z_i\right)}\right) \nonumber\\
 & & + \upmu_{\lab{r}}\cdot \exp\left(\frac{\uplambda}{\upmu_{\lab{r}}}\right)\cdot \sum_{{\mathcal{I}} \subseteq [m]} \exp\left(-\frac{1}{\upmu_{\lab{r}}}
  \cdot \sum_{i \in
  {\mathcal{I}}} z_i\right) \nonumber\\
 & = & \upmu_{\lab{r}}\cdot \log\left(\sum_{{\mathcal{I}} \subseteq [m]} {\exp\left(-\frac{1}{\upmu_{\lab{r}}}
  \cdot \sum_{i \in
  {\mathcal{I}}} z_i\right)}\right) \nonumber\\
 & & +\upmu_{\lab{r}}\cdot \underbrace{\frac{\sum_{{\mathcal{I}} \subseteq [m]} \exp\left(-\frac{1}{\upmu_{\lab{r}}}
  \cdot \sum_{i \in
  {\mathcal{I}}} z_i\right)}{\sum_{{\mathcal{I}} \subseteq [m]} \exp\left(-\frac{1}{\upmu_{\lab{r}}}
  \cdot \sum_{i \in
  {\mathcal{I}}} z_i\right)}}_{=1}\nonumber\\
 & = & f_{\lab{r}} \left(\sum_{{\mathcal{I}} \subseteq [m]} {\exp\left(-\frac{1}{\upmu_{\lab{r}}}
  \cdot \sum_{i \in
  {\mathcal{I}}} z_i\right)}\right)\nonumber\:\:,
\end{eqnarray}
with $f_{\lab{r}}(z) =
\upmu_{\lab{r}} \cdot \log z+
\upmu_{\lab{r}}$. We get from Lemma \ref{srclog} that the following example and rado risks are equivalent whenever $\upmu_{\lab{e}} = \upmu_{\lab{r}}$:
\begin{eqnarray}
\ell_{\lab{e}}(\ve{z}, \upmu_{\lab{e}}) & = & \sum_{i\in [m]} \log \left(1+\exp\left(-\frac{1}{\upmu_{\lab{e}}}
  \cdot z_i\right)\right)\:\:,\label{ploge0}\\
\ell_{\lab{r}}(\ve{z}, \upmu_{\lab{r}}) & = & \sum_{{\mathcal{I}} \subseteq [m]} {\exp\left(-\frac{1}{\upmu_{\lab{r}}}
  \cdot \sum_{i \in
  {\mathcal{I}}} z_i\right)}\:\:, \label{plogr0}
\end{eqnarray}
from which we get the statement of the Corollary by fixing $\upmu = \upmu_{\lab{e}} = \upmu_{\lab{r}}$.

\subsection{Proof of Lemma \ref{srcsql}}\label{proof_lem_srcsql}

We proceed as in the proof of Lemma \ref{srclog}. 
We have $\varphi'_{\lab{r}}(z) = z$,
${\varphi'_{\lab{r}}}^{-1}(z) = z$ and
$\varphi^\star_{\lab{r}}(z) = \varphi_{\lab{r}}(z)$. Therefore, the Lagrange multiplier
$\uplambda$ in (\ref{plambda}) is 
\begin{eqnarray}
\uplambda & = & \frac{\upmu_{\lab{r}}}{2^m} + \frac{1}{2^m}
  \cdot \sum_{{\mathcal{I}} \subseteq [m]} \sum_{i \in
  {\mathcal{I}}} z_i\label{deflambdasql0}\\
 & = & \frac{\upmu_{\lab{r}}}{2^m} + \frac{1}{2}
  \cdot \sum_{i \in
  [m]} z_i\:\:,\label{deflambdasql}
\end{eqnarray}
since any $i$ belongs exactly to half of the subsets of $[m]$. We obtain:
\begin{eqnarray}
q^*_{\mathcal{I}}(\ve{z}) & = & \frac{1}{2^m} - \frac{1}{
  \upmu_{\lab{r}}}\cdot \sum_{i \in
  {\mathcal{I}}} z_i + \frac{1}{2 \upmu_{\lab{r}}}
  \cdot \sum_{i \in
  [m]} z_i \:\:, \forall {\mathcal{I}} \subseteq [m]\:\:.\nonumber
\end{eqnarray}
On the other hand, we also have $\varphi'_{\lab{s}} (z) =
2z-1$, ${\varphi'_{\lab{s}}}^{-1} (z) =
(1+z)/2$ and $\varphi^\star_{\lab{s}}(z) = -(1/4) + (1/4)\cdot (1+z)^2$, which yields from (\ref{pp1}):
\begin{eqnarray}
p^*_i(\ve{z}) & = &
\frac{1}{2}\cdot\left( 1 - \frac{1}{
  \upmu_{\lab{e}}}\cdot z_i\right)\:\:, \forall i \in [m]\:\:.\label{esql1}
\end{eqnarray}
We then check that for any $i\in [m]$, we have
\begin{eqnarray}
\lefteqn{\sum_{{\mathcal{I}} \subseteq [m]} 1_{i \in {\mathcal{I}}}
  \cdot q^*_{\mathcal{I}}(\ve{z})}\nonumber\\
 & = & \sum_{{\mathcal{I}} \subseteq [m]} 1_{i \in {\mathcal{I}}}
  \cdot \left( \frac{1}{2^m} - \frac{1}{
  \upmu_{\lab{r}}}\cdot \sum_{i \in
  {\mathcal{I}}} z_i + \frac{1}{2 \upmu_{\lab{r}}}
  \cdot \sum_{i \in
  [m]} z_i\right) \nonumber\\
 & = & \frac{1}{2} - \frac{1}{
  \upmu_{\lab{r}}}\cdot  \sum_{{\mathcal{I}} \subseteq [m]} 1_{i \in
  {\mathcal{I}}}\cdot \sum_{i \in
  {\mathcal{I}}} z_i + \frac{2^{m-2}}{\upmu_{\lab{r}}}
  \cdot \sum_{i \in
  [m]} z_i \nonumber\\
 & = & \frac{1}{2} - \frac{2^{m-1}}{
  \upmu_{\lab{r}}}\cdot  z_i - \frac{1}{
  \upmu_{\lab{r}}}\cdot \sum_{{\mathcal{I}} \subseteq [m]\backslash\{i\}} \sum_{i \in
  {\mathcal{I}}} z_i \nonumber\\
 & & + \frac{2^{m-2}}{\upmu_{\lab{r}}}
  \cdot \sum_{i \in
  [m]} z_i \nonumber\\
 & = & \frac{1}{2} - \frac{2^{m-1}}{
  \upmu_{\lab{r}}}\cdot  z_i - \frac{2^{m-2}}{
  \upmu_{\lab{r}}}\cdot \sum_{i \in
  [m]\backslash\{i\}} z_i \nonumber\\
 & & + \frac{2^{m-2}}{\upmu_{\lab{r}}}
  \cdot \sum_{i \in
  [m]} z_i \nonumber\\
 & = & \frac{1}{2} - \frac{2^{m-1}}{
  \upmu_{\lab{r}}}\cdot  z_i  +\frac{2^{m-2}}{\upmu_{\lab{r}}}
  \cdot z_i\nonumber\\
 & = & \frac{1}{2} \left( 1 - \frac{2^{m-1}}{
  \upmu_{\lab{r}}}\cdot  z_i\right)\:\:.\label{lsql1}
\end{eqnarray}
We check that eq. (\ref{lsql1})
equals eq. (\ref{esql1}) whenever $\upmu_{\lab{e}} = \upmu_{\lab{r}} /
2^{m-1} $.
Hence eq. (\ref{eq111}) holds. We conclude that
$\varphi_{\lab{r}}$ is proportionate to $\varphi_{\lab{e}}
= \varphi_{\lab{s}}$ whenever $\upmu_{\lab{e}} = \upmu_{\lab{r}}/
2^{m-1} $.

\subsection{Proof of Corollary \ref{corsql}}\label{proof_cor_corsql}

Consider $\varphi_{\lab{r}}(z) \defeq (1/2)\cdot z^2$ and $\varphi_{\lab{e}}
= \varphi_{\lab{s}}$. We obtain from eq. (\ref{simpe}):
\begin{eqnarray}
\lefteqn{-{\mathcal{L}}_{\lab{e}}^*(\ve{z})}\nonumber\\
 & = & f_{\lab{e}} \left(\sum_{i\in [m]} 
\left(1- \frac{1}{\upmu_{\lab{e}}}
  \cdot z_i\right)^2\right) \:\:,\nonumber
\end{eqnarray}
with $f_{\lab{e}}(z) =
(\upmu_{\lab{e}}/4) \cdot z + (\upmu_{\lab{e}}m/4)$. We have also $\varphi_{\lab{r}}^\star (z) =
(1/2)\cdot z^2$, and so using eq. (\ref{simpr}) and $\uplambda$ in eq. (\ref{deflambdasql0}),
we obtain
\begin{eqnarray}
\lefteqn{-{\mathcal{L}}_{\lab{r}}^*(\ve{z})}\nonumber\\
 & = & -\frac{\upmu_{\lab{r}}}{2^m} - \frac{1}{2^m}
  \cdot \sum_{{\mathcal{I}} \subseteq [m]} \sum_{i \in
  {\mathcal{I}}} z_i \nonumber\\
 & & + \frac{1}{2 \upmu_{\lab{r}}}\sum_{{\mathcal{I}} \subseteq [m]}\left( 
  \sum_{i \in
  {\mathcal{I}}} z_i  -\frac{\upmu_{\lab{r}}}{2^m} - \frac{1}{2^m}
  \cdot \sum_{{\mathcal{I}} \subseteq [m]} \sum_{i \in
  {\mathcal{I}}} z_i\right)^2\nonumber\\
 & = & -\frac{\upmu_{\lab{r}}}{2^m} - \frac{1}{2^m}
  \cdot \sum_{{\mathcal{I}} \subseteq [m]} \sum_{i \in
  {\mathcal{I}}} z_i +\frac{\upmu_{\lab{r}}}{2^{m+1}}\nonumber\\
 & & - \frac{1}{2^m}\cdot \underbrace{\sum_{{\mathcal{I}} \subseteq [m]}
 \left(\sum_{i \in
  {\mathcal{I}}} z_i  - \frac{1}{2^m}
  \cdot \sum_{{\mathcal{I}} \subseteq [m]} \sum_{i \in
  {\mathcal{I}}} z_i\right)}_{=0}\nonumber\\
 & & + \frac{1}{2 \upmu_{\lab{r}}}\sum_{{\mathcal{I}} \subseteq [m]}\left( 
  \sum_{i \in
  {\mathcal{I}}} z_i  -\frac{1}{2^m}
  \cdot \sum_{{\mathcal{I}} \subseteq [m]} \sum_{i \in
  {\mathcal{I}}} z_i\right)^2\nonumber\\
 & = & -\frac{\upmu_{\lab{r}}}{2^{m+1}} - \frac{1}{2^m}
  \cdot \sum_{{\mathcal{I}} \subseteq [m]} \sum_{i \in
  {\mathcal{I}}} z_i \nonumber\\
 & & + \frac{2^{m-1}}{\upmu_{\lab{r}}}\cdot \frac{1}{2^m}
  \cdot \sum_{{\mathcal{I}} \subseteq [m]}\left( 
  \sum_{i \in
  {\mathcal{I}}} z_i  -\frac{1}{2^m}
  \cdot \sum_{{\mathcal{I}} \subseteq [m]} \sum_{i \in
  {\mathcal{I}}} z_i\right)^2\nonumber\\
 & = & -\frac{\upmu_{\lab{r}}}{2^{m+1}} \nonumber\\
 & & - \expect_{{\mathcal{I}} \sim
   [m]}\left[ \sum_{i \in
  {\mathcal{I}}} z_i\right] + \frac{2^{m-1}}{\upmu_{\lab{r}}}\cdot \var_{{\mathcal{I}} \sim
   [m]}\left[ \sum_{i \in
  {\mathcal{I}}} z_i\right] \nonumber\\
 & = & -\frac{\upmu_{\lab{r}}}{2^{m+1}} \nonumber\\
 & & + \frac{\upmu_{\lab{r}}}{2^{m-1}}\cdot \left(-\left(
\begin{array}{l}
\expect_{{\mathcal{I}} \sim
   [m]}\left[ \frac{2^{m-1}}{\upmu_{\lab{r}}}\cdot \sum_{i \in
  {\mathcal{I}}} z_i\right] \\
- \frac{\upmu_{\lab{r}}}{2^{m-1}} \cdot \var_{{\mathcal{I}} \sim
   [m]}\left[ \frac{2^{m-1}}{\upmu_{\lab{r}}}\sum_{i \in
  {\mathcal{I}}} z_i\right] 
\end{array}\right)\right)\nonumber\\
 & = & f_{\lab{r}} \left(-\left(
\begin{array}{l}
\expect_{{\mathcal{I}} \sim
   [m]}\left[ \frac{2^{m-1}}{\upmu_{\lab{r}}}\cdot \sum_{i \in
  {\mathcal{I}}} z_i\right] \\
- \frac{\upmu_{\lab{r}}}{2^{m-1}} \cdot \var_{{\mathcal{I}} \sim
   [m]}\left[ \frac{2^{m-1}}{\upmu_{\lab{r}}}\sum_{i \in
  {\mathcal{I}}} z_i\right] 
\end{array}\right)\right) \:\:,
\end{eqnarray}
with $f_{\lab{r}}(z) =
(\upmu_{\lab{r}} / 2^{m-1})\cdot z
-(\upmu_{\lab{r}}/2^{m+1})$. Therefore, it comes from Lemma
\ref{srcsql} that the following example and rado risks are equivalent whenever $\upmu_{\lab{e}} = \upmu_{\lab{r}}/
2^{m-1}$:
\begin{eqnarray}
\ell_{\lab{e}}(\ve{z}, \upmu_{\lab{e}}) & = & \sum_{i\in [m]} 
\left(1- \frac{1}{\upmu_{\lab{e}}}
  \cdot z_i\right)^2\:\:,\nonumber\\
\ell_{\lab{r}}(\ve{z}, \upmu_{\lab{r}}) & = & -\left(\expect_{{\mathcal{I}} }\left[ \frac{2^{m-1}}{\upmu_{\lab{r}}}\cdot \sum_{i \in
  {\mathcal{I}}} z_i\right] \right.\nonumber\\
 & & \left.- \frac{\upmu_{\lab{r}}}{2^{m-1}} \cdot \var_{{\mathcal{I}} }\left[ \frac{2^{m-1}}{\upmu_{\lab{r}}}\cdot \sum_{i \in
  {\mathcal{I}}} z_i\right]\right)\:\:.\nonumber
\end{eqnarray}
There remains to fix $\upmu \defeq \upmu_{\lab{e}} = \upmu_{\lab{r}}/
2^{m-1}$ to obtain the statement of the Corollary.

\subsection{Proof of Lemma \ref{srcrelu}}\label{proof_lem_srcrelu}

Define $\bigtriangleup_d$ as the $d$-dimensional probability
simplex. Then it comes with that choice of $\varphi_{\lab{r}}(q_{{\mathcal{I}}})$:
\begin{eqnarray}
\lefteqn{\min_{\ve{q}\in {\mathbb{H}}^{2^m}} {\mathcal{L}}_{\lab{r}}(\ve{q},
\ve{z})}\nonumber\\
 & = & \min_{\ve{q}\in \bigtriangleup_{2^m}} \sum_{{\mathcal{I}} \subseteq [m]} q_{{\mathcal{I}}} \sum_{i \in
  {\mathcal{I}}} z_i \nonumber\\
 & = & \left\{
\begin{array}{rcl}
0 & \mbox{ if } & \sum_{i \in
  {\mathcal{I}}} z_i > 0, \forall {\mathcal{I}} \neq \emptyset\:\:,\\
\sum_{i  : z_i < 0} z_i & \multicolumn{2}{l}{\mbox{ otherwise}} 
\end{array}
\right.\:\:,\label{probb1}
\end{eqnarray}
since whenever no $z_i$ is negative, the minimum is achieved by
putting all the mass (1) on $q_\emptyset$, and when some are negative,
the minimum is achieved by putting all the mass on the smallest over
all ${\mathcal{I}}$ of $\sum_{i \in
  {\mathcal{I}}} z_i$, which is the one which collects all the indexes
of the negative coordinates in $\ve{z}$. 

On the other hand, remark that fixing $\varphi_{\lab{e}} \defeq
\varphi_{\lab{s}}$ still yields $\varphi_{\lab{e}}(z) =
\upchi_{[0,1]}(z) = \varphi_{\lab{r}}(z)$, yet this time we have the following on
${\mathcal{L}}_{\lab{e}}(\ve{p}, \ve{z})$:
\begin{eqnarray}
\min_{\ve{p}\in {\mathbb{R}}^{m}} {\mathcal{L}}_{\lab{r}}(\ve{q},
\ve{z}) & = & \min_{\ve{p}\in [0,1]^m} \sum_{i\in [m]} p_i
z_i\nonumber\\
 & = & - \upmu_{\lab{e}} \cdot \sum_{i\in [m]} \max\left\{0, - \frac{1}{\upmu_{\lab{e}}}\cdot z_i\right\}\:\:,\label{lrelu}
\end{eqnarray}
since the optimal choice for $p^*_i$ is to put 1 only when $z_i$ is
negative. We obtain $\ve{p}^*(\ve{z}) =
\matrice{G}_m \ve{q}^*(\ve{z})$ for any choice of $\upmu_{\lab{e}}, \upmu_{\lab{r}}$, and so $\varphi_{\lab{r}}(z)$ is
self-proportionate for any $\upmu_{\lab{e}}, \upmu_{\lab{r}}$. This ends the proof of Lemma \ref{srcrelu}.

\subsection{Proof of Corollary \ref{correlu}}\label{proof_cor_correlu}

We obtain from Lemma \ref{srcrelu} that 
$-{\mathcal{L}}_{\lab{r}}^*(\ve{z}) = f_{\lab{r}}(\ell_{\lab{r}}(\ve{z}, \upmu_{\lab{r}}))$ with $f_{\lab{r}}(z) =
\upmu_{\lab{r}} \cdot z$ and:
\begin{eqnarray}
\ell_{\lab{r}}(\ve{z}, \upmu_{\lab{r}} ) & = & \max \left\{0,
  \max_{{\mathcal{I}} \subseteq [m]} \left\{-
\frac{1}{\upmu_{\lab{r}}} \cdot \sum_{i\in {\mathcal{I}}} z_i\right\}\right\}\:\:.\label{ppaz}
\end{eqnarray}
On the other hand, it comes from eq. (\ref{lrelu}) that $-{\mathcal{L}}_{\lab{e}}^*(\ve{z}) = f_{\lab{r}}(\ell_{\lab{e}}(\ve{z}, \upmu_{\lab{e}}))$ with $f_{\lab{e}}(z) =
\upmu_{\lab{e}} \cdot z$ and:
\begin{eqnarray}
\ell_{\lab{e}}(\ve{z}, \upmu_{\lab{e}} ) & = & \sum_{i\in [m]} \max\left\{0, - \frac{1}{\upmu_{\lab{e}}}\cdot z_i\right\}\:\:.\label{ppaz}
\end{eqnarray}
This concludes the proof of Corollary \ref{correlu}.

\subsection{Proof of Lemma \ref{srcunh}}\label{proof_lem_srcunh}

The choice of 
\begin{eqnarray}
\varphi_{\lab{r}}(z) & = & 
\upchi_{\left[\frac{1}{2^m}, \frac{1}{2}\right]}(z) \label{defvarphi}\:\:,
\end{eqnarray}
under the constraint that $\ve{q}\in {\mathbb{H}}^{2^m}$, 
enforces $q^*_{{\mathcal{I}}} = 1/2^m, \forall {\mathcal{I}} \subseteq
[m]$. Furthermore, fixing $\varphi_{\lab{e}} \defeq
\varphi_{\lab{s}}$ indeed yields
\begin{eqnarray}
\varphi_{\lab{e}} & = &
\upchi_{\left[\frac{1}{2^m}, \frac{1}{2}\right]}(z)
+\upchi_{\left[\frac{1}{2^m}, \frac{1}{2}\right]}(1-z)\nonumber\\
 & = & \upchi_{\left\{\frac{1}{2}\right\}}(z)\:\:,
\end{eqnarray} 
which enforces $p^*_i = 1/2$, $\forall i$. Since each $i$ belongs to
exactly $2^{m-1}$ subsets of $[m]$, we obtain $\ve{p}^*(\ve{z}) =
\matrice{G}_m \ve{q}^*(\ve{z})$, for any
$\upmu_{\lab{e}} , \upmu_{\lab{r}}$, and so $\varphi_{\lab{r}}$ is proportionate to $\varphi_{\lab{e}}
= \varphi_{\lab{s}}$ for any $\upmu_{\lab{e}}, \upmu_{\lab{r}}$.

\subsection{Proof of Corollary \ref{corunh}}\label{proof_cor_corunh}

We obtain from Lemma \ref{srcunh} that $-{\mathcal{L}}_{\lab{r}}^*(\ve{z}) = f_{\lab{r}}(\ell_{\lab{r}}(\ve{z}, \upmu_{\lab{r}}))$ with $f_{\lab{r}}(z) = z$ and:
\begin{eqnarray}
\ell_{\lab{r}}(\ve{z}, \upmu_{\lab{r}} ) & = & \expect_{{\mathcal{I}} }\left[ -\frac{1}{\upmu_{\lab{r}}}\cdot \sum_{i \in
  {\mathcal{I}}} z_i\right]\:\:.\nonumber
\end{eqnarray}
On the other hand, it comes from eq. (\ref{lrelu}) that $-{\mathcal{L}}_{\lab{e}}^*(\ve{z}) = f_{\lab{r}}(\ell_{\lab{e}}(\ve{z}, \upmu_{\lab{e}}))$ with $f_{\lab{e}}(z) =
(1/2) \cdot z$ and:
\begin{eqnarray}
\ell_{\lab{e}}(\ve{z}, \upmu_{\lab{e}} ) & = & \sum_i -\frac{1}{\upmu_{\lab{e}}}\cdot  z_i\:\:.\nonumber
\end{eqnarray}
This concludes the proof of Corollary \ref{correlu}.

\subsection{Proof of Theorem \ref{regrado}}\label{proof_thm_regrado}

The key to the poof is the constraint $\ve{q}\in {\mathbb{H}}^m$ in
eq. (\ref{optr}). Since $f_{\lab{e}}(z) = a_{\lab{e}}\cdot z + b_{\lab{e}}$, we
have ${\mathcal{L}}_{\lab{e}}^*(\ve{z}) = a_{\lab{e}} \cdot \left(\ell_{\lab{e}}(\ve{z})+\upomega\right) +
b_{\lab{e}} - a_{\lab{e}} \cdot \upomega$ for any $\upomega \in
{\mathbb{R}}$. It follows from eq. (\ref{ppa}) that $a_{\lab{e}} \cdot \left(\ell_{\lab{e}}(\ve{z})+\upomega\right) +
b_{\lab{e}} - a_{\lab{e}} \cdot \upomega = 
{\mathcal{L}}_{\lab{r}}^*(\ve{z}) + b = \sum_{{\mathcal{I}} \subseteq [m]} q^*_{{\mathcal{I}}} \sum_{i \in
  {\mathcal{I}}} z_i + \upmu_{\lab{r}} \sum_{{\mathcal{I}} \subseteq
  [m]} \varphi_{\lab{r}} (q^*_{{\mathcal{I}}}) + b$, and so
\begin{eqnarray*}
\lefteqn{a_{\lab{e}} \cdot \left(\ell_{\lab{e}}(\ve{z})+\upomega\right) +
b_{\lab{e}}}\nonumber\\
 & = & -\left\{\min_{\ve{q}\in {\mathbb{H}}^m} \left( \sum_{{\mathcal{I}} \subseteq [m]} q_{{\mathcal{I}}} \sum_{i \in
  {\mathcal{I}}} z_i + \upmu_{\lab{r}} \sum_{{\mathcal{I}} \subseteq
  [m]} \varphi_{\lab{r}} (q_{{\mathcal{I}}})\right) - a_{\lab{e}} \upomega\right\}\nonumber\\
& & + b\\
 & = & -\min_{\ve{q}\in {\mathbb{H}}^m} \left( \sum_{{\mathcal{I}} \subseteq [m]} q_{{\mathcal{I}}} \left(\sum_{i \in
  {\mathcal{I}}} z_i - a_{\lab{e}} \upomega\right)  + \upmu_{\lab{r}} \sum_{{\mathcal{I}} \subseteq
  [m]} \varphi_{\lab{r}} (q_{{\mathcal{I}}})\right) \nonumber\\
& & + b\:\:,
\end{eqnarray*}
since $\ve{q}\in {\mathbb{H}}^m$ and $a_{\lab{e}}, \upomega, a$ are
not a function of $\ve{q}$. We thus get $a_{\lab{e}} \cdot \left(\ell_{\lab{e}}(\ve{z})+\upomega\right) +
b_{\lab{e}} = a_{\lab{r}} \cdot
f_{\lab{r}}\left(\tilde{\ell}_{\lab{r}}(\ve{z})\right) + b_{\lab{r}}$,
where $\tilde{\ell}_{\lab{r}}(\ve{z})$ equals $\ell_{\lab{r}}(\ve{z})$ in
which each $\sum_{i \in
  {\mathcal{I}}} z_i$ is replaced by $\sum_{i \in
  {\mathcal{I}}} z_i - a_{\lab{e}} \upomega$. For $z_i =
\ve{\theta}^\top (y_i\cdot \ve{x}_i)$ and $\upomega =
\Omega(\ve{\theta})$, we obtain that whenever $\ve{\theta}\neq \ve{0}$, $\forall {\mathcal{I}} \subseteq [m]$,
\begin{eqnarray}
\sum_{i \in
  {\mathcal{I}}} z_i + a_{\lab{e}} \upomega & = &
\ve{\theta}^\top\left(\rado_{\ve{\sigma}} -
  \frac{a_{\lab{e}}\Omega(\ve{\theta})}{\|\ve{\theta}\|_2^2} \cdot \ve{\theta}\right)\:\:,\label{regz}
\end{eqnarray}
for $\sigma_i = y_i$ iff $i\in {\mathcal{I}}$ (and $-y_i$ otherwise),
and the statement of the Theorem follows.

\subsection{Proof of Theorem \ref{thadarL2}}\label{proof_thm_thadarL2}

The proof of the Theorem contains two parts, the first of which
follows \adaboostSS's exponential convergence rate proof, and the
second departs from this proof to cover \radaboostNoL.

We use the fact that $\alpha_{\iota (t)} \rado_{j \iota(t)} =
\alpha_{\iota (t)} \cdot \ve{1}_{\iota(t)}^\top \ve{\rado}_j =
(\ve{\theta}_{T}-\ve{\theta}_{T-1})^\top \ve{\rado}_j$ to unravel the
weights as:
\begin{eqnarray}
\lefteqn{w_{Tj}}\nonumber\\
 & = & \frac{w_{(T-1)j}}{Z_T} \cdot
    \exp\left(-\alpha_{\iota (T)} \rado_{j \iota(T)}  + \delta_T\right)\nonumber\\
 & = & \frac{w_{(T-1)j}}{Z_T} \cdot
    \exp\left(\begin{array}{l}
-(\ve{\theta}_{T}-\ve{\theta}_{T-1})^\top \ve{\rado}_{j} \\
+ \upomega\cdot
    (\|\ve{\theta}_{T}\|_2^2-\|\ve{\theta}_{T-1}\|_2^2)
\end{array}\right) \nonumber\\
 & = & \frac{w_{(T-1)j}}{Z_T} \cdot
    \exp\left(\begin{array}{l}
-\ve{\theta}_{T}^\top \left(\ve{\rado}_{j} - \upomega\cdot
    \ve{\theta}_{T}\right)\\
+ \ve{\theta}_{T-1}^\top \left(\ve{\rado}_{j} - \upomega\cdot
    \ve{\theta}_{T-1}\right) 
\end{array}\right) \nonumber\\
 & = & \frac{w_{0}}{\prod_{t=1}^T Z_t} \cdot
    \exp\left(\begin{array}{l}
-\ve{\theta}_{T}^\top \left(\ve{\rado}_{j} - \upomega\cdot
    \ve{\theta}_{T}\right)\\
+ \ve{\theta}_{0}^\top \left(\ve{\rado}_{j} - \upomega\cdot
    \ve{\theta}_{0}\right)
\end{array}\right) \label{tel1}\\
 & = & \frac{w_{0}}{\prod_{t=1}^T Z_t} \cdot
    \exp\left(-\ve{\theta}_{T}^\top \left(\ve{\rado}_{j} - \upomega\cdot
    \ve{\theta}_{T}\right)\right) \label{tel2}\:\:,
\end{eqnarray}
since the sums telescope in eq.  (\ref{tel1}) when we unravel the
weight update and $\ve{\theta}_0 = \ve{0}$. We therefore
get
\begin{eqnarray}
\ell^{\tiny{\mathrm{exp}}}_{\lab{r}}({\mathcal{S}}_r,
  \ve{\theta}, \|.\|_2^2)  & = & \prod_{t=1}^T Z_t\:\:,\label{bz1}
\end{eqnarray}
as in the classical \adaboostSS~analysis \citep{ssIBj}. This time
however, we have, letting $\tilde{\rado}_{j \iota(t)} \defeq \rado_{j \iota(t)}
   /\rado_{*\iota(t)} \in [-1,1]$ and $\tilde{\alpha}_{\iota(t)} \defeq
   \rado_{*\iota(t)}\cdot \alpha_t$ for short,
\begin{eqnarray}
\lefteqn{Z_{t+1}}\nonumber\\
 & = & \sum_{j\in [n]} w_{tj} \cdot \exp\left(-\alpha_{\iota(t)} \rado_{j \iota(t)} +
  \delta_t\right) \nonumber\\
 & = & \exp(\delta_t)\cdot \sum_{j\in [n]} w_{tj} \cdot \exp\left(-\alpha_{\iota(t)} \rado_{j \iota(t)} \right) \nonumber\\
 & = & \exp(\delta_t)\cdot \sum_{j\in [n]} w_{tj} \cdot \exp\left(-\tilde{\alpha}_{\iota(t)}\tilde{\rado}_{j
     \iota(t)} \right) \nonumber\\
 & \leq & \frac{\exp(\delta_t)}{2}\nonumber\\
 & & \cdot \sum_{j\in [n]} w_{tj} \cdot \left(\begin{array}{c}(1+\tilde{\rado}_{j
       \iota(t)})\cdot \exp\left(-\tilde{\alpha}_{\iota(t)} \right) \\
+ (1-\tilde{\rado}_{j
       \iota(t)})\cdot \exp\left(\tilde{\alpha}_{\iota(t)} \right) \end{array}\right) \label{pconv}\\
 &  & = \exp(\delta_t)\cdot \sqrt{1-r^2_t}\label{p60}\\
& = & \exp\left(\upomega\cdot(\|\ve{\theta}_t\|_2^2 -
\|\ve{\theta}_{t-1}\|_2^2) - \frac{1}{2}\ln
\frac{1}{1-r_t^2}\right)\:\:.  \nonumber
\end{eqnarray}
This is where our proof follows a different path from
\adaboostSS's: in eq. (\ref{p60}), we do not upperbound the $\sqrt{1-r^2_t}$ term, so it
can absorb more easily the new $\exp(\delta_t)$ factor which appears because of regularization.

Ineq. (\ref{pconv}) holds because of the convexity of $\exp$, and
eq. (\ref{p60}) is an equality when $r_t < \upgamma$. If $r_t >
\upgamma$ is clamped to $r_t \leftarrow \upgamma$ by the
weak learner in (\ref{defmuL2}), then we have instead the derivation
\begin{eqnarray}
\lefteqn{\sum_{j\in [n]} w_{tj} \cdot \left(\begin{array}{c}(1+\tilde{\rado}_{j
       \iota(t)})\cdot \exp\left(-\tilde{\alpha}_{\iota(t)} \right) \\
+ (1-\tilde{\rado}_{j
       \iota(t)})\cdot \exp\left(\tilde{\alpha}_{\iota(t)}
     \right) \end{array}\right)}\nonumber\\
 & = & \left(1+r_t\right)\cdot \sqrt{\frac{1 -
     \upgamma}{1+\upgamma}} + \left(1-r_t\right)\cdot \sqrt{\frac{1 +
     \upgamma}{1-\upgamma}} \label{fedc}\nonumber\\
 & \leq & 2\sqrt{1-\upgamma^2}\:\:,\label{bz1}
\end{eqnarray}
since function in (\ref{fedc}) is decreasing on $r_t > 0$. If $r_t <
-\upgamma$ is clamped to $r_t \leftarrow -\upgamma$, we get the
same conclusion as in ineq (\ref{bz1}) because this time
$\tilde{\alpha}_{\iota(t)} = (1/2)\cdot\ln((1 -
\upgamma)/(1+\upgamma))$. Summarising, whether $r_t$ has been
clamped or not by the weak learner in (\ref{defmuL2}), we get
\begin{eqnarray}
\lefteqn{Z_{t+1}}\nonumber\\
& \leq & \exp\left(\upomega\cdot(\|\ve{\theta}_t\|_2^2 -
\|\ve{\theta}_{t-1}\|_2^2) - \frac{1}{2}\ln
\frac{1}{1-r_t^2}\right)\:\:,  \label{plast}
\end{eqnarray}
with the additional fact that $|r_t|\leq \upgamma$.
For any feature index $k \in [d]$, let ${\mathcal{F}}_k\subseteq [T]$
the iteration indexes for which $\iota(t) = k$. Letting $\uplambda_{\Gamma}$ ($>0$) the
largest eigenvalue of $\Upgamma$, we obtain:
\begin{eqnarray}
\lefteqn{\prod_{t=1}^T Z_t}\nonumber\\
 & \leq & \exp\left(\upomega\cdot \|\ve{\theta}_T\|_{\Upgamma}^2
  - \sum_t \frac{1}{2}\log
\frac{1}{1-r_t^2}\right)\nonumber\\
 & \leq & \exp\left(\upomega \uplambda_{\Gamma}\cdot \|\ve{\theta}_T\|_2^2
  - \sum_t \frac{1}{2}\log
\frac{1}{1-r_t^2}\right)\nonumber\\
  & & = \exp\left(- \frac{1}{2}\cdot \sum_{k\in [d]} \Lambda_k \right) \:\:,\label{eqdelta1}
\end{eqnarray}
With
\begin{eqnarray}
\Lambda_k & \defeq &\log
\frac{1}{\prod_{t : \iota(t) \in {\mathcal{F}}_k} (1-r_t^2)}
\nonumber\\
 & & - \frac{\upomega \uplambda_{\Gamma}}{2 \rado^2_{*k}}\log^2 \prod_{t : \iota(t) \in {\mathcal{F}}_k}\left(\frac{1+r_t}{1-r_t}\right)\:\:.
\end{eqnarray}
Since $(\sum_{l=1}^u a_l)^2 \leq u\sum_{l=1}^u a_l^2$ and $\min_{k}
\max_j |\rado_{jk}|\leq |\rado_{*k}|$, $\Lambda_k$ satisfies:
\begin{eqnarray}
\Lambda_k & \geq & \sum_{t : \iota(t) \in {\mathcal{F}}_k} \left\{\log
\frac{1}{1-r_t^2}\right.
\nonumber\\
 & & \left. - \frac{T_k \upomega \uplambda_{\Gamma}}{2
     M^2}\log^2 \frac{1+r_t}{1-r_t} \right\}\:\:,
\end{eqnarray}
with $T_k \defeq |{\mathcal{F}}_k|$ and $M\defeq \min_{k} \max_j |\rado_{jk}|$.
For any $a>0$, let
\begin{eqnarray}
f_a(z) & \defeq & \frac{1}{az^2}\cdot \left(\log
\frac{1}{1-z^2} - a \cdot \log^2 \frac{1+z}{1-z}\right)-1\:\:.\nonumber
\end{eqnarray}
It satisfies
\begin{eqnarray}
f_a(z) & \approx_0 & \left(\frac{1}{a} - 5\right) + \left(\frac{1}{2a}
  - \frac{8}{3}
\right)\cdot z^2 \nonumber\\
 & & + \left(\frac{1}{3a}
  - \frac{92}{45}
\right)\cdot z^4 + o(z^4)\:\:. 
\end{eqnarray}
Since $f_a(z)$ is continuous for any $a\neq 0$, $\forall 0 < a < 1/5$, $\exists z_*(a) > 0$ such that $f_a(z) \geq 0,
\forall z \in [0, z_*]$. So, for any such $a< 1/5$ and any $\upomega$
satisfying $\upomega < (2 a M^2)/
(T_k \uplambda_{\Gamma})$, as long as each $r_t \leq z_*(a)$, we shall obtain
\begin{eqnarray}
\Lambda_k & \geq & a \sum_{t : \iota(t) \in {\mathcal{F}}_k} r_t^2\:\:.
\end{eqnarray}
There remains to tune $\upgamma \leq z_*(a)$, and remark that if we fix $a = 1/7$, then numerical calculations reveal that $z_*(a) >
0.98$, and if $a = 1/10$ then numerical calculations give $z_*(a) >
0.999$, completing the statement of Theorem \ref{thadarL2}.

\subsection{Proof of Theorem \ref{thadarINF}}\label{proof_thm_thadarINF}

We consider the case $\Omega(.) = \|.\|_\infty$, from which we shall
derive the case $\Omega(.) = \|.\|_1$.
We proceed as in the proof of Theorem \ref{thadarL2}, with the main
change that we have now $\delta_t =
\upomega\cdot(\|\ve{\theta}_t\|_\infty-\|\ve{\theta}_{t-1}\|_\infty)$,
so in place of $\Lambda_k$ in ineq . (\ref{eqdelta1}) we have to use,
letting $k_*$ any feature that gives the $\ell_\infty$ norm,
\begin{eqnarray}
\Lambda_{k} & \defeq & \left\{
\begin{array}{lcl}
\begin{array}{c}
\sum_{t : \iota(t) \in {\mathcal{F}}_{k}}  \log
\frac{1}{1-r_t^2}\\
 - \frac{\upomega}{\rado_{*{k}}}\left|\sum_{t : \iota(t) \in
     {\mathcal{F}}_{k}} \log \frac{1+r_t}{1-r_t}\right|
\end{array} & \mbox{ if } & k = k_*\\
\sum_{t : \iota(t) \in {\mathcal{F}}_{k}}  \log
\frac{1}{1-r_t^2} & \multicolumn{2}{l}{\mbox{ otherwise}} 
\end{array} \right.\:\:.\label{deflambda1}
\end{eqnarray}
It also comes
\begin{eqnarray}
\lefteqn{\Lambda_{k_*}}\nonumber\\
 & \geq & \sum_{t : \iota(t) \in {\mathcal{F}}_{k_*}}  \left\{\log
\frac{1}{1-r_t^2} - \frac{\upomega}{\rado_{*{k_*}}} \log \frac{1+|r_t|}{1-|r_t|}\right\}\nonumber\\
 & \geq & \sum_{t : \iota(t) \in {\mathcal{F}}_{k_*}}  \left\{\log
\frac{1}{1-r_t^2} - \frac{\upomega}{M} \log \frac{1+|r_t|}{1-|r_t|}\right\}\:\:, \label{defdeltainf}
\end{eqnarray}
with $M\defeq \min_{k} \max_j |\rado_{jk}|$.
Let us analyze $\Lambda_{k_*}$ and define for any $b>0$
\begin{eqnarray}
g_b(z) & \defeq & \log \frac{1}{1-z^2} - b \cdot
\log\frac{1+z}{1-z}\nonumber\\
 & & -\left( -2bz +z^2 - \frac{2bz^3}{3}\right)\label{defg}\:\:.
\end{eqnarray}
Inspecting $g_b$ shows that $g_b(0) = 0$, $g'_b(0) = 0$ and $g_b(z)$
is convex over $[0,1)$ for any $b\leq 3$, which shows that $g_b(z)
\geq 0$, $\forall z \in [0,1)$, $\forall b \leq 3$, and so, after
dividing by $bz^2$ and reorganising, yields in these cases:
\begin{eqnarray}
\lefteqn{\frac{1}{bz^2}\cdot \left(\log \frac{1}{1-z^2} - b \cdot
\log\frac{1+z}{1-z}\right) -1}\nonumber\\
 & \geq & \left(-\frac{2}{z} +
\left(\frac{1}{b}-1\right) - \frac{2z}{3}\right)\:\:.\label{pp1}
\end{eqnarray} 
Hence, both functions being continuous on $(0,1)$, the function in the left-hand side zeroes before the one in the
right-hand side (when this one does on $(0,1)$). The zeroes of the polynomial
\begin{eqnarray}
p_b(z) & \defeq & -\frac{2z^2}{3} + \left(\frac{1}{b}-1\right)z - 2
\end{eqnarray}
exist iff $b \leq \sqrt{3}/(4+\sqrt{3})$, in which case any
$z\in [0,1)$ must
satisfy
\begin{eqnarray}
z & \geq & \frac{3}{4}\cdot\left( \frac{1}{b} - 1 -
  \sqrt{\left(\frac{1}{b} - 1\right)^2 - \frac{16}{3}}\right) \label{condz}
\end{eqnarray}
to guarantee that $p_b(z)\geq 0$. Whenever this happens, we shall have
from (\ref{pp1}):
\begin{eqnarray}
\log \frac{1}{1-z^2} - b \cdot
\log\frac{1+z}{1-z} & \geq & b z^2\:\:. \label{condinf}
\end{eqnarray}
Since \omegaweak~is a $\gwl$-weak learner, if
we can guarantee that the right-hand side of ineq. (\ref{condz}) is no
more than $\gwl$, then there is nothing more to require from the
weak learner to have ineq. (\ref{condinf}) --- and therefore to have
$\Lambda_{k_*} \geq b \gwl^2 \cdot |{\mathcal{F}}_{k_*}|$. This yields
equivalently the following constraint on $b$:
\begin{eqnarray}
b & \leq & \frac{\frac{8\gwl}{3}}{\frac{16\gwl^2}{9} +
  \frac{8\gwl}{3} + \frac{16}{3}}\:\:. \label{consta}
\end{eqnarray}
Since $\gwl \leq 1$, ineq (\ref{consta}) ensured as long as
\begin{eqnarray}
b & \leq &  \frac{\frac{8\gwl}{3}}{\frac{16}{9} +
  \frac{8}{3} + \frac{16}{3}} = \frac{3\gwl}{11}\:\:,\label{constb}
\end{eqnarray}
which also guarantees $b \leq \sqrt{3}/(4+\sqrt{3})$. 
So, letting $T_* \defeq |{\mathcal{F}}_{k_*}|$ and recollecting 
\begin{eqnarray}
b & \defeq &
\frac{\upomega}{\min_{k} \max_j
  |\rado_{jk}|}\label{defbb}
\end{eqnarray} 
from eq. (\ref{defdeltainf}), we obtain from ineqs
  (\ref{defdeltainf}) and (\ref{condinf}):
\begin{eqnarray}
\Lambda_{k_*} & \geq & \frac{\upomega T_* \gwl^2}{\min_{k} \max_j
  |\rado_{jk}|}\:\:.
\end{eqnarray}
We need to ensure $\upomega \leq 3 \min_{k} \max_j
  |\rado_{jk}|\gwl /11$ from ineq
. (\ref{constb}), which holds if we pick it according to
eq. (\ref{bomegaLINF}). In this case, we finally obtain 
\begin{eqnarray}
\Lambda_{k_*} & \geq & (a \gwl  T_*) \cdot \gwl^2\:\:.
\end{eqnarray}
Now, since $\log(1/(1-x^2)) \geq x^2$, we also
have for $k\neq k_*$ in eq. (\ref{deflambda1}), 
\begin{eqnarray}
\Lambda_k & = & \sum_{t : \iota(t) \in {\mathcal{F}}_{k}}  \log
\frac{1}{1-r_t^2}\nonumber\\
 & \geq & \sum_{t : \iota(t) \in {\mathcal{F}}_{k}} r_t^2 \nonumber\\
 & \geq &
 |{\mathcal{F}}_{k}| \gwl^2\:\:, \forall k \neq k_*\:\:.
\end{eqnarray}
So, we finally obtain from
eq. (\ref{bz1}) and ineq. (\ref{eqdelta1}),
\begin{eqnarray}
\ell^{\tiny{\mathrm{exp}}}_{\lab{r}}({\mathcal{S}}_r,
  \ve{\theta}, \|.\|_2^2)  & \leq & \exp\left(-\frac{\tilde{T} \gwl^2}{2}\right)\:\:,\label{bzinf}
\end{eqnarray}
with $\tilde{T} \defeq (T - T_*) + a \gwl \cdot T_*$, as claimed when
$\Omega(.) = \|.\|_\infty$. The
case $\Omega = \|.\|_1$ follows form the fact that all $\Lambda_k$
match the bound of $\Lambda_{k_*}$.

\subsection{Proof of Theorem \ref{thadarSLOPE}}\label{proof_thm_thadarSLOPE}

We use the proof of Theorem \ref{thadarINF}, since when $\Omega(.) =
\|.\|_\Phi$, eq. (\ref{deflambda1}) becomes 
\begin{eqnarray}
\Lambda_{k} & \defeq & 
\sum_{t : \iota(t) \in {\mathcal{F}}_{k}}  \log
\frac{1}{1-r_t^2}\\
 & & - \frac{\xi_k}{\rado_{*{k}}}\left|\sum_{t : \iota(t) \in
     {\mathcal{F}}_{k}} \log \frac{1+r_t}{1-r_t}\right| \nonumber\\
 & \geq & \sum_{t : \iota(t) \in {\mathcal{F}}_{k}}  \left\{\log
\frac{1}{1-r_t^2} - \frac{\xi_k}{\max_j |\rado_{jk}|} \log \frac{1+|r_t|}{1-|r_t|}\right\}\:\:,\label{deflambda2}
\end{eqnarray}
assuming without loss of generality that the classifier at iteration
$T$, $\ve{\theta}_T$,
satisfies $|\theta_{Tk}| \geq |\theta_{T(k+1)}|$
for $k = 1, 2, ..., d-1$. We recall that $\xi_k \defeq \Phi^{-1}(1-kq/(2d))$ where
$\Phi^{-1}(.)$ is the quantile of the standard normal distribution and
$q\in (0,1)$ is the user-fixed $q$-value. The constraint $b\leq
3\gwl/11$ from ineq. (\ref{constb}) now has to hold with 
\begin{eqnarray}
b = b_k & \defeq &
\frac{\xi_k}{\max_j
  |\rado_{jk}|}\label{defbb}\:\:.
\end{eqnarray} 
Now, fix
\begin{eqnarray}
a & \defeq & \min \left\{\frac{3 \gwl}{11}, \frac{\Phi^{-1}(1-q/(2d))}{
    \min_ k \max_j
  |\rado_{jk}|}\right\}\label{defaa}\:\:.
\end{eqnarray}
Remark that
\begin{eqnarray}
\xi_k & \defeq &  \Phi^{-1}\left(1-\frac{kq}{2d}\right)\nonumber\\
 & \geq & \Phi^{-1}\left(1-\frac{q}{2d}\right) \nonumber\\
 & \geq & a \min_{k'} \max_j
  |\rado_{jk'}|\label{propxik}\:\:.
\end{eqnarray}
Suppose $q$ is chosen such that 
\begin{eqnarray}
\xi_k & \leq & \frac{3 \gwl}{11} \cdot \max_j
  |\rado_{jk}|\:\:, \forall k \in [d]\:\:.\label{defint}
\end{eqnarray}
This ensures $b_k \leq
3\gwl/11$ ($\forall k \in [d]$) in ineq. (\ref{constb}), while ineq. (\ref{propxik}) ensures
\begin{eqnarray}
\Lambda_{k} & \geq & b_k \sum_{t : \iota(t) \in {\mathcal{F}}_{k}}
r^2_t\label{ddf1}\\
& \geq & \frac{\xi_k}{\min_{k'}\max_j
  |\rado_{jk'}|}\cdot  \sum_{t : \iota(t) \in {\mathcal{F}}_{k}}
r^2_t\label{ddf3}\\
 & \geq & a |{\mathcal{F}}_{k}| \gwl^2\label{ddf2}\:\:.
\end{eqnarray}
Ineq. (\ref{ddf1}) holds because of ineqs (\ref{deflambda2}) and (\ref{condinf}). Ineq. (\ref{ddf2}) holds because of the weak
learning assumption and ineq. (\ref{defint}).
So, we obtain, under the weak learning assumption,
\begin{eqnarray}
\ell^{\tiny{\mathrm{exp}}}_{\lab{r}}({\mathcal{S}}_r,
  \ve{\theta}, \|.\|_\Phi)  & \leq & \exp\left(-\frac{a T \gwl^2}{2}\right)\:\:.\label{bzinfSLOPE}
\end{eqnarray}
Ensuring ineq. (\ref{defint}) is done if, after replacing $\xi_k$ by
its expression and reorganising, we can ensure
\begin{eqnarray}
q & \geq & 2\cdot \max_k \frac{q_{N,k}}{q_{D,k}}\:\:,
\end{eqnarray}
with
\begin{eqnarray}
(0,1) \ni q_{N,k} & \defeq & 1 - \Phi\left(\frac{3\gwl}{11}\cdot \max_j
  |\rado_{jk}|\right)\label{defqn}\:\:,\\
(0,1] \ni q_{D,k} & \defeq & \frac{k}{d} \label{defqd}\:\:.\\
\end{eqnarray}

\subsection{Proof of Theorem \ref{thdp1}}\label{proof_thm_thdp1}

Suppose wlog that the example index on which ${\mathcal{S}}_{\lab{e}}$ and
${\mathcal{S}}'_{\lab{e}}$ differ is $m$, and let $\ve{e}_m$ and
$\ve{e}'_m$ denote the two distinct edge vectors of the neighbouring datasets.
For $n= 1$, let $\ve{\rado}$ denote a rado created from first
picking uniformly at random ${\mathcal{I}} \in 2^m$ and then using \dprad~on the
singleton ${\mathcal{S}}_{\lab{r}} \defeq
\{\ve{\rado}_{{\mathcal{I}}}\}$ with:
\begin{eqnarray}
\ve{\rado}_{{\mathcal{I}}} & \defeq & \sum_{i \in {\mathcal{I}}} y_i
\cdot \ve{x}_i\:\:.
\end{eqnarray}
Let $a({\mathcal{S}}_{\lab{e}}) \defeq \sum_{{\mathcal{I}}'\subseteq [m-1]}
\mu(\ve{\rado}
  | \ve{\rado}_1 = \ve{\rado}_{\mathcal{I}'}, {\mathcal{S}} = {\mathcal{S}}_{\lab{e}})$, where
  $\mu(\ve{\rado}
  | .)$ is the density of the singleton output of \dprad, and
  $b({\mathcal{S}}_{\lab{e}}) \defeq \sum_{{\mathcal{I}}'\subseteq
    [m], m \in {\mathcal{I}}}
\mu(\ve{\rado}
  | \ve{\rado}_1 = \ve{\rado}_{\mathcal{I}'}, {\mathcal{S}} =
  {\mathcal{S}}_{\lab{e}})$. We have:
\begin{eqnarray}
\frac{\mu(\ve{\rado} |
  {\mathcal{S}}_{\lab{e}})}{\mu(\ve{\rado} |
  {\mathcal{S}}'_{\lab{e}})} & = &
\frac{a({\mathcal{S}}_{\lab{e}}) +
  b({\mathcal{S}}_{\lab{e}})}{a({\mathcal{S}}'_{\lab{e}}) +
  b({\mathcal{S}}'_{\lab{e}})}\nonumber\\
& = &
\frac{a({\mathcal{S}}_{\lab{e}}) + b({\mathcal{S}}_{\lab{e}})}{a({\mathcal{S}}_{\lab{e}}) + b({\mathcal{S}}'_{\lab{e}})}\label{feq1}\\
& \leq &
\max\left\{\frac{b({\mathcal{S}}_{\lab{e}})}{b({\mathcal{S}}'_{\lab{e}})}, \frac{b({\mathcal{S}}'_{\lab{e}})}{b({\mathcal{S}}_{\lab{e}})}\right\}\label{feq2}\:\:.
\end{eqnarray}
Eq. (\ref{feq1}) follows from the fact that when ${\mathcal{I}}' \subseteq
  [m-1]$, $\mu(\ve{\rado}
  | \ve{\rado}_1 = \ve{\rado}_{\mathcal{I}'}, {\mathcal{S}} =
  {\mathcal{S}}_{\lab{e}})= \mu(\ve{\rado}
  | \ve{\rado}_1 = \ve{\rado}_{\mathcal{I}'}, {\mathcal{S}} =
  {\mathcal{S}}'_{\lab{e}})$. Now, for any \textit{fixed} ${\mathcal{I}}'\subseteq
    [m]$ such that $m \in {\mathcal{I}}'$, we have
\begin{eqnarray}
\lefteqn{\mu(\ve{\rado}
  | \ve{\rado}_1 = \ve{\rado}_{\mathcal{I}'}, {\mathcal{S}} =
  {\mathcal{S}}_{\lab{e}})}\nonumber\\
 & \defeq & \left(\frac{\varepsilon}{2r_{\lab{e}}}\right)^d
 \exp(-\varepsilon \cdot \|\ve{\rado}_{\mathcal{I}'} - \ve{\rado}\|_1
 / r_{\lab{e}}) \nonumber\\
 & = & \left(\frac{\varepsilon}{2r_{\lab{e}}}\right)^d
 \exp(-\varepsilon \cdot \|\ve{\rado}'_{\mathcal{I}'} -
 \ve{\rado} + \ve{e}_m -\ve{e}'_m\|_1 / r_{\lab{e}})\label{defppi}\\
 & \leq & \underbrace{\left(\frac{\varepsilon}{2r_{\lab{e}}}\right)^d
 \exp \left(-\frac{\varepsilon}{r_{\lab{e}}} \cdot \|\ve{\rado}'_{\mathcal{I}'} -
 \ve{\rado} \|_1\right)}_{\defeq \mu(\ve{\rado}
  | \ve{\rado}_1 = \ve{\rado}_{\mathcal{I}'}, {\mathcal{S}} =
  {\mathcal{S}}'_{\lab{e}})}  \nonumber\\
 & & \cdot \exp \left(\frac{\varepsilon}{r_{\lab{e}}} \cdot \|\ve{e}_m -\ve{e}'_m\|_1\right)\nonumber\\
 & & = \exp \left(\frac{\varepsilon}{r_{\lab{e}}} \cdot \|\ve{e}_m
   -\ve{e}'_m\|_1\right)\nonumber\\
 & & \cdot \mu(\ve{\rado}
  | \ve{\rado}_1 = \ve{\rado}_{\mathcal{I}'}, {\mathcal{S}} =
  {\mathcal{S}}'_{\lab{e}})\nonumber\:\:.
\end{eqnarray}
where $\ve{\rado}'_{\mathcal{I}'} \defeq \ve{\rado}_{\mathcal{I}'} -
\ve{e}_m +\ve{e}'_m$ in eq. (\ref{defppi}) is the rado that is created from the same
$\mathcal{I}'$ but using  ${\mathcal{S}}'_{\lab{e}}$ and its
potentially different
example $\ve{e}'_m$. The inequality
holds because of the triangle inequality. Since $\|\ve{e}_m
-\ve{e}'_m\|_1 \leq r_{\lab{e}}$ by assumption, we get
$\mu(\ve{\rado}
  | \ve{\rado}_1 = \ve{\rado}_{\mathcal{I}'}, {\mathcal{S}} =
  {\mathcal{S}}_{\lab{e}}) \leq \exp(\varepsilon)\cdot \mu(\ve{\rado}
  | \ve{\rado}_1 = \ve{\rado}_{\mathcal{I}'}, {\mathcal{S}} =
  {\mathcal{S}}'_{\lab{e}})$, and so, summing over all ${\mathcal{I}}'\subseteq
    [m]$ such that $m \in {\mathcal{I}}'$, we get
$b({\mathcal{S}}_{\lab{e}})/ b({\mathcal{S}}'_{\lab{e}}) \leq
\exp(\varepsilon)$. Furthermore, we also have by symmetry
$b({\mathcal{S}}'_{\lab{e}})/ b({\mathcal{S}}_{\lab{e}}) \leq
\exp(\varepsilon)$. So the delivery of one rado is
$\varepsilon$-differentially private. The composition Theorem
\citep{drTA} achieves the proof of the first point of Theorem
\ref{thdp1}. To prove the second point, we first define the
(unregularized) log-loss,
\begin{eqnarray}
\ell^{\tiny{\mathrm{log}}}_{\lab{e}}\left({\mathcal{S}}_{\lab{e}},
  \ve{\theta}\right) & \defeq & \frac{1}{m}\cdot \sum_{i}
\log\left(1+\exp\left(-\ve{\theta}^\top
    (y_i \cdot \ve{x}_i)\right)\right) \:\:.
\end{eqnarray}
We exploit the following
inequalities that hold for the log-loss and exp-rado loss:

\begin{eqnarray}
\lefteqn{\frac{1}{m} \cdot \log \ell^{\tiny{\mathrm{exp}}}_{\lab{r}}({\mathcal{S}}^{*,{\textsc{dp}}}_{\lab{r}},
  \ve{\theta})}\nonumber\\
 & = &  \frac{1}{m} \cdot \log \frac{1}{2^m} \sum_{\ve{\sigma} \in \Sigma_m}
 \exp\left(-\ve{\theta}^\top \left(\ve{\rado}_{\ve{\sigma}} +
     \ve{z}_{\ve{\sigma}}\right)\right) \nonumber\\
 & \leq & \frac{1}{m} \cdot \log \left( \left( \frac{1}{2^m} \sum_{\ve{\sigma} \in \Sigma_m}
 \exp\left(-\ve{\theta}^\top \ve{\rado}_{\ve{\sigma}} \right)\right) \cdot
\left( \sum_{\ve{\sigma} \in \Sigma_m}
 \exp\left(-\ve{\theta}^\top
     \ve{z}_{\ve{\sigma}}\right) \right)\right) \label{r1}\\
 &  & = \frac{1}{m} \cdot \log \frac{1}{2^m} \sum_{\ve{\sigma} \in \Sigma_m}
 \exp\left(-\ve{\theta}^\top \ve{\rado}_{\ve{\sigma}} \right) +  \frac{1}{m}
 \cdot \log  \sum_{\ve{\sigma} \in \Sigma_m}
 \exp\left(-\ve{\theta}^\top
     \ve{z}_{\ve{\sigma}}\right) \nonumber\\
 &  = & \log 2 + \frac{1}{m} \cdot \log \frac{1}{2^m} \sum_{\ve{\sigma} \in \Sigma_m}
 \exp\left(-\ve{\theta}^\top \ve{\rado}_{\ve{\sigma}} \right) +  \frac{1}{m}
 \cdot \log  \frac{1}{2^m} \sum_{\ve{\sigma} \in \Sigma_m}
 \exp\left(-\ve{\theta}^\top
     \ve{z}_{\ve{\sigma}}\right) \nonumber\\
 &  \leq & \log 2 + \frac{1}{m} \cdot \log \frac{1}{2^m} \sum_{\ve{\sigma} \in \Sigma_m}
 \exp\left(-\ve{\theta}^\top \ve{\rado}_{\ve{\sigma}} \right) +  \frac{1}{m}\max_{\ve{\sigma}}\ve{\theta}^\top
     \ve{z}_{\ve{\sigma}}\nonumber\\
 &  \leq & \log 2 + \frac{1}{m} \cdot \log \frac{1}{2^m} \sum_{\ve{\sigma} \in \Sigma_m}
 \exp\left(-\ve{\theta}^\top \ve{\rado}_{\ve{\sigma}} \right) +
 \frac{1}{m} \max_{\ve{\sigma}} \Omega^\star(\ve{z}_{\ve{\sigma}})
 \Omega(\ve{\theta}) \label{cs2}\\
 & & = \ell^{\tiny{\mathrm{log}}}_{\lab{e}}\left({\mathcal{S}}_{\lab{e}},
  \ve{\theta}\right) + \frac{1}{m} \max_{\ve{\sigma}} \Omega^\star(\ve{z}_{\ve{\sigma}})
 \Omega(\ve{\theta}) \label{cs3}\\
 &  = &
 \ell^{\tiny{\mathrm{log}}}_{\lab{e}}\left({\mathcal{S}}_{\lab{e}},
   \ve{\theta}, (1/m)\cdot \max_{\ve{\sigma}} \Omega^\star(\ve{z}_{\ve{\sigma}})\cdot\Omega\right)\:\:,\nonumber
\end{eqnarray}
where ineq. (\ref{r1}) comes from the fact that $\sum_i a_ib_i \leq
(\sum_i a_i)(\sum_i b_i)$ when all $a_i, b_i\geq 0$,
ineq. (\ref{cs2}) is Cauchy-Schwartz and eq. (\ref{cs3}) is Lemma 2 in
\citep{npfRO}.\\

\noindent \textbf{Remarks on $\varepsilon_a$}: let us explain why the protection
of examples using the same noise level as rados is conservative but in
fact necessary in the worst case, considering for
simplicity the protection of a single rado / example.
The proof of Theorem \ref{thdp1} exploits a conservative
upperbound for the likelihood ratio:
\begin{eqnarray}
\frac{\mu(\ve{\rado} |
  {\mathcal{S}}_{\lab{e}})}{\mu(\ve{\rado} |
  {\mathcal{S}}'_{\lab{e}})} =
\frac{a({\mathcal{S}}_{\lab{e}}) +
  b({\mathcal{S}}_{\lab{e}})}{a({\mathcal{S}}_{\lab{e}}) +
  b({\mathcal{S}}'_{\lab{e}})} \leq 
\max\left\{\frac{b({\mathcal{S}}_{\lab{e}})}{b({\mathcal{S}}'_{\lab{e}})}, \frac{b({\mathcal{S}}'_{\lab{e}})}{b({\mathcal{S}}_{\lab{e}})}\right\}\label{dpratio1}\:\:,
\end{eqnarray}
and then upperbounds the $\max$ by $\exp \varepsilon$ to get the DP
requirement. The same strategy can be used to protect the example, but
the bound is sometimes more conservative in this
case. Indeed, whereas one examples participates in generating half the total number
of DP rados, one example participates in only $1/m$ of the generation of
DP examples. For a single DP example $\ve{e}$, the equality in (\ref{dpratio1})
becomes $\mu(\ve{e} |
  {\mathcal{S}}_{\lab{e}}) / \mu(\ve{e} |
  {\mathcal{S}}'_{\lab{e}}) = (a'({\mathcal{S}}_{\lab{e}}) +
  b'({\mathcal{S}}_{\lab{e}})/(a'({\mathcal{S}}_{\lab{e}}) +
  b'({\mathcal{S}}'_{\lab{e}})$ with $a'({\mathcal{S}}_{\lab{e}})
  \defeq (1-(1/m)) \cdot \mu(\ve{e} |
  {\mathcal{S}}_{\lab{e}}\backslash \{\ve{e}_m\})$ and:
\begin{eqnarray}
b'({\mathcal{S}}_{\lab{e}}) \defeq \frac{1}{m} \cdot \mu(\ve{e} |
  \{\ve{e}_m\}) \:\:, b'({\mathcal{S}}'_{\lab{e}}) \defeq \frac{1}{m}
  \cdot \mu(\ve{e} |
  \{\ve{e}'_m\})\:\:.
\end{eqnarray}
Let $u \defeq \mu(\ve{e} |
  \{\ve{e}'_m\}) / \mu(\ve{e} |
  {\mathcal{S}}'_{\lab{e}}\backslash \{\ve{e}'_m\})$ ($=\mu(\ve{e} |
  \{\ve{e}'_m\}) / \mu(\ve{e} |
  {\mathcal{S}}_{\lab{e}}\backslash \{\ve{e}_m\})$). If we use the
  same amount of protection as for one rado, then we get
\begin{eqnarray}
\frac{\mu(\ve{e} |
  {\mathcal{S}}_{\lab{e}})}{\mu(\ve{e} |
  {\mathcal{S}}'_{\lab{e}})} & \leq & f_u(\varepsilon)\:\:,\label{eqp}
\end{eqnarray}
where $\varepsilon$ is the rado privacy budget and 
\begin{eqnarray}
f_u(\varepsilon) & \defeq & \frac{(m-1) + u \exp(\varepsilon)}{m-1 +
  u}\:\:.
\end{eqnarray}
$f_u(\varepsilon)$ is always $<\exp(\varepsilon)$, so if we use this
$\exp(\varepsilon)$ bound to pick the noise level, then 
we are in fact putting more protection over examples than necessary (remember that the
protection is also conservative for rados, but to a lesser
extent). However, this choice would not be so bad
in the worst case since $\lim_{u\rightarrow \infty}
f_u(\varepsilon) = \exp(\varepsilon)$. To summarise, without
constraints on $u$, and to be sure to meet the DP requirements in any case, we would err on the conservative side, as we did
for rados in ineq. (\ref{feq2}), and pick
$\varepsilon_{\lab{e}}=\varepsilon$, \textit{i.e.} the same
amount of noise for examples. Yet, as we explain in the body of the paper, the results are
exceedingly in favor of \radaboostNoL~in this case. To give a more
balanced picture, we chose to compute an ``approximate'' privacy budget
$\varepsilon_{\lab{e}} = \varepsilon_a$ for $n$ examples, which we simply
fix to be $\varepsilon_a \defeq n\cdot \ln(f_{u\defeq
  1}(\varepsilon/n))$ ($< \varepsilon$) where $\varepsilon$ is the
privacy budget for $n$ rados. So, we have
\begin{eqnarray}
\varepsilon_a & = & n \cdot \ln\left(1+\frac{\exp(\varepsilon/n)-1}{m}\right)\label{defepse}\:\:.
\end{eqnarray}
Again, when $u>1$, fixing $\varepsilon = \varepsilon_a$ to protect
examples would fail to achieve $\varepsilon$-differential privacy.

Nevertheless, one can reasonably consider that the ``optimal'' differentially private picture
of \adaboostSS~shall thus be representable as a stretching of its curves in
between the figures for $\varepsilon_a$ and $\varepsilon$. 

\section{Additional experiments}\label{exp_expes}

\subsection{Supports for rados (complement to Table \ref{tc1_errs_rr})}\label{exp_compl_tc1_errs_rr}

\begin{sidewaystable}[t]
\begin{center}
{\small
\begin{tabular}{|crr||r|r||r|r|r|r|r|}
\hline \hline
 & & &   \adaboostSS  & $\ell_1$-\adaboostSS &
 \multicolumn{5}{c|}{{\scriptsize \radaboostNoL}} \\ 
 & & & & & \multicolumn{1}{c|}{$\upomega = 0$} & \multicolumn{1}{c|}{$\Omega = \|.\|_{\matrice{\tiny{i}}_d}^2$} & \multicolumn{1}{c|}{$\Omega = \|.\|_1$} &
 \multicolumn{1}{c|}{$\Omega = \|.\|_\infty$} & \multicolumn{1}{c|}{$\Omega = \|.\|_{\Phi}$}\\
domain & $m$ & $d$ & \multicolumn{1}{c||}{supp.$\pm\sigma$} & \multicolumn{1}{c||}{supp.$\pm\sigma$} &
\multicolumn{1}{c|}{supp.$\pm\sigma$} &
\multicolumn{1}{c|}{supp.$\pm\sigma$} &
\multicolumn{1}{c|}{supp.$\pm\sigma$} &
\multicolumn{1}{c|}{supp.$\pm\sigma$} & 
\multicolumn{1}{c|}{supp.$\pm\sigma$}\\ \hline
Fertility & 100 & 9 & 36.67$\pm$36.3 & 14.44$\pm$5.36 & 37.78$\pm$31.1 &
36.67$\pm$34.8 & $\bullet$ 42.22$\pm$31.0 & $\circ$ 24.44$\pm$22.1 &
32.22$\pm$17.7\\
Sonar & 208 & 60 & 57.83$\pm$3.69 & 1.83$\pm$0.52 & $\bullet$ 14.17$\pm$3.62 & 14.00$\pm$3.16 &
13.67$\pm$3.99 & 12.83$\pm$4.45 & $\circ$ 12.67$\pm$4.17\\
Haberman & 306 & 3 & 70.00$\pm$10.6 & 33.33$\pm$0.00 & $\bullet$ 66.67$\pm$22.2 &
$\bullet$ 66.67$\pm$15.7 & 56.67$\pm$16.1 & 60.00$\pm$26.3 &
$\circ$ 50.00$\pm$17.6\\
Ionosphere & 351 & 33 & 76.97$\pm$8.23 & 3.64$\pm$1.27 & $\bullet$ 13.64$\pm$3.85 &
13.03$\pm$3.51 & 11.82$\pm$3.63 & $\circ$ 11.21$\pm$4.75 & $\circ$ 11.21$\pm$4.53\\
Breastwisc & 699 & 9 & 90.00$\pm$3.51 & 11.11$\pm$0.00 & 51.11$\pm$12.0 &
84.44$\pm$7.77 & $\circ$ 48.89$\pm$9.37 & 84.44$\pm$10.7 & $\bullet$ 86.67$\pm$4.68\\
Transfusion & 748 & 4 & 77.50$\pm$14.2 & 25.00$\pm$0.00 & $\circ$ 67.50$\pm$20.6 &
70.00$\pm$23.0 & $\circ$ 67.50$\pm$16.9 & $\circ$ 67.50$\pm$23.7 &
$\bullet$ 72.50$\pm$14.2\\
Banknote & 1372 & 4 & 100.00$\pm$0.00 & 25.00$\pm$0.00 & $\circ$ 40.00$\pm$12.9 &
$\bullet$ 50.00$\pm$0.00 & 47.50$\pm$7.91 & $\bullet$ 50.00$\pm$0.00 & 47.50$\pm$7.91\\
Winered & 1599 & 11 & 79.09$\pm$6.14 & 9.09$\pm$0.00 & $\circ$ 25.45$\pm$5.75 & $\bullet$ 27.27$\pm$6.06
& $\bullet$ 27.27$\pm$7.42 & $\circ$ 25.45$\pm$7.17 & $\bullet$ 27.27$\pm$7.42\\
Abalone & 4177 & 10 & 64.00$\pm$6.99 & 19.00$\pm$3.16 & $\bullet$ 30.00$\pm$6.67 & $\circ$ 10.00$\pm$0.00
& 12.00$\pm$6.32 & $\circ$ 10.00$\pm$0.00 & 11.00$\pm$3.16\\
Winewhite & 4898 & 11 & 66.36$\pm$9.63 & 9.09$\pm$0.00 & $\bullet$ 28.18$\pm$2.87 &
$\bullet$ 28.18$\pm$2.87 & 20.91$\pm$4.39 & 27.27$\pm$0.00 & $\circ$
18.18$\pm$0.00\\
Smartphone & 7352 & 561 & 5.53$\pm$0.24 & 0.36$\pm$0.00 & $\circ$ 0.18$\pm$0.00 & 71.21$\pm$20.1 & $\circ$ 0.18$\pm$0.00 & $\bullet$ 74.72$\pm$19.7 & 24.69$\pm$9.87\\
Firmteacher & 10800 & 16 & 48.12$\pm$30.8 & 10.00$\pm$3.22 & 24.38$\pm$7.48 & $\bullet$ 25.62$\pm$9.52 & 21.25$\pm$4.37 & $\circ$ 20.62$\pm$4.22 & $\circ$ 20.62$\pm$9.34\\
Eeg & 14980 & 14 & 14.29$\pm$3.37 & 8.57$\pm$3.01 & $\bullet$ 39.29$\pm$13.2 & $\circ$ 38.57$\pm$9.04 &
$\bullet$ 39.29$\pm$14.0 & $\circ$ 38.57$\pm$13.1 & $\bullet$ 39.29$\pm$10.8\\
Magic & 19020 & 10 & 45.00$\pm$7.07 & 10.00$\pm$0.00 & $\circ$ 10.00$\pm$0.00 &
$\bullet$ 51.00$\pm$3.16 & $\circ$ 10.00$\pm$0.00 & 49.00$\pm$7.38 &
$\circ$ 10.00$\pm$0.00\\
Hardware & 28179 & 96 & 11.98$\pm$7.56 & 2.19$\pm$0.33 & $\circ$
1.04$\pm$0.00 & 20.94$\pm$3.12 & $\circ$ 1.04$\pm$0.00 & $\bullet$
22.08$\pm$1.89 & 21.25$\pm$1.49\\\Xhline{2\arrayrulewidth}
Marketing & 45211 & 27 & 65.19$\pm$3.58 & 7.40$\pm$0.00 &
7.41$\pm$0.00 &
12.96$\pm$3.60 & $\circ$ 3.70$\pm$0.00 & $\bullet$ 13.33$\pm$4.35 & $\circ$ 3.70$\pm$0.00\\
Kaggle & 120269 & 11 & 28.18$\pm$5.16 & 18.18$\pm$0.00 & $\bullet$ 17.27$\pm$2.87 & $\circ$ 9.09$\pm$0.00 & 15.45$\pm$4.39 & 10.00$\pm$2.87 & 14.55$\pm$4.69\\
\hline\hline
\end{tabular}
}
\end{center}
\caption{Supports of \adaboostSS~and $\ell_1$-\adaboostSS~vs \radaboostNoL~for the results
  displayed in Table  \ref{tc1_errs_rr} (supp.\%($\ve{\theta}$) $\defeq 100\cdot
  \|\ve{\theta}\|_0/d$). For each domain, the sparsest of \radaboostNoL's method (in
  average) is indicated with "$\circ$", and the least sparse is indicated with "$\bullet$".}
  \label{tc1_supp_rr}
\end{sidewaystable}

Table \ref{tc1_supp_rr} in this Appendix provides the
supports used to summarize Table  \ref{tc1_errs_rr}. 

\subsection{Experiments on \textit{class-wise} rados}\label{exp_compl_tc1}

Tables \ref{tc1_errs_cw} and \ref{tc1_supp_cw} provide the test errors
and supports for \radaboostNoL~when trained with class-wise rados,
that is, rados that sum examples of the same class. The experiments do
not display that class-wise rados allow for a better training of
\radaboostNoL, as test errors are on par with \radaboostNoL~trained
with plain random rados (see Table \ref{tc1_errs_rr}).

\begin{sidewaystable}[t]
\begin{center}
{\footnotesize
\begin{tabular}{||c|||crr||r||r|rr|rr|rr|rr|}\hline \hline
&  & & &   \adaboostSS$\wedge$  &
 \multicolumn{9}{c|}{{\scriptsize \radaboostNoL}} \\ 
 & & & &$\ell_1$-\adaboostSS & \multicolumn{1}{c|}{$\upomega = 0$} &  \multicolumn{2}{c|}{$\Omega = \|.\|_{\matrice{\tiny{i}}_d}^2$} & \multicolumn{2}{c|}{$\Omega = \|.\|_1$} &
\multicolumn{2}{c|}{$\Omega = \|.\|_\infty$}&
\multicolumn{2}{c|}{$\Omega = \|.\|_{\Phi}$}\\
 & domain & $m$ & $d$ & \multicolumn{1}{c||}{err$\pm\sigma$} & \multicolumn{1}{c|}{err$\pm\sigma$} & \multicolumn{1}{c}{err$\pm\sigma$}
& \multicolumn{1}{c|}{$\Delta$} & \multicolumn{1}{c}{err$\pm\sigma$}
& \multicolumn{1}{c|}{$\Delta$} & \multicolumn{1}{c}{err$\pm\sigma$}
& \multicolumn{1}{c|}{$\Delta$} & \multicolumn{1}{c}{err$\pm\sigma$}
& \multicolumn{1}{c|}{$\Delta$} \\ \hline
+ & Fertility & 100 & 9 & 40.00$\pm$14.1 &  48.00$\pm$16.1 &
42.00$\pm$9.19 & 10.00 & 47.00$\pm$12.5 & 7.00 & 42.00$\pm$13.2 &
10.00 & \cellcolor{Gray}37.00$\pm$10.6 & 16.00\\
- & Sonar & 208 & 60 & 24.57$\pm$9.11 &  27.86$\pm$11.4 &  25.52$\pm$7.29
& 5.74 & 26.40$\pm$7.80 & 3.90 & \cellcolor{Gray}24.98$\pm$10.6 & 4.40 &
25.02$\pm$9.05 & 4.38\\
- & Haberman & 306 & 3 & 25.15$\pm$6.53 & 26.76$\pm$6.92 &
\cellcolor{Gray}25.48$\pm$6.32 & 1.28 & 25.78$\pm$6.72 & 1.30 & 25.77$\pm$7.93 & 1.99
& 25.78$\pm$7.18 & 1.31\\
- & Ionosphere & 351 & 33 & 13.11$\pm$6.36 & 17.67$\pm$6.17 &
14.22$\pm$6.08 & 3.43 & 15.65$\pm$5.51 & 2.29 & \cellcolor{Gray}13.67$\pm$5.65 & 5.11
& 15.08$\pm$6.38 & 3.73\\
 - & Breastwisc & 699 & 9 & 3.00$\pm$1.96 & \cellcolor{Gray}3.43$\pm$1.93 &
3.57$\pm$2.54 & 0.86 & 3.57$\pm$2.15 & 1.29 & 3.72$\pm$2.44 & 1.14 &
\cellcolor{Gray}3.43$\pm$1.93 & 1.00\\
+ & Transfusion & 748 & 4 & 39.17$\pm$7.01 & 35.56$\pm$5.15 &
34.90$\pm$5.09 & 4.01 & 34.76$\pm$7.25 & 4.68 & \cellcolor{Gray}33.43$\pm$5.53 & 5.46
& 33.95$\pm$5.14 & 4.27\\
- & Banknote & 1372 & 4 & 2.70$\pm$1.46 & \cellcolor{Gray}13.70$\pm$2.30 &
13.78$\pm$3.48 & 0.73 & 13.92$\pm$3.16 & 1.46 & 13.78$\pm$3.73 & 1.09
& \cellcolor{Gray}13.70$\pm$3.83 & 1.31\\
+ & Winered & 1599 & 11 & 26.33$\pm$2.75 & 27.64$\pm$3.16 &
\cellcolor{Gray}26.27$\pm$2.11 & 2.06 & 27.39$\pm$2.86 & 1.12 & 26.64$\pm$3.12 & 1.56
& 27.64$\pm$3.34 & 0.69\\
+ & Abalone & 4177 & 10 & 22.98$\pm$2.70 & 24.59$\pm$2.65 &
24.18$\pm$2.51 & 0.00 & 24.11$\pm$2.39 & 0.48 & 24.18$\pm$2.51 & 0.19
& \cellcolor{Gray}24.08$\pm$2.67 & 0.26\\
+ & Winewhite & 4898 & 11 & 30.73$\pm$2.20 & 31.97$\pm$1.57 &
31.44$\pm$1.49 & 0.65 & \cellcolor{Gray}31.01$\pm$2.17 & 0.74 & 31.32$\pm$1.99 & 0.49
& 31.38$\pm$2.05 & 0.63\\
+ & Smartphone & 7352 & 561 & 0.00$\pm$0.00 & 0.67$\pm$0.25 & 
0.18$\pm$0.24 & 0.04 & 0.46$\pm$0.29 & 0.00 & 0.18$\pm$0.22 & 0.05 &
\cellcolor{Gray}0.16$\pm$0.22 & 0.05\\
- & Firmteacher & 10800 & 16 & 44.44$\pm$1.34 & \cellcolor{Gray}40.10$\pm$4.65 & 40.34$\pm$7.09 & 3.35 & 40.86$\pm$5.16 & 1.98 & 40.87$\pm$3.91 & 1.44 & 40.88$\pm$5.29 & 1.86\\
-& Eeg & 14980 & 14 & 45.38$\pm$2.04 & 44.79$\pm$1.62 &
44.47$\pm$1.49 & 0.69 & 44.45$\pm$1.27 & 0.51 & \cellcolor{Gray}44.41$\pm$1.96 & 0.49
& 44.54$\pm$1.41 & 0.45\\
+ & Magic & 19020 & 10 & 21.07$\pm$1.09 & 21.51$\pm$0.99 &
\cellcolor{Gray}21.38$\pm$1.17 & 0.24 & 26.41$\pm$1.08 & 0.00 &
21.42$\pm$0.99 & 0.18 & 25.84$\pm$1.94 & 0.57\\
+ & Hardware & 28179 & 96 & 16.77$\pm$0.73 & 8.85$\pm$0.68 & 6.39$\pm$0.71
& 0.20 & 9.06$\pm$3.76 & 2.04 & 5.78$\pm$1.85 & 0.71 &
\cellcolor{Gray}5.43$\pm$1.57 & 1.10\\\Xhline{2\arrayrulewidth}
- & Marketing & 45211 & 27 & 30.68$\pm$1.01 &  28.03$\pm$0.45 &
\cellcolor{Gray} 27.79$\pm$0.50 & 0.23 & 27.87$\pm$0.58 & 0.14 & 27.94$\pm$0.53 & 0.06 & 27.90$\pm$0.61 & 0.08\\
+ & Kaggle & 120269 & 11 & 47.80$\pm$0.47 & \cellcolor{Gray} 15.99$\pm$2.89 & 
16.88$\pm$0.51 & 0.02 & \cellcolor{Gray} 15.99$\pm$2.89 & 0.90 & 16.16$\pm$2.36 & 0.74
& 16.02$\pm$2.88 & 0.88\\
\hline\hline
\end{tabular}
}
\end{center}
\caption{Results of \adaboostSS~\citep{ssIBj} vs \radaboostNoL~(trained
  with random \textit{class-wise} rados). Conventions follow Table
  \ref{tc1_errs_rr}. On each domain, the leftmost column shows a "+"
  when \radaboostNoL~performs better when trained with class-wise
  rados (instead of just plain random rados), and "-" when it performs worse.}
  \label{tc1_errs_cw}
\end{sidewaystable}

\begin{sidewaystable}[t]
\begin{center}
{\footnotesize
\begin{tabular}{|crr||r|r||r|r|r|r|r|}
\hline \hline
 & & &   \adaboostSS  & $\ell_1$-\adaboostSS &
 \multicolumn{5}{c|}{{\scriptsize \radaboostNoL}} \\ 
 & & & & & \multicolumn{1}{c|}{$\upomega = 0$} & \multicolumn{1}{c|}{$\Omega = \|.\|_{\matrice{\tiny{i}}_d}^2$} & \multicolumn{1}{c|}{$\Omega = \|.\|_1$} &
 \multicolumn{1}{c|}{$\Omega = \|.\|_\infty$} & \multicolumn{1}{c|}{$\Omega = \|.\|_{\Phi}$}\\
domain & $m$ & $d$ & \multicolumn{1}{c||}{supp.$\pm\sigma$} & \multicolumn{1}{c||}{supp.$\pm\sigma$} &
\multicolumn{1}{c|}{supp.$\pm\sigma$} &
\multicolumn{1}{c|}{supp.$\pm\sigma$} &
\multicolumn{1}{c|}{supp.$\pm\sigma$} &
\multicolumn{1}{c|}{supp.$\pm\sigma$} & 
\multicolumn{1}{c|}{supp.$\pm\sigma$}\\ \hline
Fertility & 100 & 9 & 36.67$\pm$36.3 & 14.44$\pm$5.36 & 47.78$\pm$41.0 &
42.22$\pm$42.2 & 42.22$\pm$38.1 & 32.22$\pm$19.9 &
37.78$\pm$28.8\\
Sonar & 208 & 60 & 57.83$\pm$3.69 & 1.83$\pm$0.52 & 13.67$\pm$6.61 & 12.17$\pm$4.78 &
12.83$\pm$3.24 & 13.33$\pm$5.56 & 15.00$\pm$4.51\\
Haberman & 306 & 3 & 70.00$\pm$10.5 & 33.33$\pm$0.00 & 73.33$\pm$14.1 &
70.00$\pm$10.5 & 83.33$\pm$17.6 & 80.00$\pm$23.3 &
80.00$\pm$17.2\\
Ionosphere & 351 & 33 & 76.97$\pm$8.23 & 3.64$\pm$1.27 & 16.36$\pm$5.57 &
18.18$\pm$4.29 & 16.97$\pm$7.03 & 15.15$\pm$5.15 & 16.36$\pm$4.99\\
Breastwisc & 699 & 9 & 90.00$\pm$3.51 & 11.11$\pm$0.00 & 46.67$\pm$11.5 &
47.78$\pm$9.15 & 43.33$\pm$9.73 & 52.22$\pm$12.9 & 53.33$\pm$12.6\\
Transfusion & 748 & 4 & 77.50$\pm$14.2 & 25.00$\pm$0.00 & 67.50$\pm$20.6 &
82.50$\pm$16.9 & 67.50$\pm$12.1 & 82.50$\pm$12.1 &
75.00$\pm$16.7\\
Banknote & 1372 & 4 & 100.00$\pm$0.00 & 25.00$\pm$0.00 & 45.00$\pm$10.6 &
42.50$\pm$12.1 & 45.00$\pm$10.5 & 45.00$\pm$10.5 &
45.00$\pm$10.5\\
Winered & 1599 & 11 & 79.09$\pm$6.14 & 9.09$\pm$0.00 & 30.91$\pm$6.36 & 24.55$\pm$7.48
& 27.27$\pm$6.06 & 30.91$\pm$7.67 & 29.09$\pm$5.75\\
Abalone & 4177 & 10 & 64.00$\pm$6.99 & 19.00$\pm$3.16 & 57.00$\pm$10.6 &
10.00$\pm$0.00 & 14.00$\pm$5.16 & 10.00$\pm$0.00 & 13.00$\pm$4.83\\
Winewhite & 4898 & 11 & 66.36$\pm$9.63 & 9.09$\pm$0.00 & 20.00$\pm$3.83 &
19.09$\pm$2.87 & 20.91$\pm$6.14 & 20.91$\pm$4.39 & 19.09$\pm$2.87\\
Smartphone & 7352 & 561 & 5.53$\pm$0.24 & 0.36$\pm$0.00 & 0.18$\pm$0.00 &
30.78$\pm$19.7 & 0.18$\pm$0.00 & 29.86$\pm$20.5 & 30.09$\pm$13.3\\
Firmteacher & 10800 & 16 & 48.12$\pm$30.8 & 10.00$\pm$3.22 & 30.00$\pm$16.9 & 26.25$\pm$13.1 & 25.00$\pm$7.22 & 25.62$\pm$12.0 & 23.75$\pm$8.74\\
Eeg & 14980 & 14 & 14.29$\pm$3.37 & 8.57$\pm$3.01 & 38.57$\pm$7.68 & 42.14$\pm$4.05 &
40.00$\pm$3.69 & 42.86$\pm$0.00 & 39.29$\pm$3.76\\
Magic & 19020 & 10 & 45.00$\pm$7.07 & 10.00$\pm$0.00 & 20.00$\pm$0.00 & 38.00$\pm$4.22
& 10.00$\pm$0.00 & 32.00$\pm$4.22 & 11.00$\pm$3.16\\
Hardware & 28179 & 96 & 11.98$\pm$7.56 & 2.19$\pm$0.33 & 5.00$\pm$0.96 &
20.10$\pm$7.15 & 1.67$\pm$0.54 & 17.08$\pm$7.51 &
25.21$\pm$22.5\\\Xhline{2\arrayrulewidth}
Marketing & 45211 & 27 & 65.19$\pm$3.58 & 7.40$\pm$0.00 & 5.19$\pm$1.91 & 5.56$\pm$1.95 & 5.56$\pm$1.95 & 5.56$\pm$1.95 & 5.93$\pm$2.59\\
Kaggle & 120269 & 11 & 28.18$\pm$5.16 & 18.18$\pm$0.00 & 17.27$\pm$2.87 &
19.09$\pm$2.87 & 17.27$\pm$2.87 & 18.18$\pm$0.00 & 17.27$\pm$2.87\\
\hline\hline
\end{tabular}
}
\end{center}
\caption{Supports of \adaboostSS~vs \radaboostNoL~for the results
  displayed in Table  \ref{tc1_errs_cw} in this Appendix. Conventions follow Table \ref{tc1_supp_rr} in this
  Appendix.}
  \label{tc1_supp_cw}
\end{sidewaystable}

\subsection{Test errors and supports for rados (comparison last vs
  best empirical classifier)}\label{exp_compl_tc1_errs_supp_all}

In the paper's main experiments, the classifier kept out of the sequence, for both
\adaboostSS~and \radaboostNoL, is the best empirical classifier, that
is, the classifier which minimizes the
empirical risk out of the training sample. This setting makes sense if
the objective is just the minimization of the test error without any constraint, and it is
also applicable in a privacy setting where the data and the learner
are distant parties (in this case, the learner sends the sequence of classifiers
$\ve{\theta}_1, \ve{\theta}_2, ..., \ve{\theta}_T$ to the party
holding the data, which can then select the best in the
sequence). Yet, one may wonder how the
algorithms compare when the classifier returned is just the last one
in the sequence, that is, $\ve{\theta}_T$.

Tables \ref{tc1_errs_rr_all} and \ref{tc1_supp_rr_all} provide
errors and supports comparing the versions of \radaboostNoL~when the
best empirical classifier is selected ($^\star$), or when the last classifier in
the sequence is kept ($^\dagger$). They are therefore subsuming Tables
\ref{tc1_errs_rr} (for test errors) and \ref{tc1_supp_rr}
(for supports). 

The intuition tells that not
selecting the classifier in the sequence produced ($^\dagger$) should
produce either no better, or eventually worse results than when selecting the
classifier to keep from the sequence $\ve{\theta}_0, \ve{\theta}_1,
..., \ve{\theta}_T$. The results
display that it is the case, for both \adaboostSS~and \radaboostNoL,
and the phenomenon is more visible as the domain size increases. The
degradation for \radaboostNoL~appears to be significantly worse than that for
\adaboostSS~on three domains, Fertility, Firmteacher and Kaggle, since
not selecting the classifier using the training data incurs an increase
of 8$\%$ and more on the test error for these domains. However, for
the majority of the domains, the variation in test error does not
exceed 1$\%$, and on three domains (Winewhite, Smartphone, Eeg), the
absence of selection of the classifier actually does not increase the test
error at all. 

Therefore, even when not marginal, the fact that the
test error significantly increases only on a minority of the
domains for \radaboostNoL~calls for a rather domain-specific selection procedure
of the classifier in the sequence, rather than an all-purpose
selection procedure. Furthermore, on domains for which not
selecting the classifier produces the worst results, such a more efficient selection
procedure of the classifier might
actually be bypassed by a more careful \textit{crafting} of the rados, since
when class-wise random rados are used (results not shown), picking the
last classifier for domain Kaggle reduces the test error by
approximately 10$\%$
compared to random rados (the test error drops to 32.68$\pm$10.9 instead of 42.41$\pm$9.32 for
\slope). Such a specific crafting of rados is an interesting and non trivial problem that deserves
further attention.

\begin{sidewaystable}[t]
\begin{center}
{\scriptsize
\begin{tabular}{|crr||r||rr|rrr|rrr|rrr|rrr|}
\hline \hline
 & \hspace{-0.2cm}& \hspace{-0.4cm}&   \adaboostSS$\wedge$  &
 \multicolumn{14}{c|}{{\scriptsize \radaboostNoL}} \\ 
 & \hspace{-0.2cm}& \hspace{-0.4cm}& $\ell_1$-\adaboostSS\hspace{-0.19cm}& \multicolumn{2}{c|}{$\upomega = 0$} &  \multicolumn{3}{c|}{$\Omega = \|.\|_{\matrice{\tiny{i}}_d}^2$} & \multicolumn{3}{c|}{$\Omega = \|.\|_1$} &
\multicolumn{3}{c|}{$\Omega = \|.\|_\infty$}&
\multicolumn{3}{c|}{$\Omega = \|.\|_{\Phi}$}\\
domain & \hspace{-0.2cm}$m$ & \hspace{-0.4cm}$d$ &
\multicolumn{1}{c||}{err$\pm\sigma$ \hspace{-0.19cm} } & &
\multicolumn{1}{c|}{err$\pm\sigma$} & & \multicolumn{1}{c}{err$\pm\sigma$}
& \multicolumn{1}{c|}{$\Delta$} & & \multicolumn{1}{c}{err$\pm\sigma$}
& \multicolumn{1}{c|}{$\Delta$} & & \multicolumn{1}{c}{err$\pm\sigma$}
& \multicolumn{1}{c|}{$\Delta$} & & \multicolumn{1}{c}{err$\pm\sigma$}
& \multicolumn{1}{c|}{$\Delta$} \\ \hline
\hspace{-0.27cm} Fertility$^\dagger$ & \hspace{-0.49cm} 100 & \hspace{-0.4cm} 9
&  48.00$\pm$15.5 \hspace{-0.19cm} & & 49.00$\pm$13.7 & & 49.00$\pm$14.5 &
4.00 & & \cellcolor{Gray}48.00$\pm$16.9 & 5.00 & &
\cellcolor{Gray}48.00$\pm$11.4 & 7.00 & & 50.00$\pm$13.3 & 3.00\\
\hspace{-0.27cm} Fertility$^\star$ & \hspace{-0.49cm} 100 & \hspace{-0.4cm} 9 &  40.00$\pm$14.1
\hspace{-0.19cm} & & 40.00$\pm$14.9 &
& 41.00$\pm$16.6 & 8.00 &  &
41.00$\pm$14.5 & 4.00 &  & 41.00$\pm$21.3 &
6.00 & & \cellcolor{Gray}38.00$\pm$14.0 & 7.00\\\hline
\hspace{-0.27cm} Sonar$^\dagger$ & \hspace{-0.49cm} 208 &
\hspace{-0.4cm} 60 &
 24.02$\pm$5.43 \hspace{-0.19cm} & & 29.38$\pm$8.45 & & \cellcolor{Gray}24.50$\pm$7.53 &
6.81 & & \cellcolor{Gray}24.50$\pm$9.10 & 6.81 & & 26.00$\pm$5.83 & 7.21 & & 26.45$\pm$7.76 & 1.93\\
\hspace{-0.27cm} Sonar$^\star$ & \hspace{-0.49cm} 208 & \hspace{-0.4cm} 60 & 24.57$\pm$9.11
\hspace{-0.19cm} &  & 27.88$\pm$4.33 & & 25.05$\pm$7.56 &
8.14 & & \cellcolor{Gray}24.05$\pm$8.41 & 4.83 & & 24.52$\pm$8.65 & 10.12 & 
& 25.00$\pm$13.4 & 3.83\\\hline
\hspace{-0.27cm} Haberman$^\dagger$ & \hspace{-0.49cm} 306 & \hspace{-0.4cm} 3 &
 30.45$\pm$12.4 \hspace{-0.19cm} & & 27.19$\pm$11.2 & &
\cellcolor{Gray}26.15$\pm$9.75 & 5.62 & & 26.49$\pm$9.21 & 3.26 & & 28.12$\pm$10.3 &
3.65 & & 27.47$\pm$9.18 & 9.54\\
\hspace{-0.27cm} Haberman$^\star$ & \hspace{-0.49cm} 306 & \hspace{-0.4cm} 3 &   25.15$\pm$6.53
\hspace{-0.19cm} &  & 25.78$\pm$7.18 &
 & \cellcolor{Gray}24.83$\pm$6.18
& 1.62 & & 25.80$\pm$6.71 & 1.32 & & 25.48$\pm$7.37 & 1.62 & 
& 25.78$\pm$7.18 & 1.65\\\hline
\hspace{-0.27cm} Ionosphere$^\dagger$ & \hspace{-0.49cm} 351 & \hspace{-0.4cm}
33 &   10.56$\pm$4.88 \hspace{-0.19cm} & & 17.68$\pm$7.38 & &
15.11$\pm$6.05 & 4.86 &  & 16.54$\pm$6.46 & 3.42 & & \cellcolor{Gray}15.10$\pm$8.74 &
2.86 & & 17.38$\pm$10.6 & 2.30\\
\hspace{-0.27cm} Ionosphere$^\star$ & \hspace{-0.49cm} 351 & \hspace{-0.4cm} 33 &   13.11$\pm$6.36
\hspace{-0.19cm} &  & 14.51$\pm$7.36 &
& 13.64$\pm$5.99 & 5.43 & & 14.24$\pm$6.15 & 2.83 &  & \cellcolor{Gray}13.38$\pm$4.44 & 3.15
&  & 14.25$\pm$5.04 & 3.41\\\hline
\hspace{-0.27cm} Breastwisc$^\dagger$ & \hspace{-0.49cm} 699 & \hspace{-0.4cm} 9
&    3.44$\pm$2.46 \hspace{-0.19cm} & & 3.72$\pm$2.06 & &
3.87$\pm$3.38 & 1.43 & & \cellcolor{Gray}3.16$\pm$2.44 & 3.14 & & 3.44$\pm$2.65 & 2.14
& & 4.72$\pm$3.02 & 1.14\\
\hspace{-0.27cm} Breastwisc$^\star$ & \hspace{-0.49cm} 699 & \hspace{-0.4cm} 9 &    3.00$\pm$1.96
\hspace{-0.19cm} & & 3.43$\pm$2.25 & & \cellcolor{Gray}2.57$\pm$1.62 &
1.14 &  & 3.29$\pm$2.24 & 0.86 & & 2.86$\pm$2.13 & 0.86 &  & 3.00$\pm$2.18 &
0.29\\\hline
\hspace{-0.27cm} Transfusion$^\dagger$ & \hspace{-0.49cm} 748 & \hspace{-0.4cm}
4 &    39.68$\pm$6.39 \hspace{-0.19cm} & & 39.02$\pm$6.60 & &
\cellcolor{Gray}38.35$\pm$7.06 & 1.99 & & 39.01$\pm$7.63 & 1.88 & & 38.75$\pm$7.15 &
2.28 & & 38.74$\pm$6.67 & 2.14\\
\hspace{-0.27cm} Transfusion$^\star$ & \hspace{-0.49cm} 748 & \hspace{-0.4cm} 4 &    39.17$\pm$7.01
\hspace{-0.19cm} &  & 37.97$\pm$7.42 &
& 37.57$\pm$5.60 & 2.40 &  & 36.50$\pm$6.78 & 2.14 &  & 37.43$\pm$8.08 & 1.21
&  & \cellcolor{Gray}36.10$\pm$8.06 & 3.21\\\hline
\hspace{-0.27cm} Banknote$^\dagger$ & \hspace{-0.49cm} 1 372 & \hspace{-0.4cm} 4
&   2.70$\pm$1.69 \hspace{-0.19cm} & & 15.09$\pm$3.50 & &
14.07$\pm$3.80 & 0.51 & & 14.14$\pm$3.25 & 1.02 & & 14.43$\pm$3.90 &
0.51 & & \cellcolor{Gray}13.27$\pm$3.45 & 2.70\\
\hspace{-0.27cm} Banknote$^\star$ & \hspace{-0.49cm} 1 372 & \hspace{-0.4cm} 4 &   2.70$\pm$1.46
\hspace{-0.19cm} & 
& 14.00$\pm$4.16 &  & \cellcolor{Gray}12.02$\pm$2.74
& 0.73 & & 13.63$\pm$2.75 & 1.39 &
 & 12.17$\pm$2.77 & 0.80 &
& 13.63$\pm$3.02 & 1.39\\\hline
\hspace{-0.27cm} Winered$^\dagger$ & \hspace{-0.49cm} 1 599 & \hspace{-0.4cm} 11
&   26.08$\pm$2.14 \hspace{-0.19cm} & & 28.14$\pm$3.23 & &
\cellcolor{Gray}27.77$\pm$3.83 & 1.38 & & 28.14$\pm$3.71 & 1.26 & & 28.02$\pm$3.64 &
0.88 & & 27.83$\pm$3.73 & 1.06\\
\hspace{-0.27cm} Winered$^\star$ & \hspace{-0.49cm} 1 599 & \hspace{-0.4cm} 11 &   26.33$\pm$2.75
\hspace{-0.19cm} & 
& 28.02$\pm$3.32 &  & 27.83$\pm$3.95
& 1.19 &  & \cellcolor{Gray}27.45$\pm$4.17 & 1.00 &  & 27.58$\pm$3.76 & 1.12 &
 & \cellcolor{Gray}27.45$\pm$3.34 & 1.25\\\hline
\hspace{-0.27cm} Abalone$^\dagger$ & \hspace{-0.49cm} 4 177 & \hspace{-0.4cm} 10
&   23.03$\pm$2.13 \hspace{-0.19cm} & & 26.48$\pm$1.75 & & 25.81$\pm$1.49
& 0.55 & & \cellcolor{Gray}24.18$\pm$0.94 & 0.43 & & 25.93$\pm$1.38 & 0.50 & & 24.40$\pm$0.87 & 0.32\\
\hspace{-0.27cm} Abalone$^\star$ & \hspace{-0.49cm} 4 177 & \hspace{-0.4cm} 10 &   22.98$\pm$2.70
\hspace{-0.19cm} &
 & 26.57$\pm$2.31 &  & 24.18$\pm$2.51
& 0.00 & & 24.13$\pm$2.48 & 0.14 &  & 24.18$\pm$2.51 & 0.00 &
& \cellcolor{Gray}24.11$\pm$2.59 & 0.07\\\hline
\hspace{-0.27cm} Winewhite$^\dagger$ & \hspace{-0.49cm} 4 898 & \hspace{-0.4cm} 11 &  
 30.67$\pm$1.96 \hspace{-0.19cm} & & \cellcolor{Gray}31.77$\pm$2.51 & & 32.32$\pm$2.51 & 1.29 & &
 32.50$\pm$1.92 & 0.51 & & 32.12$\pm$2.67 & 0.98 & & 32.26$\pm$1.82 & 1.14\\
\hspace{-0.27cm} Winewhite$^\star$ & \hspace{-0.49cm} 4 898 & \hspace{-0.4cm} 11 &  
 30.73$\pm$2.20
\hspace{-0.19cm} 
&  & 32.63$\pm$2.52 &
 & \cellcolor{Gray}31.85$\pm$1.66 & 1.18 & & 32.16$\pm$1.73 & 1.31 & & 32.16$\pm$2.02 & 0.90
&  & 31.97$\pm$2.26 & 1.12\\\hline
\hspace{-0.27cm} Smartphone$^\dagger$ & \hspace{-0.49cm} 7 352 &  \hspace{-0.4cm} 561 &
 0.00$\pm$0.00 \hspace{-0.19cm} & & 0.67$\pm$0.24 & & 0.20$\pm$0.13 & 0.00 & &
 0.46$\pm$0.16 & 0.00 & & \cellcolor{Gray}0.19$\pm$0.11 & 0.03 & & 0.22$\pm$0.13 & 0.00\\
\hspace{-0.27cm} Smartphone$^\star$ & \hspace{-0.49cm} 7 352 &  \hspace{-0.4cm} 561 &
 0.00$\pm$0.00 \hspace{-0.19cm} &  & 0.67$\pm$0.25 & &
\cellcolor{Gray} 0.19$\pm$0.22 & 0.00 &
 & 0.44$\pm$0.29 & 0.03 &
 & 0.20$\pm$0.24 & 0.01
& & \cellcolor{Gray} 0.19$\pm$0.22 & 0.04\\\hline
\hspace{-0.27cm} Firmteacher$^\dagger$ & \hspace{-0.49cm} 10 800 &  \hspace{-0.4cm} 16 & 
46.00$\pm$1.32 \hspace{-0.19cm} & & 46.62$\pm$1.86 & & 46.12$\pm$1.80 & 1.22 & &
46.47$\pm$2.18 & 0.72 & & 46.58$\pm$2.38 & 0.78 & & \cellcolor{Gray}46.04$\pm$2.47 & 1.24\\
\hspace{-0.27cm} Firmteacher$^\star$ & \hspace{-0.49cm} 10 800 &  \hspace{-0.4cm} 16 & 
44.44$\pm$1.34 \hspace{-0.19cm} & & 40.58$\pm$4.87 &  &
40.89$\pm$3.95 & 2.35 & & 39.81$\pm$4.37 & 2.89 &   & 38.91$\pm$4.51 &
3.56 &  & \cellcolor{Gray}38.01$\pm$6.15 & 5.02\\\hline
\hspace{-0.27cm} Eeg$^\dagger$ & \hspace{-0.49cm} 14 980 & \hspace{-0.4cm} 14 & 
45.60$\pm$1.96 \hspace{-0.19cm} & & \cellcolor{Gray}43.86$\pm$1.97 & & 44.03$\pm$2.04 & 0.57 & &
43.95$\pm$2.04 & 0.53 & & 43.88$\pm$1.57 & 0.28 & & 44.05$\pm$1.88 & 0.49\\
\hspace{-0.27cm} Eeg$^\star$ & \hspace{-0.49cm} 14 980 & \hspace{-0.4cm} 14 & 
45.38$\pm$2.04 \hspace{-0.19cm} &
 & 44.09$\pm$2.32 &  & 44.01$\pm$1.48 &
0.40 &  & 43.89$\pm$2.19 & 0.89
&  & 44.07$\pm$2.02 & 0.81 &  & \cellcolor{Gray}43.87$\pm$1.40
& 0.95\\\hline
\hspace{-0.27cm} Magic$^\dagger$ & \hspace{-0.49cm} 19 020 & \hspace{-0.4cm} 10
&  21.81$\pm$0.81 \hspace{-0.19cm} & & 37.23$\pm$1.02 & & \cellcolor{Gray}22.38$\pm$0.78 & 0.26 & &
26.41$\pm$0.85 & 0.00 & & 23.11$\pm$1.16 & 1.43 & & 26.41$\pm$0.85 &
0.00\\
\hspace{-0.27cm} Magic$^\star$ & \hspace{-0.49cm} 19 020 & \hspace{-0.4cm} 10 &  21.07$\pm$1.09
\hspace{-0.19cm} &
 & 37.51$\pm$0.46 &  &
\cellcolor{Gray}22.11$\pm$1.32 & 0.28 &  & 26.41$\pm$1.08 & 0.00 & &
23.00$\pm$1.71 & 0.66 &  &
26.41$\pm$1.08 & 0.00\\\hline
\hspace{-0.27cm} Hardware$^\dagger$ & \hspace{-0.49cm}  28 179 &
\hspace{-0.4cm} 96 & 16.84$\pm$0.74 \hspace{-0.19cm} & & 9.41$\pm$0.55 & &
11.52$\pm$0.91 & 0.07 & & 11.51$\pm$1.20 & 0.57 & & 13.88$\pm$0.91 &
0.09 & & \cellcolor{Gray}7.20$\pm$0.64 & 0.05\\
\hspace{-0.27cm} Hardware$^\star$ & \hspace{-0.49cm}  28 179 & \hspace{-0.4cm}  96 &
 16.77$\pm$0.73 \hspace{-0.19cm} &  & 9.41$\pm$0.71 & &
6.43$\pm$0.74 & 0.18 &  & 11.72$\pm$1.24 & 0.41 & & 6.50$\pm$0.67 &
0.10 & & \cellcolor{Gray}6.42$\pm$0.69 & 0.13\\
\Xhline{3\arrayrulewidth}
\hspace{-0.27cm} Marketing$^\dagger$ & \hspace{-0.49cm}  45 211 & \hspace{-0.4cm}
27 & 30.67$\pm$0.61 \hspace{-0.19cm} & & 28.37$\pm$1.21 & &
28.06$\pm$1.04 & 0.45 & & 28.02$\pm$0.88 & 0.00 & & \cellcolor{Gray}27.87$\pm$1.33 &
1.27 & & 28.02$\pm$0.88 & 0.00\\
\hspace{-0.27cm} Marketing$^\star$ & \hspace{-0.49cm} 45 211 & \hspace{-0.4cm} 27 &  30.68$\pm$1.01 \hspace{-0.19cm} & & 27.70$\pm$0.69 & &
27.33$\pm$0.73 & 0.33 &   &
28.02$\pm$0.47 & 0.00 &  & \cellcolor{Gray}27.19$\pm$0.87 &
0.51 &   & 28.02$\pm$0.47 & 0.00\\\hline
\hspace{-0.27cm} Kaggle$^\dagger$ & \hspace{-0.49cm} 120 269 &
\hspace{-0.4cm} 11 &  48.30$\pm$0.67 \hspace{-0.19cm}  & &
44.99$\pm$1.98 & & 44.04$\pm$5.80 & 2.34 &  & 44.37$\pm$3.93 & 1.38 &
& 43.19$\pm$7.17 & 3.04 & & \cellcolor{Gray}42.41$\pm$9.32 & 4.25\\
\hspace{-0.27cm} Kaggle$^\star$ & \hspace{-0.49cm} 120 269 & \hspace{-0.4cm} 11 &  47.80$\pm$0.47 \hspace{-0.19cm} & & 39.22$\pm$8.47 &  &
16.90$\pm$0.51 & 0.00 & & 16.90$\pm$0.51 & 0.00 & & \cellcolor{Gray}16.89$\pm$0.50 &
0.01 & & 16.90$\pm$0.51 & 0.00\\
\hline\hline
\end{tabular}
}
\end{center}
\caption{Best result of \adaboostSS/$\ell_1$-\adaboostSS~\cite{ssIBj,xxrsSA}, vs \radaboostNoL~(with or without
  regularization, trained with $n=m$ random rados (above bold horizontal
  line) / $n=10 000$ rados (below bold horizontal line)), according to the
  expected true error of $\ve{\theta}_T$, when the classifier $\ve{\theta}_T$
  returned is the last classifier of the sequence ("$^\dagger$";
  $\ve{\theta}_T=\ve{\theta}_{1000}$), or when it is the classifier
  minimizing the empirical risk in the sequence ("$^\star$";
  $\ve{\theta}_T=\ve{\theta}_{\leq 1000}$). Table shows the best result
  over all $\upomega$s, as well as 
  the difference between the worst and best
  ($\Delta$). Shaded cells display the best result of
  \radaboostNoL. All domains but Kaggle are
  UCI \cite{blUR}.}
  \label{tc1_errs_rr_all}
\end{sidewaystable}

\begin{sidewaystable}[t]
\begin{center}
{\footnotesize
\begin{tabular}{|crr||r|r||r|r|r|r|r|}
\hline \hline
 & & &   \adaboostSS  & $\ell_1$-\adaboostSS &
 \multicolumn{5}{c|}{{\scriptsize \radaboostNoL}} \\ 
 & & & & & \multicolumn{1}{c|}{$\upomega = 0$} & \multicolumn{1}{c|}{$\Omega = \|.\|_{\matrice{\tiny{i}}_d}^2$} & \multicolumn{1}{c|}{$\Omega = \|.\|_1$} &
 \multicolumn{1}{c|}{$\Omega = \|.\|_\infty$} & \multicolumn{1}{c|}{$\Omega = \|.\|_{\Phi}$}\\
domain & $m$ & $d$ & \multicolumn{1}{c|}{supp.$\pm\sigma$} & \multicolumn{1}{c||}{supp.$\pm\sigma$} &
\multicolumn{1}{c|}{supp.$\pm\sigma$} &
\multicolumn{1}{c|}{supp.$\pm\sigma$} &
\multicolumn{1}{c|}{supp.$\pm\sigma$} &
\multicolumn{1}{c|}{supp.$\pm\sigma$} & 
\multicolumn{1}{c|}{supp.$\pm\sigma$}\\ \hline
Fertility$^\dagger$ & 100 & 9 & 100.00$\pm$0.00 & 11.11$\pm$0.00 & 86.67$\pm$10.2 & 86.67$\pm$10.2 & 82.22$\pm$15.0 & 93.33$\pm$7.77 & 83.33$\pm$10.8\\
Fertility$^\star$ & 100 & 9 & 36.67$\pm$36.3 & 14.44$\pm$5.36 & 37.78$\pm$31.1 &
36.67$\pm$34.8 &  42.22$\pm$31.0 &  24.44$\pm$22.1 &
32.22$\pm$17.7\\ \hline
Sonar$^\dagger$ & 208 & 60 & 58.50$\pm$4.04 & 2.50$\pm$1.41& 15.67$\pm$3.62 & 16.50$\pm$2.88 & 15.17$\pm$3.09 & 14.67$\pm$3.50 & 14.50$\pm$2.94\\
Sonar$^\star$ & 208 & 60 & 57.83$\pm$3.69 & 1.83$\pm$0.52 &  14.17$\pm$3.62 & 14.00$\pm$3.16 &
13.67$\pm$3.99 & 12.83$\pm$4.45 &  12.67$\pm$4.17\\\hline
Haberman$^\dagger$ & 306 & 3 & 100.00$\pm$0.00 & 33.33$\pm$0.00& 66.67$\pm$22.2 & 60.00$\pm$26.3 & 73.33$\pm$21.1 & 66.67$\pm$22.2 & 76.67$\pm$22.5\\
Haberman$^\star$ & 306 & 3 & 70.00$\pm$10.6 & 33.33$\pm$0.00 &  66.67$\pm$22.2 &
 66.67$\pm$15.7 & 56.67$\pm$16.1 & 60.00$\pm$26.3 &
 50.00$\pm$17.6\\\hline
Ionosphere$^\dagger$ & 351 & 33 & 80.30$\pm$4.34 & 3.94$\pm$2.04 & 15.15$\pm$2.47 & 14.24$\pm$2.05 & 13.64$\pm$3.27 & 13.64$\pm$2.58 & 15.15$\pm$2.86\\
Ionosphere$^\star$ & 351 & 33 & 76.97$\pm$8.23 & 3.64$\pm$1.27 &  13.64$\pm$3.85 &
13.03$\pm$3.51 & 11.82$\pm$3.63 &  11.21$\pm$4.75 & 
11.21$\pm$4.53\\\hline
Breastwisc$^\dagger$ & 699 & 9 & 100.00$\pm$0.00 & 11.11$\pm$0.00& 50.00$\pm$12.0 & 90.00$\pm$3.51 & 57.78$\pm$8.76 & 87.78$\pm$3.51 & 84.44$\pm$7.77\\
Breastwisc$^\star$ & 699 & 9 & 90.00$\pm$3.51 & 11.11$\pm$0.00 & 51.11$\pm$12.0 &
84.44$\pm$7.77 &  48.89$\pm$9.37 & 84.44$\pm$10.7 & 
86.67$\pm$4.68\\\hline
Transfusion$^\dagger$ & 748 & 4 & 100.00$\pm$0.00 & 25.00$\pm$0.00& 75.00$\pm$11.8 & 87.50$\pm$13.2 & 80.00$\pm$15.8 & 80.00$\pm$15.8 & 85.00$\pm$12.9\\
Transfusion$^\star$ & 748 & 4 & 77.50$\pm$14.2 & 25.00$\pm$0.00 &  67.50$\pm$20.6 &
70.00$\pm$23.0 &  67.50$\pm$16.9 &  67.50$\pm$23.7 &
 72.50$\pm$14.2\\\hline
Banknote$^\dagger$ & 1 372 & 4 & 100.00$\pm$0.00 & 25.00$\pm$0.00& 37.50$\pm$13.2 & 50.00$\pm$0.00 & 50.00$\pm$0.00 & 50.00$\pm$0.00 & 35.00$\pm$12.9\\
Banknote$^\star$ & 1 372 & 4 & 100.00$\pm$0.00 & 25.00$\pm$0.00 &  40.00$\pm$12.9 &
 50.00$\pm$0.00 & 47.50$\pm$7.91 &  50.00$\pm$0.00 &
47.50$\pm$7.91\\\hline
Winered$^\dagger$ & 1 599 & 11 & 84.55$\pm$6.14 & 9.09$\pm$0.00& 27.27$\pm$7.42 & 27.27$\pm$7.42 & 31.82$\pm$6.43 & 29.09$\pm$9.39 & 26.36$\pm$7.96\\
Winered$^\star$ & 1 599 & 11 & 79.09$\pm$6.14 & 9.09$\pm$0.00 &  25.45$\pm$5.75 &  27.27$\pm$6.06
&  27.27$\pm$7.42 &  25.45$\pm$7.17 & 
27.27$\pm$7.42\\\hline
Abalone$^\dagger$ & 4 177 & 10 & 65.00$\pm$5.27 & 19.00$\pm$3.16& 40.00$\pm$13.3 & 83.00$\pm$4.83 & 10.00$\pm$0.00 & 86.00$\pm$5.16 & 25.00$\pm$24.2\\
Abalone$^\star$ & 4 177 & 10 & 64.00$\pm$6.99 & 19.00$\pm$3.16 &  30.00$\pm$6.67 &  10.00$\pm$0.00
& 12.00$\pm$6.32 &  10.00$\pm$0.00 & 11.00$\pm$3.16\\\hline
Winewhite$^\dagger$ & 4 898 & 11 & 65.45$\pm$7.17 & 9.09$\pm$0.00& 27.27$\pm$0.00 & 30.00$\pm$6.14 & 19.09$\pm$2.87 & 28.18$\pm$5.16 & 18.18$\pm$0.00\\
Winewhite$^\star$ & 4 898 & 11 & 66.36$\pm$9.63 & 9.09$\pm$0.00 &  28.18$\pm$2.87 &
 28.18$\pm$2.87 & 20.91$\pm$4.39 & 27.27$\pm$0.00 & 
18.18$\pm$0.00\\\hline
Smartphone$^\dagger$ & 7 352 & 561 & 5.63$\pm$0.27 & 0.36$\pm$0.00& 0.18$\pm$0.00 & 78.81$\pm$0.33 & 0.18$\pm$0.00 & 81.48$\pm$0.27 & 40.98$\pm$0.27\\
Smartphone$^\star$ & 7 352 & 561 & 5.53$\pm$0.24 & 0.36$\pm$0.00 & 
0.18$\pm$0.00 & 71.21$\pm$20.1 &  0.18$\pm$0.00 & 
74.72$\pm$19.7 & 24.69$\pm$9.87\\\hline
Firmteacher$^\dagger$ & 10 800 & 16 & 100.00$\pm$0.00 & 10.00$\pm$3.23& 40.62$\pm$6.07 & 39.38$\pm$7.82 & 42.50$\pm$7.10 & 40.00$\pm$7.34 & 42.50$\pm$6.45\\
Firmteacher$^\star$ & 10 800 & 16 & 48.12$\pm$30.8 & 10.00$\pm$3.23 &
24.38$\pm$7.48 &  25.62$\pm$9.52 & 21.25$\pm$4.37 & 
20.62$\pm$4.22 &  20.62$\pm$9.34\\\hline
Eeg$^\dagger$ & 14 980 & 14 & 15.00$\pm$2.26 & 8.57$\pm$3.01& 47.14$\pm$6.90 & 47.86$\pm$5.88 & 42.14$\pm$7.86 & 45.00$\pm$4.82 & 47.14$\pm$6.02\\
Eeg$^\star$ & 14 980 & 14 & 14.29$\pm$3.37 & 8.57$\pm$3.01 &  39.29$\pm$13.2 &  38.57$\pm$9.04 &
 39.29$\pm$14.0 &  38.57$\pm$13.1 & 
39.29$\pm$10.8\\\hline
Magic$^\dagger$ & 19 020 & 10 & 78.00$\pm$6.32 & 10.00$\pm$0.00& 11.00$\pm$3.16 & 49.00$\pm$3.16 & 10.00$\pm$0.00 & 50.00$\pm$0.00 & 10.00$\pm$0.00\\
Magic$^\star$ & 19 020 & 10 & 45.00$\pm$7.07 & 10.00$\pm$0.00 &  10.00$\pm$0.00 &
 51.00$\pm$3.16 &  10.00$\pm$0.00 & 49.00$\pm$7.38 &
 10.00$\pm$0.00\\\hline
Hardware$^\dagger$ & 28 179 & 96 & 16.35$\pm$0.70 & 2.91$\pm$0.82& 1.04$\pm$0.00 & 91.67$\pm$0.00 & 1.04$\pm$0.00 & 91.67$\pm$0.00 & 65.21$\pm$0.54\\
Hardware$^\star$ & 28 179 & 96 & 11.98$\pm$7.56 & 2.19$\pm$0.33 & 
1.04$\pm$0.00 & 20.94$\pm$3.12 &  1.04$\pm$0.00 & 
22.08$\pm$1.89 & 21.25$\pm$1.49\\\Xhline{3\arrayrulewidth}
Marketing$^\dagger$ & 45 211 & 27 & 64.07$\pm$3.51 & 7.41$\pm$0.00 & 8.89$\pm$1.91 & 15.93$\pm$3.51 & 3.70$\pm$0.00 & 16.67$\pm$3.15 & 3.70$\pm$0.00\\
Marketing$^\star$ & 45 211 & 27 & 65.19$\pm$3.58 & 7.40$\pm$0.00 &
7.41$\pm$0.00 &
12.96$\pm$3.60 &  3.70$\pm$0.00 &  13.33$\pm$4.35 &
3.70$\pm$0.00\\\hline
Kaggle$^\dagger$ & 120 269 & 11 & 36.36$\pm$0.00 & 18.18$\pm$0.00& 19.09$\pm$2.87 & 33.64$\pm$6.14 & 18.18$\pm$0.00 & 40.00$\pm$6.36 & 19.09$\pm$2.87\\
Kaggle$^\star$ & 120 269 & 11 & 28.18$\pm$5.16 & 18.18$\pm$0.00 &  17.27$\pm$2.87 &  9.09$\pm$0.00 & 15.45$\pm$4.39 & 10.00$\pm$2.87 & 14.55$\pm$4.69\\
\hline\hline
\end{tabular}
}
\end{center}
\caption{Supports of \adaboostSS~and $\ell_1$-\adaboostSS~vs \radaboostNoL~for the results
  displayed in Table  \ref{tc1_errs_rr_all} (supp.\%($\ve{\theta}$) $\defeq 100\cdot
  \|\ve{\theta}\|_0/d$), when the classifier $\ve{\theta}_T$
  returned is the last classifier of the sequence ("$^\dagger$";
  $\ve{\theta}_T=\ve{\theta}_{1000}$), or when it is the classifier
  minimizing the empirical risk in the sequence ("$^\star$";
  $\ve{\theta}_T=\ve{\theta}_{\leq 1000}$).}
  \label{tc1_supp_rr_all}
\end{sidewaystable}

\end{document}